\documentclass[oneside,11pt]{article}

\usepackage{times,url,mathrsfs}
\usepackage{amsmath,wrapfig,color}
\usepackage{amsfonts}
\usepackage{graphicx}
\usepackage{subfigure}
\newtheorem{theorem}{Theorem}
\newtheorem{lemma}{Lemma}

\begin{document}
\title{\huge Coding for Random Projections\vspace{0.5in}}

\author{ \bf{Ping Li} \\
         Department of Statistics \& Biostatistics\\\hspace{0.1in}
         Department of Computer Science\\
       Rutgers University\\
          Piscataway, NJ 08854\\
       \texttt{pl314@rci.rutgers.edu}\\\\
       \and
         \bf{Michael Mitzenmacher}\\
         School of Engineering and Applied Sciences\\
         Harvard University\\
         Cambridge, MA 02138\\
         \texttt{michaelm@eecs.harvard.edu}
       \and
         \bf{Anshumali Shrivastava}\\
         Department of Computer Science\\
         Cornell University\\
         {Ithaca, NY 14853}\\
        \texttt{anshu@cs.cornell.edu}}

\date{}

\maketitle

\begin{abstract}
The method of random projections has become very popular for large-scale applications in statistical learning, information retrieval, bio-informatics  and other applications.  Using a well-designed \textbf{coding} scheme for the projected data, which determines the number of bits needed for each projected value and how to allocate these bits, can significantly improve the effectiveness of the algorithm, in storage cost as well as computational speed.   In this paper, we study a number of simple coding schemes, focusing on the task of similarity estimation and on an application to training linear classifiers. We demonstrate that \textbf{uniform quantization} outperforms the standard existing influential method~\cite{Proc:Datar_SCG04}. Indeed, we argue that in many cases coding with just a small number of bits suffices.  Furthermore, we also  develop a \textbf{non-uniform 2-bit} coding scheme that generally performs well in practice, as confirmed by our experiments on training linear support vector machines (SVM).

\end{abstract}

\newpage\clearpage

\section{Introduction}

The method of random projections has become popular for large-scale machine learning applications such as classification, regression, matrix factorization, singular value decomposition, near neighbor search, bio-informatics, and more~\cite{Proc:Papadimitriou_PODS98,Proc:Dasgupta_FOCS99,Proc:Bingham_KDD01,Article:Buher_Tompa,Proc:Fradkin_KDD03,Proc:Li_Hastie_Church_COLT06,Proc:Frund_NIPS08,Book:Vempala,Proc:Dasgupta_UAI00,Article:JL84,Proc:Wang_Li_SDM10}. In this paper, we study a number of simple and effective schemes for  \textbf{coding} the projected data, with the focus on   similarity estimation and training linear classifiers~\cite{Proc:Joachims_KDD06,Proc:Shalev-Shwartz_ICML07,Article:Fan_JMLR08,URL:Bottou_SGD}. We will closely compare our method with the influential prior coding scheme in~\cite{Proc:Datar_SCG04}.\\

Consider two high-dimensional vectors, $u, v\in\mathbb{R}^D$. The idea is to multiply them with a random normal projection matrix $\mathbf{R}\in\mathbb{R}^{D\times k}$ (where $k\ll D$), to generate two (much) shorter vectors $x, y$:
\begin{align}
x = u\times \mathbf{R} \in\mathbb{R}^k,\hspace{0.2in} y = v\times \mathbf{R} \in\mathbb{R}^k, \hspace{0.2in} \mathbf{R} = \{r_{ij}\}{_{i=1}^D}{_{j=1}^k}, \hspace{0.2in} r_{ij} \sim N(0,1) \text{ i.i.d. }
\end{align}
In real applications, the dataset will consist of a  large number of  vectors (not just two). Without loss of generality, we use one pair of data vectors ($u, v$) to demonstrate our results.\\

In this study, for convenience, we assume that the marginal Euclidian norms of the original data vectors, i.e., $\|u\|, \|v\|$, are  known. This assumption is  reasonable in practice~\cite{Proc:Li_Hastie_Church_COLT06}.  For example, the input data for feeding to a support vector machine (SVM) are usually normalized, i.e.,  $\|u\|=\|v\|=1$.  Computing the marginal norms for the entire dataset only requires one linear scan of the data, which is anyway needed during  data collection/processing. Without loss of generality, we assume $\|u\|=\|v\|=1$ in this paper.  The joint distribution of $(x_j, y_j)$ is hence a bi-variant normal:
\begin{align}
\left[\begin{array}{c}x_j\\ y_j\end{array} \right] \sim N
\left(
\left[\begin{array}{c}0\\ 0\end{array} \right],\
\left[\begin{array}{cc}1 &\rho\\ \rho &1
\end{array} \right]
\right), \text{ i.i.d.}\hspace{0.25in} j = 1, 2, ..., k.
\end{align}
where $\rho = \sum_{i=1}^D u_iv_i$ (assuming $\|u\|=\|v\|=1$). For convenience and brevity, we also restrict our attention to $\rho\geq 0$, which is a common scenario in practice. Throughout the paper, we adopt the conventional notation for the standard normal pdf  $\phi(x)$ and cdf $\Phi(x)$:
\begin{align}
\phi(x) = \frac{1}{\sqrt{2\pi}} e^{-\frac{x^2}{2}},\hspace{0.5in} \Phi(x) = \int_{-\infty}^x \phi(x) dx
\end{align}

\subsection{Uniform Quantization}

Our first proposal is perhaps the most intuitive scheme, based on a simple uniform quantization:
\begin{align}\label{eqn_hw}
h_{w}^{(j)}(u) = \left\lfloor x_j/w\right\rfloor,\hspace{0.5in} h_{w}^{(j)}(v) = \left\lfloor y_j/w\right\rfloor
\end{align}
where $w>0$ is the bin width and $\left\lfloor . \right\rfloor$ is the standard floor operation, i.e., $\left\lfloor z \right\rfloor$ is the largest integer which is smaller than or equal to $z$. For example, $\lfloor 3.1\rfloor = 3$, $\lfloor 4.99\rfloor = 4$, $\lfloor -3.1\rfloor = -4$. Later in the paper we will also use the standard ceiling operation $\lceil .\rceil$. We show that the collision probability $P_{w}=\mathbf{Pr}\left(h_{w}^{(j)}(u) = h_{w}^{(j)}(v) \right)$ is a monotonically increasing function of the similarity $\rho$, making (\ref{eqn_hw}) a suitable coding scheme for similarity estimation and  near neighbor search.

The potential benefits of coding with a small number of bits arise because the (uncoded) projected data, $x_j = \sum_{i=1}^D u_i r_{ij}$ and $y_j = \sum_{i=1}^D v_i r_{ij}$, being real-valued numbers, are neither convenient/economical for storage and transmission, nor well-suited for indexing.\\ 

Since the original data are assumed to be normalized, i.e., $\|u\|=\|v\|=1$, the marginal distribution of $x_j$ (and $y_j$) is the standard normal, which decays rapidly at the tail, e.g., $1-\Phi(3) = 10^{-3}$, $1-\Phi(6) = 9.9\times10^{-10}$. If we use $6$ as cutoff,  i.e., values with absolute value greater than 6 are just treated as $-6$ and 6,
then the number of bits needed to represent the bin the value lies in is $1+\log_2 \left\lceil \frac{6}{w}\right\rceil$. In particular, if we choose the bin width $w\geq6$, we can just record the sign of the outcome (i.e., a one-bit scheme). In general, the optimum choice of  $w$ depends on
the similarity $\rho$ and the task.  In this paper we focus on the task of similarity estimation (of $\rho$) and we will provide the optimum $w$ values for all similarity levels. Interestingly, using our uniform quantization scheme,  we find in a certain range the optimum $w$ values are quite large, and in particular are larger than 6.\\

We can  build \textbf{linear classifier} (e.g., linear SVM) using coded random projections. For example, assume the projected values are within $(-6,6)$.  If $w = 2$, then the code values output by $h_w$ will be within the set  $\{-3, -2, -1, 0, 1, 2\}$.  This means  we can represent a projected value using a vector of length 6 (with exactly one 1) and the total length of the new feature vector (fed to a linear SVM) will be $6\times k$. See more details in Section~\ref{sec_SVM}. This trick was also recently used for linear learning with binary data based on  {\em b-bit minwise hashing}~\cite{Proc:Li_Konig_WWW10,Proc:Li_Owen_Zhang_NIPS12}. Of course, we can also use the method to speed up kernel evaluations for kernel SVM  with high-dimensional data.

\textbf{Near neighbor search} is a basic problem studied since the early days of modern computing~\cite{Article:Friedman_75} with applications throughout computer science.  The use of coded projection data for near neighbor search is closely related to {\em locality  sensitive hashing (LSH)}~\cite{Proc:Indyk_STOC98}.  For example, using $k$ projections and a bin width $w$, we can naturally build a hash table with $\left(2 \lceil\frac{6}{w}\rceil\right)^k$
buckets. We map every data vector in the dataset to one of the buckets. For a query data vector, we search for similar data vectors in the same bucket.  Because the concept of LSH is well-known, we do not elaborate on the details. Compared to~\cite{Proc:Datar_SCG04}, our proposed coding scheme has better performance  for near neighbor search; the analysis will be reported in a separate technical report. This paper focuses on similarity estimation.

\subsection{Advantages over the Window-and-Offset Coding Scheme}

 \cite{Proc:Datar_SCG04} proposed the following well-known coding scheme, which uses
windows and a random offset:
\begin{align}\label{eqn_hwq}
h_{w,q}^{(j)}(u) = \left\lfloor\frac{x_j + q_j}{w}\right\rfloor,\hspace{0.3in} h_{w,q}^{(j)}(v) = \left\lfloor\frac{y_j + q_j}{w}\right\rfloor
\end{align}
where $q_j\sim uniform(0,w)$. \cite{Proc:Datar_SCG04} showed that the collision probability can be written as
\begin{align}\label{eqn_Pwq}
P_{w,q} = &\mathbf{Pr}\left(h_{w,q}^{(j)}(u) = h_{w,q}^{(j)}(v)\right)
= \int_0^w\frac{1}{\sqrt{d}}2\phi\left(\frac{t}{\sqrt{d}}\right)\left(1-\frac{t}{w}\right)dt
\end{align}
where $d = ||u-v||^2= 2(1-\rho)$ is the Euclidean distance between $u$ and $v$.  The difference between (\ref{eqn_hwq}) and our proposal (\ref{eqn_hw}) is that we do not use the additional randomization with $q\sim uniform(0,w)$ (i.e., the offset). By comparing them closely, we will demonstrate the following advantages of our scheme:\vspace{-0.0in}
\begin{enumerate}
\item Operationally, our scheme $h_{w}$ is  simpler than $h_{w,q}$. \vspace{-0.0in}
\item With a fixed $w$,  our scheme $h_{w}$  is always more accurate than $h_{w,q}$, often significantly so.\vspace{-0in}
\item For each coding scheme, we can separately find the optimum bin width $w$. We will show that the optimized  $h_w$ is also more accurate than optimized $h_{w,q}$, often significantly so.\vspace{0in}
\item For a wide range of  $\rho$ values (e.g., $\rho<0.56$), the optimum $w$ values for our scheme $h_w$ are relatively large (e.g., $>6$), while for the existing scheme $h_{w,q}$, the optimum $w$ values are  small (e.g., about $1$). This means $h_w$ requires a smaller number of bits than $h_{w,q}$.
\end{enumerate}
In summary, uniform quantization is simpler, more accurate, and uses fewer bits than the influential prior  work~\cite{Proc:Datar_SCG04} which uses the window with the random offset.

\subsection{Organization}

In Section~\ref{sec_hw}, we analyze the collision probability for the  uniform quantization  scheme and then compare it with  the collision probability of the well-known prior work~\cite{Proc:Datar_SCG04} which uses an additional random offset. Because the collision probabilities are monotone functions of the similarity $\rho$, we can always estimate $\rho$ from the observed (empirical) collision probabilities. In Section~\ref{sec_compare_hwq}, we theoretically compare the estimation variances of these two schemes and conclude that  the random offset step in~\cite{Proc:Datar_SCG04} is not needed.  \\

In Section~\ref{sec_hw2}, we develop a 2-bit non-unform coding scheme and demonstrate that its performance largely matches the performance of the uniform quantization scheme (which requires storing more bits). Interestingly, for certain range of the similarity $\rho$, we observe that only  one bit is needed. Thus, Section~\ref{sec_h1} is devoted to comparing the 1-bit scheme with our proposed methods. The comparisons show that the 1-bit scheme does not perform as well when the similarity $\rho$ is high (which is often the case applications are interested in). In Section~\ref{sec_SVM}, we provide a set of experiments on training linear SVM using all the coding schemes we have studied. The experimental results basically confirm the variance analysis. Section~\ref{sec_future} presents several directions for related future research. Finally, Section~\ref{sec_conclusion} concludes the paper.

\section{The Collision Probability of  Uninform Quantization $h_w$}\label{sec_hw}

To use our coding scheme $h_w$ (\ref{eqn_hw}), we  need  to evaluate $P_{w} = \mathbf{Pr}\left(h_{w}^{(j)}(u) = h^{(j)}_{w}(v)\right)$, the collision probability. From practitioners' perspective, as long as $P_w$ is a monotonically increasing function of the similarity $\rho$, it is a suitable coding scheme. In other words, it does not  matter whether $P_w$ has a closed-form expression, as long as we can  demonstrate its advantage over the alternative~\cite{Proc:Datar_SCG04}, whose collision probability is denoted by $P_{w,q}$.  Note that $P_{w,q}$ can be expressed in a closed-form in terms of the standard $\phi$ and $\Phi$ functions:
\begin{align}\label{eqn_Pwq2}
P_{w,q} = \mathbf{Pr}\left(h_{w,q}^{(j)}(u) = h_{w,q}^{(j)}(v)\right)
=& 2\Phi\left(\frac{w}{\sqrt{d}}\right)-1-\frac{2}{\sqrt{2\pi}w/\sqrt{d}}+\frac{2}{w/\sqrt{d}}\phi\left(\frac{w}{\sqrt{d}}\right)
\end{align}
Recall $d = 2(1-\rho)$ is the Euclidean distance $\|u-v\|^2$. It is clear that $P_{w,q}\rightarrow 1$ as $w\rightarrow\infty$.\\

The following Lemma~\ref{lem_Pst} will help derive the collision probability $P_w$ (in Theorem~\ref{thm_Pw}).
\begin{lemma}\label{lem_Pst}
Assume $\left[\begin{array}{c}x\\ y\end{array} \right] \sim N
\left(
\left[\begin{array}{c}0\\ 0\end{array} \right],\
\left[\begin{array}{cc}1 &\rho\\ \rho &1
\end{array} \right]
\right)$, $\rho\geq 0$. Then
\begin{align}
&Q_{s,t}\left(\rho\right)  = \mathbf{Pr}\left(x\in[s,t],\ y\in[s,t]\right) = \int_{s}^{t}\phi(z)\left[\Phi\left(\frac{t-\rho z}{\sqrt{1-\rho^2}}\right)- \Phi\left(\frac{s-\rho z}{\sqrt{1-\rho^2}}\right)\right]dz\\
&\frac{\partial Q_{s,t}(\rho)}{\partial \rho} =\frac{1}{2\pi}\frac{1}{(1-\rho^2)^{1/2}}\left(e^{-\frac{t^2}{(1+\rho)}}+e^{-\frac{s^2}{(1+\rho)}} -2e^{-\frac{t^2+s^2-2st\rho}{2(1-\rho^2)}}\right)  \geq 0
\end{align}
\noindent\textbf{Proof:}\ \ See Appendix~\ref{app_lem_Pst}.\hfill$\Box$\\
\end{lemma}

\begin{theorem}\label{thm_Pw}
The collision probability of the coding scheme $h_w$ defined in (\ref{eqn_hw}) is
\begin{align}\label{eqn_Pw}
P_{w} =2\sum_{i=0}^\infty\int_{iw}^{(i+1)w}\phi(z)\left[\Phi\left(\frac{(i+1)w-\rho z}{\sqrt{1-\rho^2}}\right)- \Phi\left(\frac{iw-\rho z}{\sqrt{1-\rho^2}}\right)\right]dz
\end{align}
which is a monotonically increasing function of $\rho$.  In particular, when $\rho = 0$, we have
\begin{align}
P_{w} = 2\sum_{i=0}^\infty \left[\Phi\left((i+1)w\right)- \Phi\left(iw\right)\right]^2
\end{align}
\noindent\textbf{Proof:}\ \ The proof follows from Lemma~\ref{lem_Pst} by using $s=iw$ and $t=(i+1)w$, $i=0, 1, ...$.  \hfill$\Box$\\
\end{theorem}

\begin{figure}[h!]
\begin{center}
\mbox{
\includegraphics[width = 2.2in]{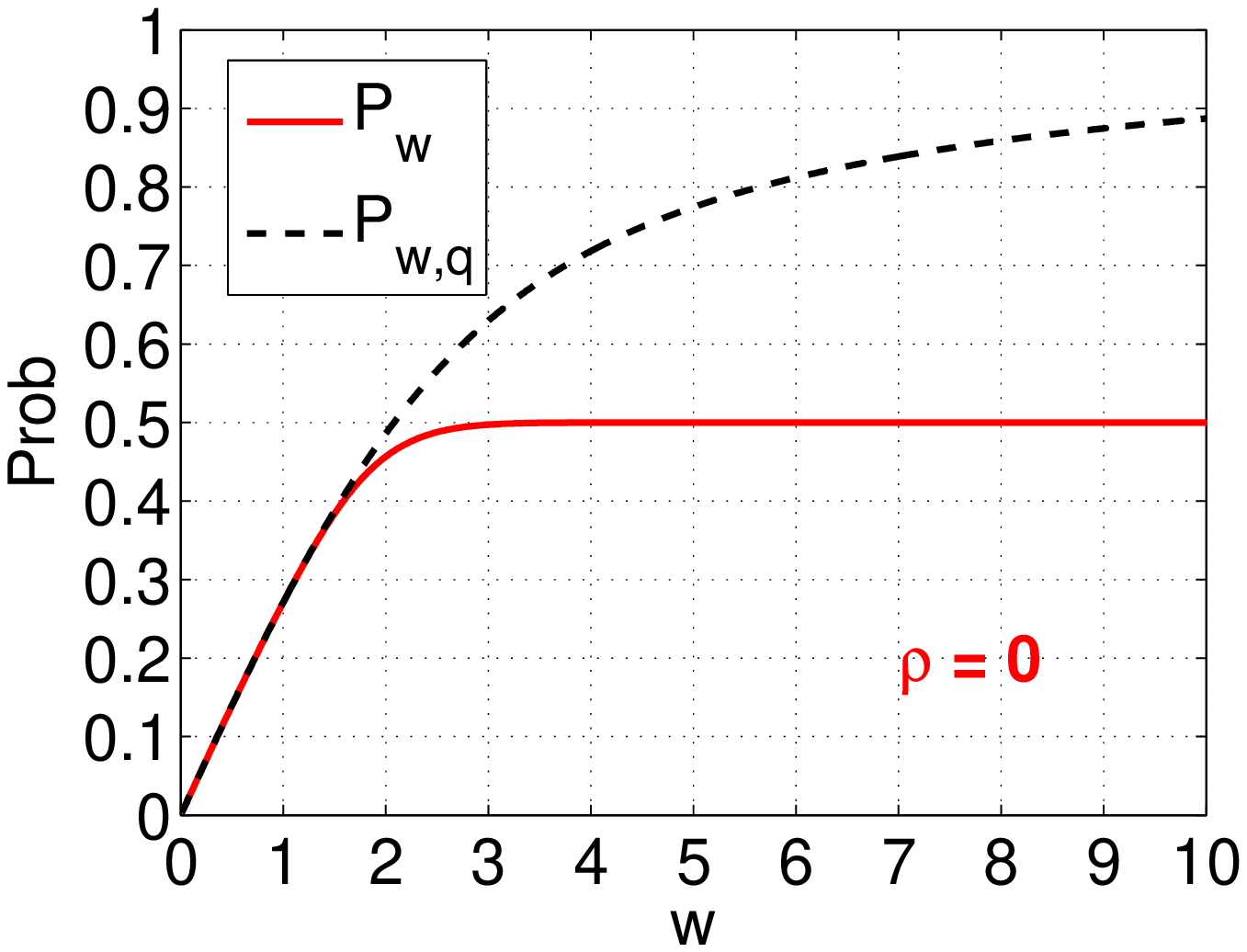}
\includegraphics[width = 2.2in]{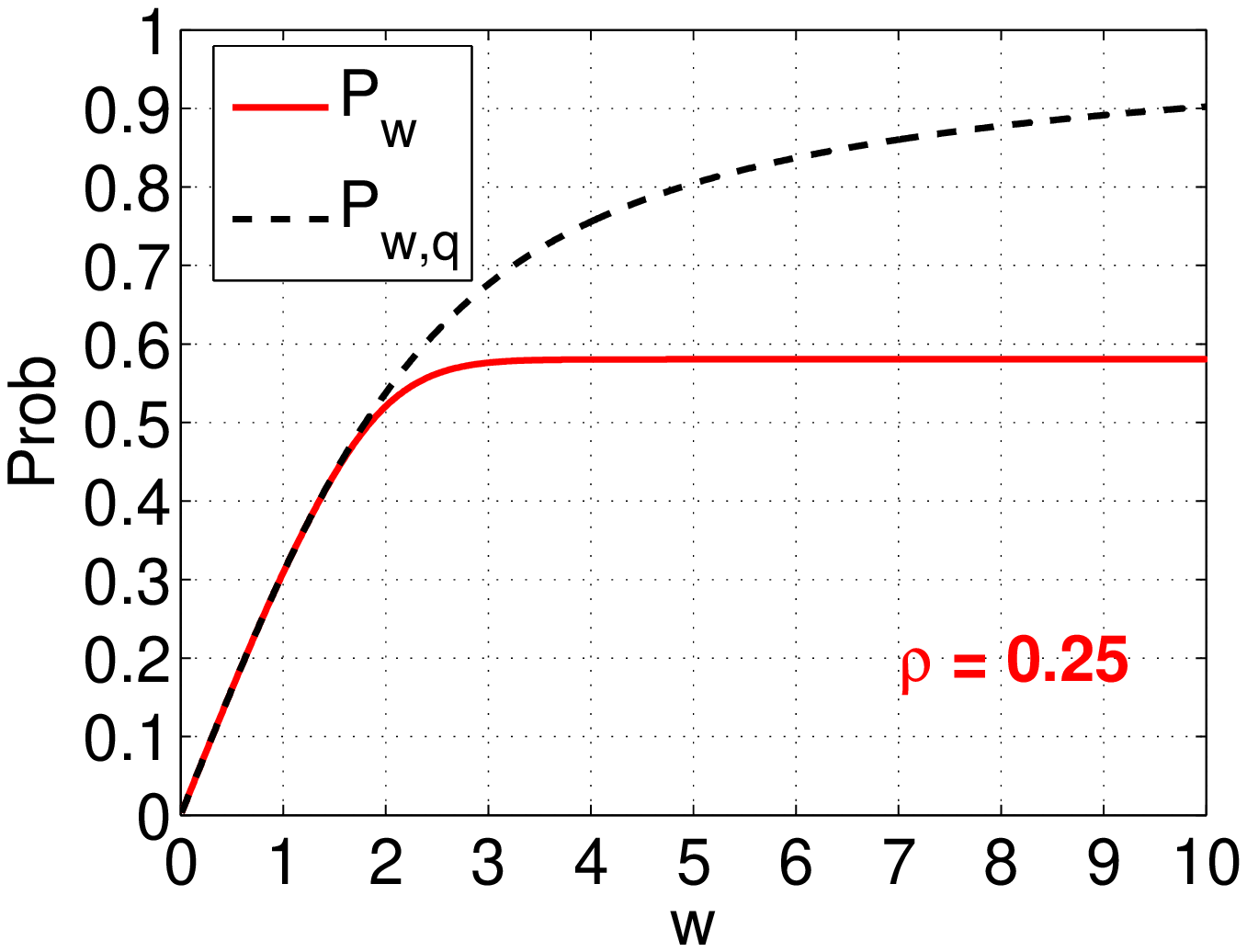}
\includegraphics[width = 2.2in]{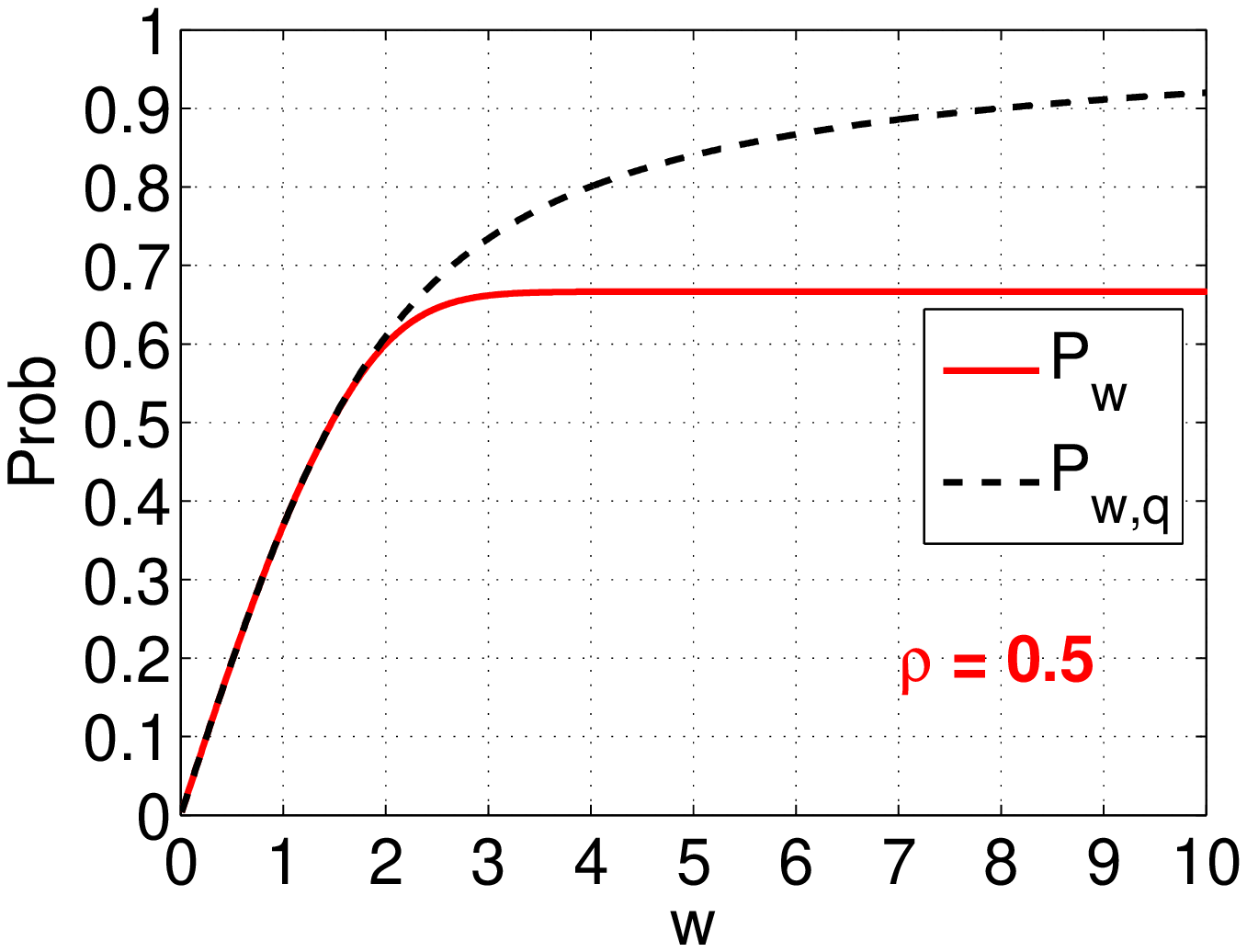}
}
\mbox{
\includegraphics[width = 2.2in]{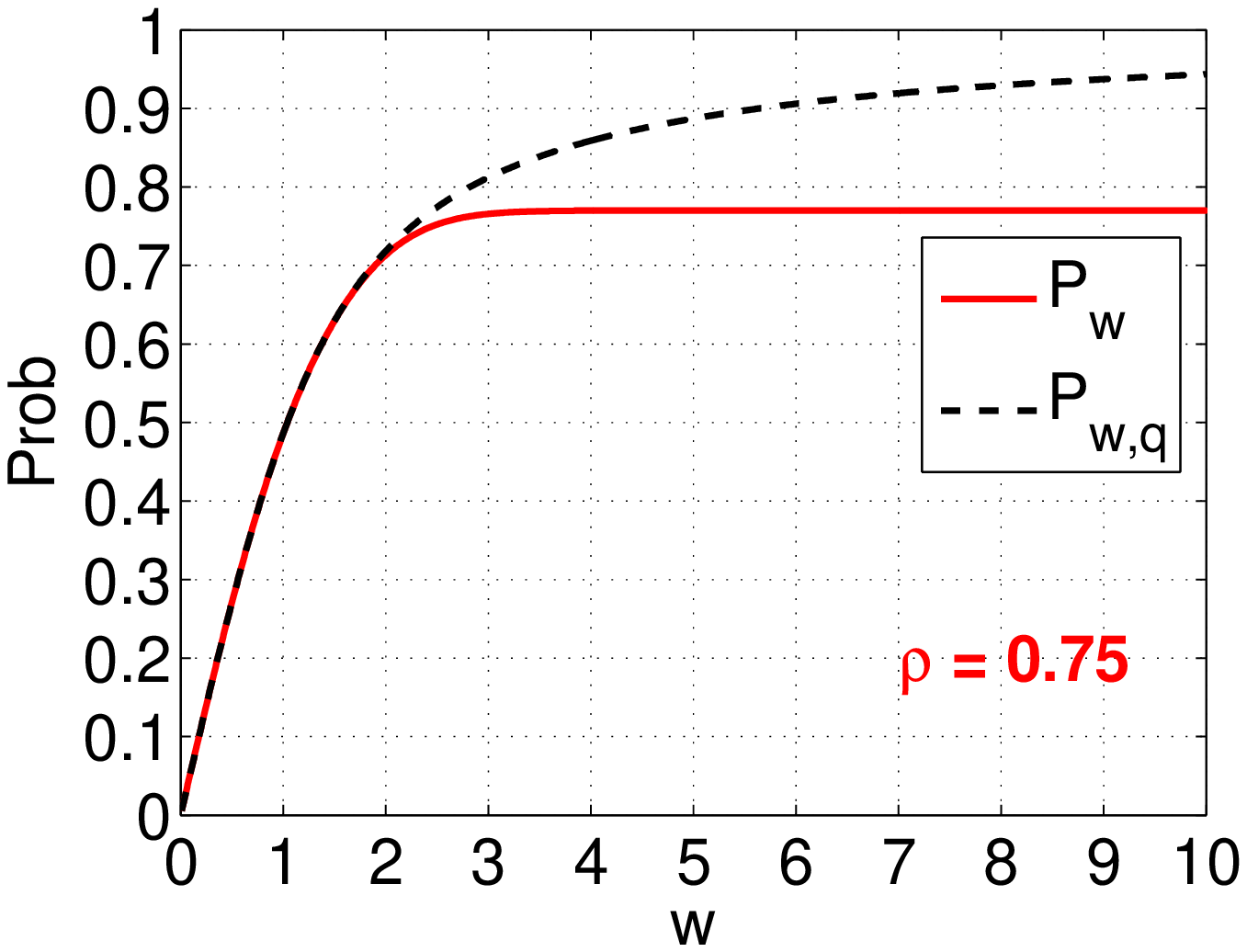}
\includegraphics[width = 2.2in]{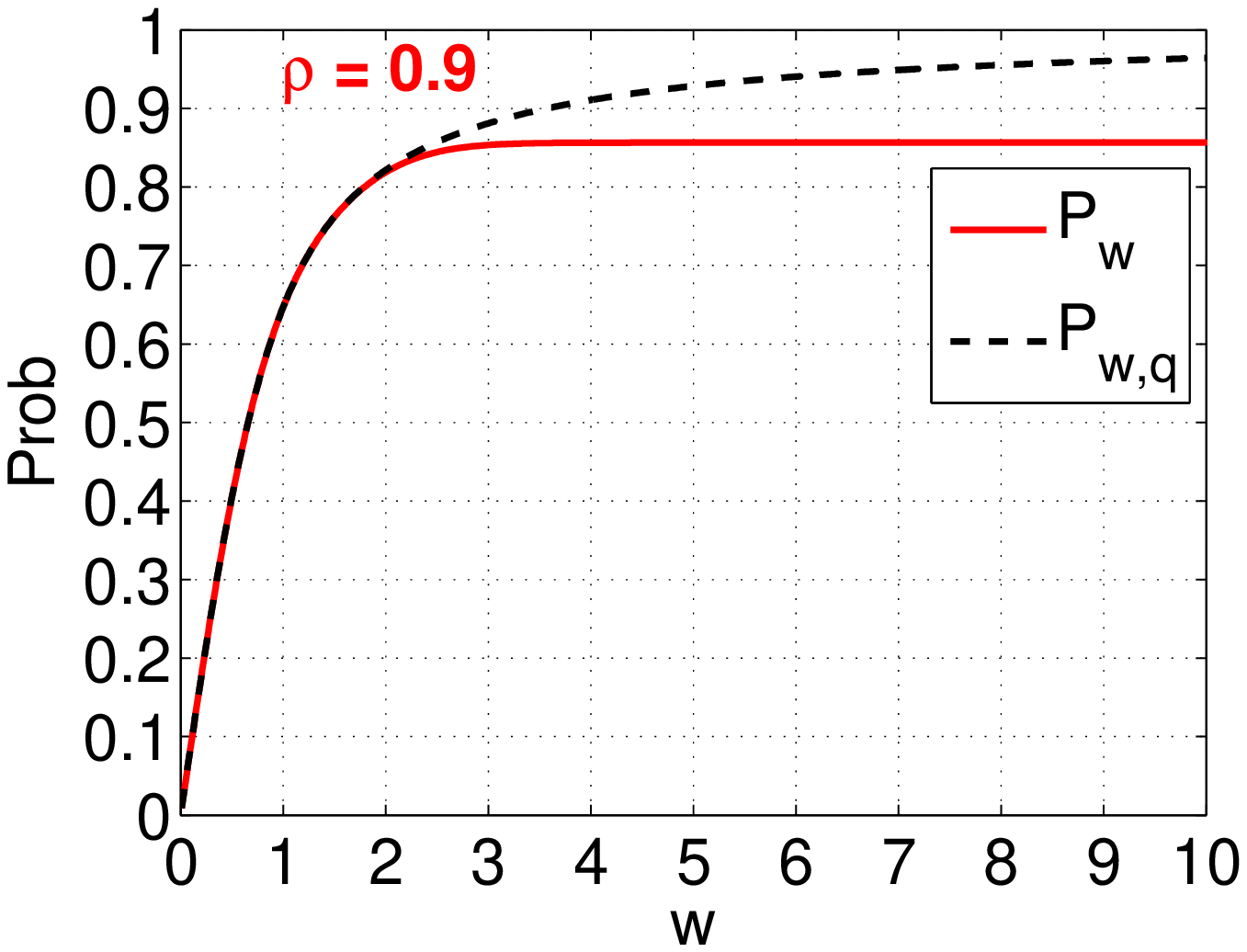}
\includegraphics[width = 2.2in]{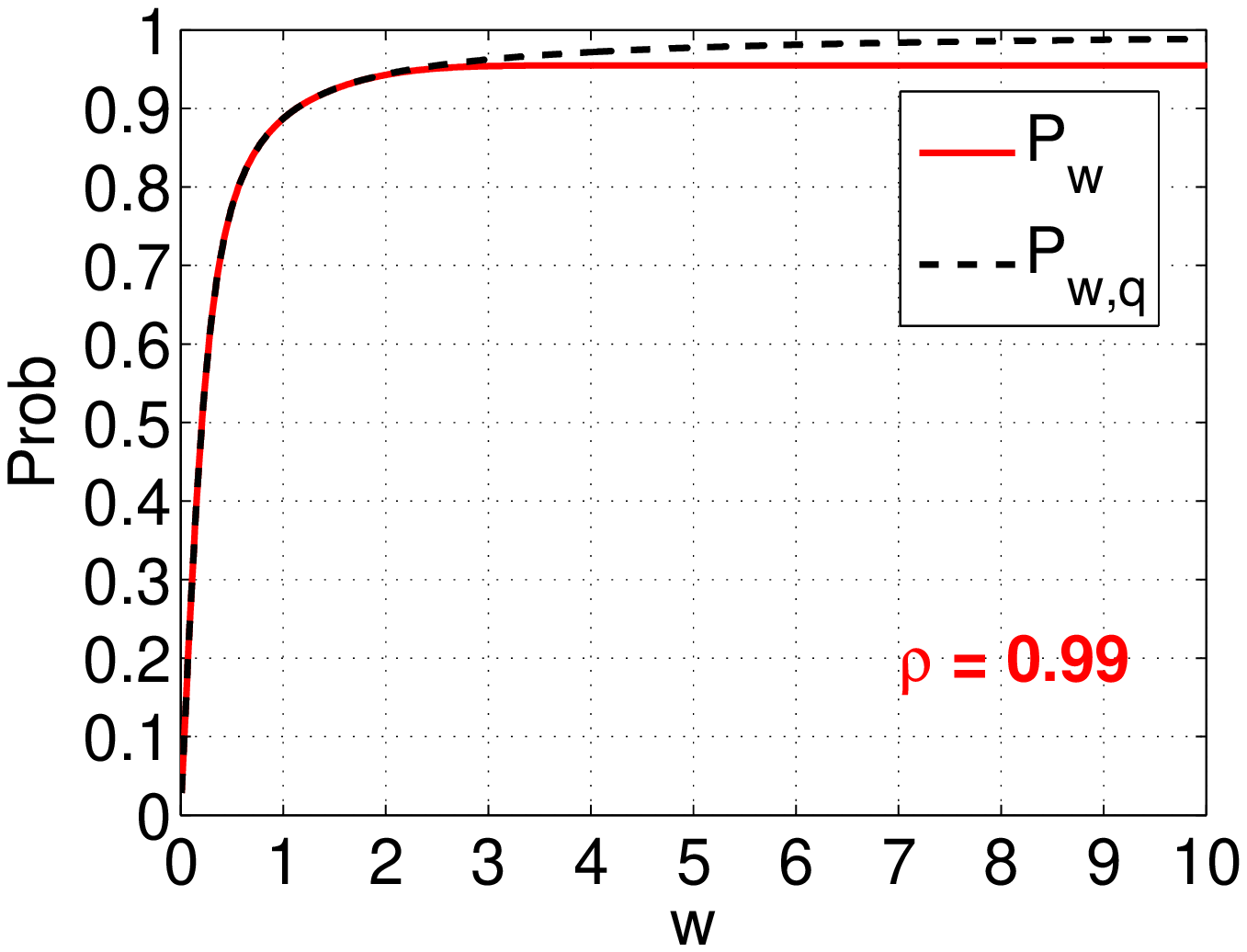}
}
\end{center}
\vspace{-.2in}
\caption{Collision probabilities, $P_w$ and $P_{w,q}$, for  $\rho = 0, 0.25, 0.5, 0.75, 0.9$, and $0.99$.  Our proposed scheme  ($h_w$) has smaller collision probabilities than the existing scheme~\cite{Proc:Datar_SCG04} ($h_{w,q}$), especially when $w>2$.}\label{fig_Pwq}
\end{figure}

Figure~\ref{fig_Pwq} plots both $P_w$ and $P_{w,q}$ for selected $\rho$ values.  The difference between $P_w$ and $P_{w,q}$ becomes apparent after about $w>2$. For example, when $\rho=0$, $P_{w}$ quickly approaches the limit 0.5 while $P_{w,q}$ keeps increasing (to 1) as $w$ increases. Intuitively, the fact that $P_{w,q}\rightarrow1$ when $\rho=0$, is undesirable because it means two orthogonal vectors will have the same coded value. Thus, it is  not surprising that our proposed scheme $h_w$ will have better performance than $h_{w,q}$. We will analyze their theoretical variances to provide precise comparisons.

\section{Analysis of Two Coding Schemes ($h_w$ and $h_{w,q}$) for Similarity Estimation}\label{sec_compare_hwq}

In both schemes (corresponding to $h_w$ and $h_{w,q}$), the collision probabilities $P_w$ and $P_{w,q}$ are monotonically  increasing functions of the similarity $\rho$. Since there is a one-to-one mapping between $\rho$ and $P_w$, we can tabulate $P_w$ for each $\rho$ (for example, at a precision of $10^{-3}$). From $k$ independent projections, we can compute the empirical $\hat{P}_w$ and $\hat{P}_{w,q}$ and find the estimates, denoted by  $\hat{\rho}_w$ and $\hat{\rho}_{w,q}$, respectively, from the tables. In this section, we  compare the estimation variances for these two estimators, to  demonstrate advantage of the proposed coding scheme $h_w$.

\newpage

Theorem~\ref{thm_Vwq} provides the variance of $h_{w}$, for estimating $\rho$ from $k$ random projections.

\begin{theorem}\label{thm_Vwq}
\begin{align}
&Var\left(\hat{\rho}_{w,q}\right) = \frac{V_{w,q}}{k} + O\left(\frac{1}{k^2}\right),\hspace{0.2in}\text{where}\\\label{eqn_Vwq}
&V_{w,q} = d^2/4\left(\frac{w/\sqrt{d}}{\phi\left(w/\sqrt{d}\right)-1/\sqrt{2\pi}}\right)^2P_{w,q}(1-P_{w,q}),\ \ \ \ d = 2(1-\rho)
\end{align}
\noindent\textbf{Proof:}\ \ See Appendix~\ref{app_thm_Vwq}.  \hfill$\Box$\\
\end{theorem}

\begin{figure}[h!]
\begin{center}
\includegraphics[width = 3in]{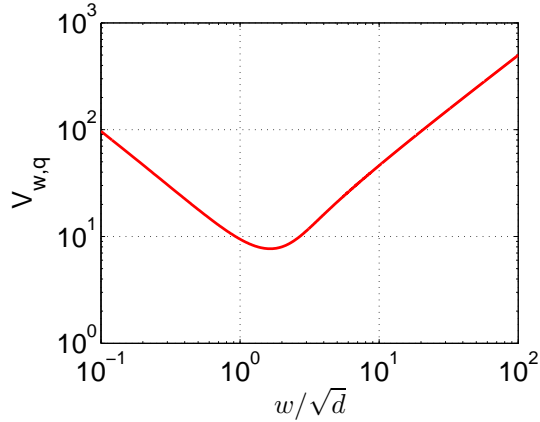}
\end{center}
\vspace{-0.2in}
\caption{The variance factor $V_{w,q}$  (\ref{eqn_Vwq}) without the $\frac{d^2}{4}$ term, i.e., $V_{w,q}\times 4/d^2$.}\label{fig_Vwq}
\end{figure}

Figure~\ref{fig_Vwq} plots the variance factor $V_{w,q}$ defined in (\ref{eqn_Vwq}) without the $\frac{d^2}{4}$ term. (Recall $d = 2(1-\rho)$.)  The minimum is 7.6797 (keeping four digits), attained at $w/\sqrt{d} = 1.6476$. The plot also suggests that the performance of this popular scheme can be sensitive to the choice of the bin width $w$. This is a practical disadvantage. Since we do not know $\rho$ (or $d$) in advance and we must specify $w$ in advance, the  performance  of this scheme might be unsatisfactory, as one can not really find one ``optimum'' $w$ for all pairs in a dataset.

In comparison, our proposed scheme has smaller variance and is not as sensitive to the choice of $w$.

\begin{figure}[h!]
\begin{center}
\includegraphics[width=3in]{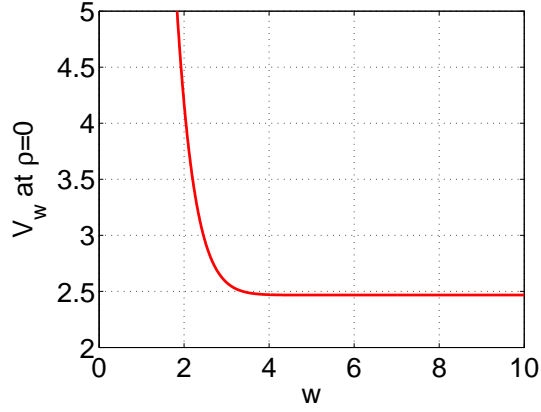}
\end{center}
\vspace{-0.25in}
\caption{The minimum of $\left. V_w\right|_{\rho=0} \rightarrow \pi^2/4$, as $w\rightarrow\infty$.}\label{fig_Vw0}
\end{figure}

\begin{theorem}\label{thm_Vw}
\begin{align}\vspace{-0.15in}
&Var\left(\hat{\rho}_{w}\right) = \frac{V_{w}}{k} + O\left(\frac{1}{k^2}\right),\hspace{0.2in}\text{where}\\\label{eqn_Vw}
&V_{w} = \frac{\pi^2(1-\rho^2)P_w(1-P_w)}{\left[\sum_{i=0}^\infty \left(e^{-\frac{(i+1)^2w^2}{(1+\rho)}}
+e^{-\frac{i^2w^2}{(1+\rho)}}-2e^{-\frac{w^2}{2(1-\rho^2)}}e^{-\frac{i(i+1)w^2}{1+\rho}}\right)\right]^2}
\end{align}
In particular, when $\rho=0$, we have
\begin{align}\label{eqn_Vw0}
\left. V_w\right|_{\rho=0}
=&\left[\frac{\sum_{i=0}^\infty \left(\Phi((i+1)w) - \Phi(iw)\right)^2}{\sum_{i=0}^\infty \left(\phi((i+1)w) - \phi(iw)\right)^2}\right]\left[\frac{1/2-\sum_{i=0}^\infty \left(\Phi((i+1)w) - \Phi(iw)\right)^2}{\sum_{i=0}^\infty \left(\phi((i+1)w) - \phi(iw)\right)^2}\right]
\end{align}
\noindent\textbf{Proof:}\ \ See Appendix~\ref{app_thm_Vw}. \hfill$\Box$
\end{theorem}
\noindent\textbf{Remark:}\ At $\rho= 0$, the minimum is $V_w   = \frac{\pi^2}{4}$ attained at $w\rightarrow \infty$, as shown in Figure~\ref{fig_Vw0}. Note that when $w\rightarrow\infty$,
we have $\sum_{i=0}^\infty \left(\Phi((i+1)w) - \Phi(iw)\right)^2\rightarrow1/4$ and $\sum_{i=0}^\infty \left(\phi((i+1)w) - \phi(iw)\right)^2\rightarrow1/(2\pi)$, and hence
$\left. V_w\right|_{\rho=0}\rightarrow \left[\frac{1/4}{1/(2\pi)}\right]\left[\frac{1/2-1/4}{1/(2\pi)}\right] = \frac{\pi^2}{4}$. In comparison,  Theorem~\ref{thm_Vwq} says that  when $\rho=0$ (i.e., $d=2$) we have  $V_{w,q} = 7.6797$, which is significantly larger than $\pi^2/4=2.4674$. 

\vspace{0.1in}

\begin{figure}[h!]
\begin{center}
\mbox{
\includegraphics[width=2.2in]{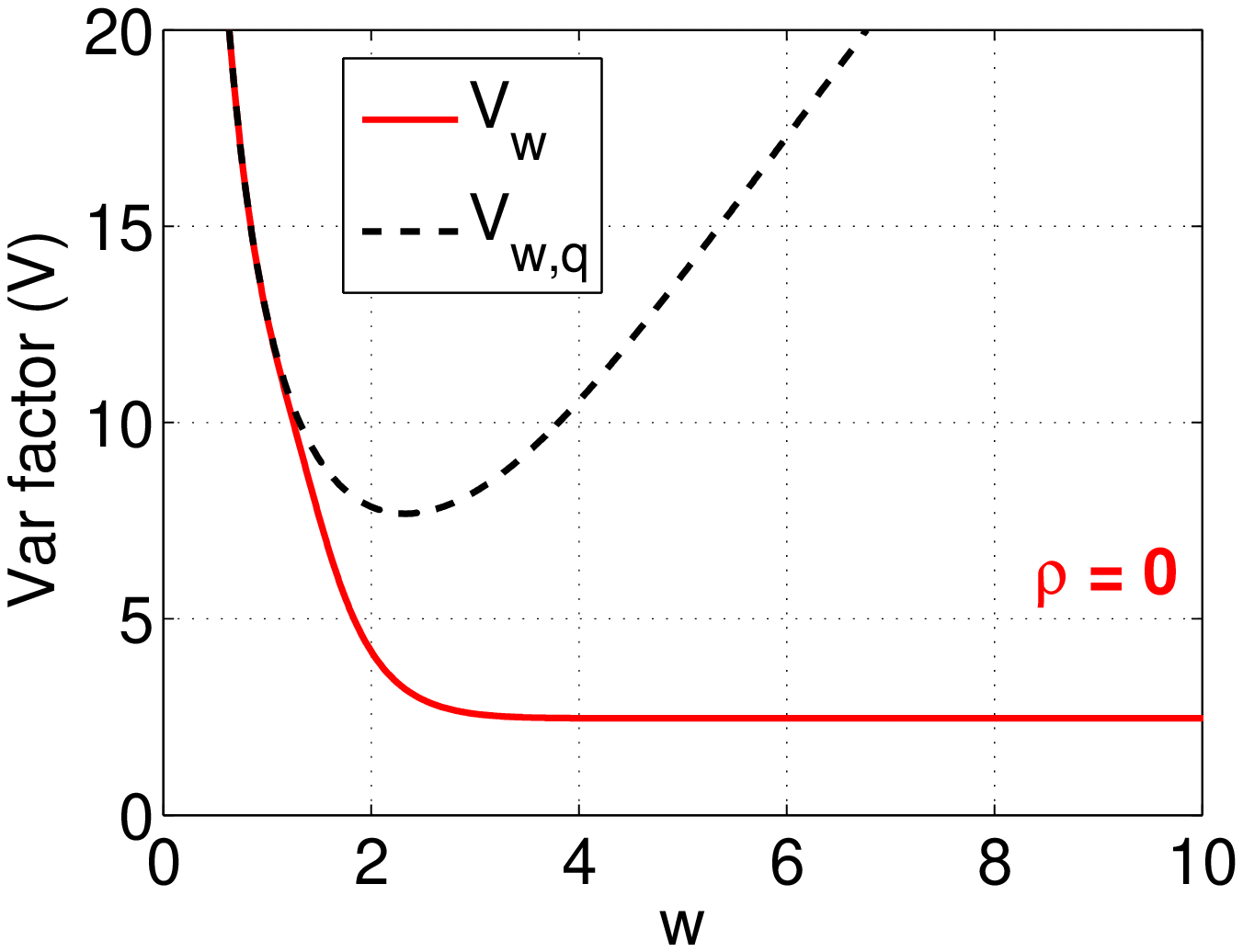}
\includegraphics[width=2.2in]{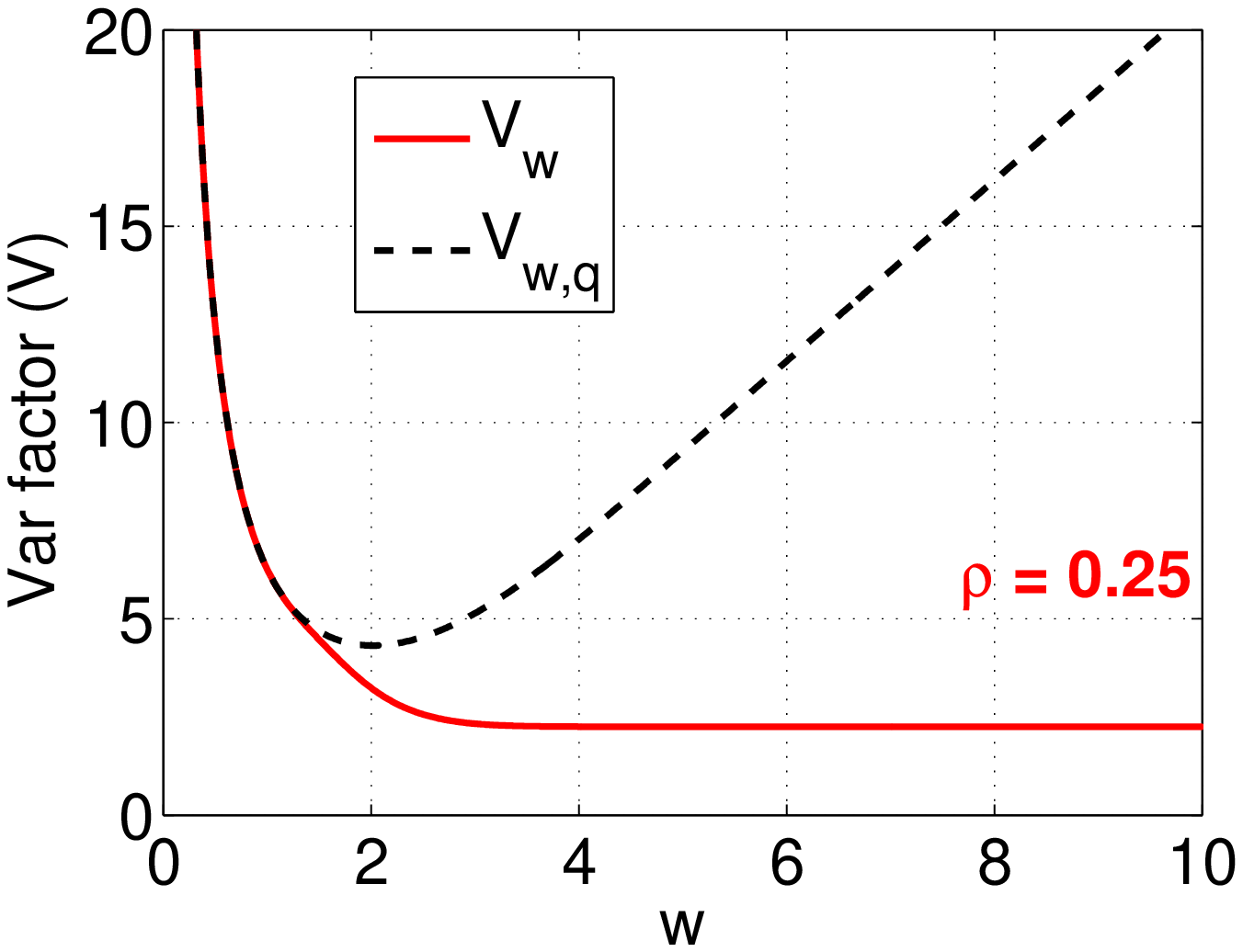}
\includegraphics[width=2.2in]{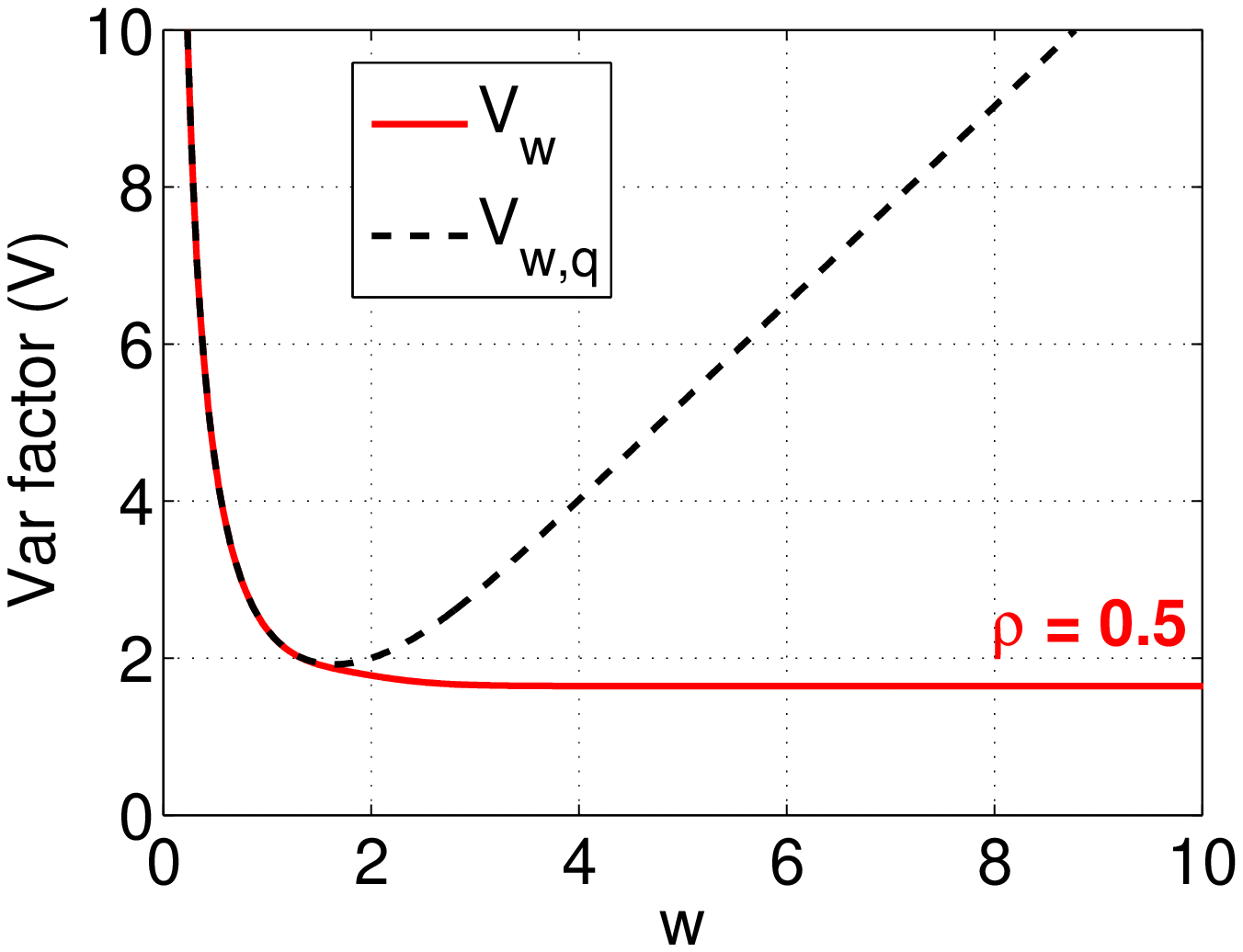}
}
\mbox{
\includegraphics[width=2.2in]{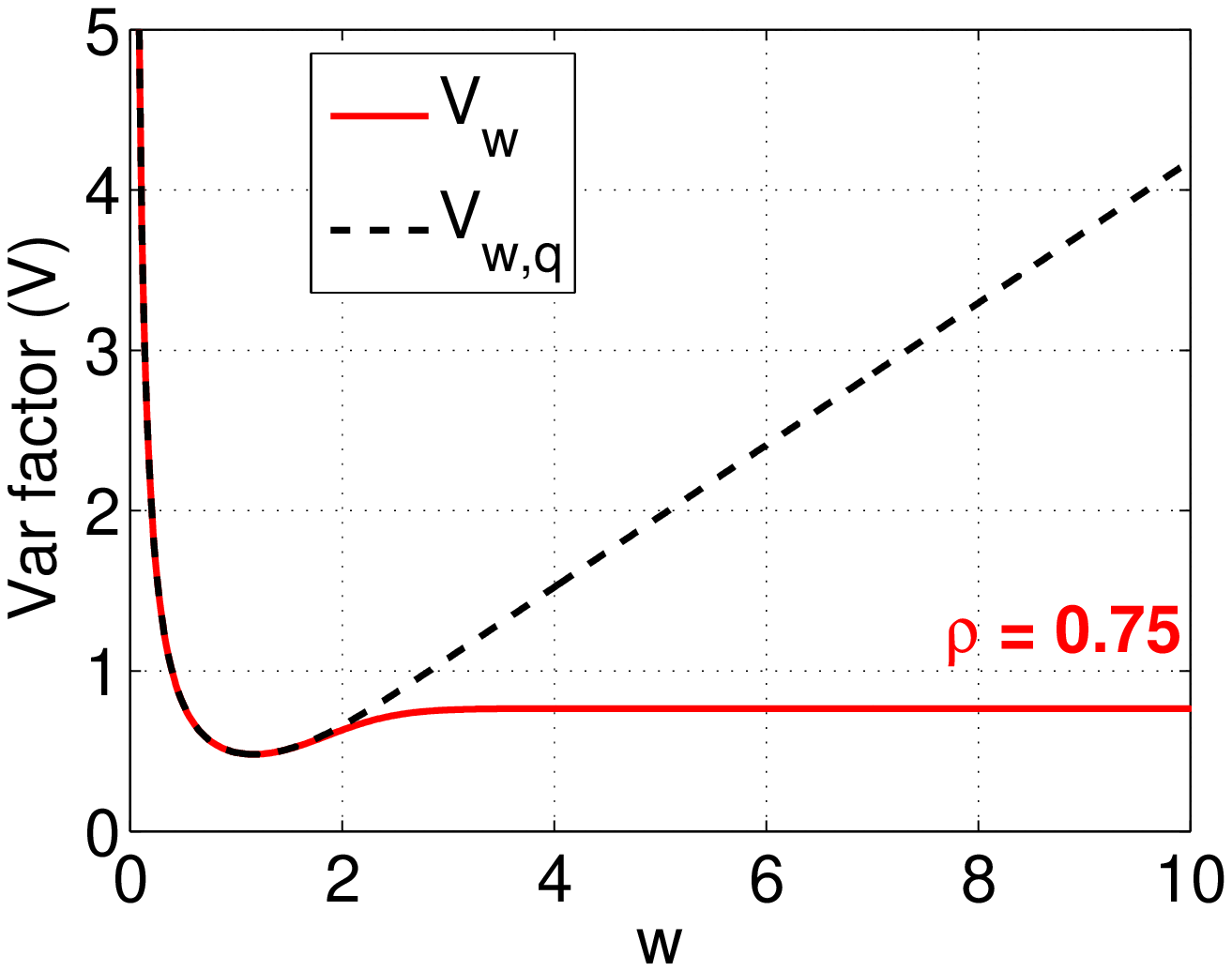}
\includegraphics[width=2.2in]{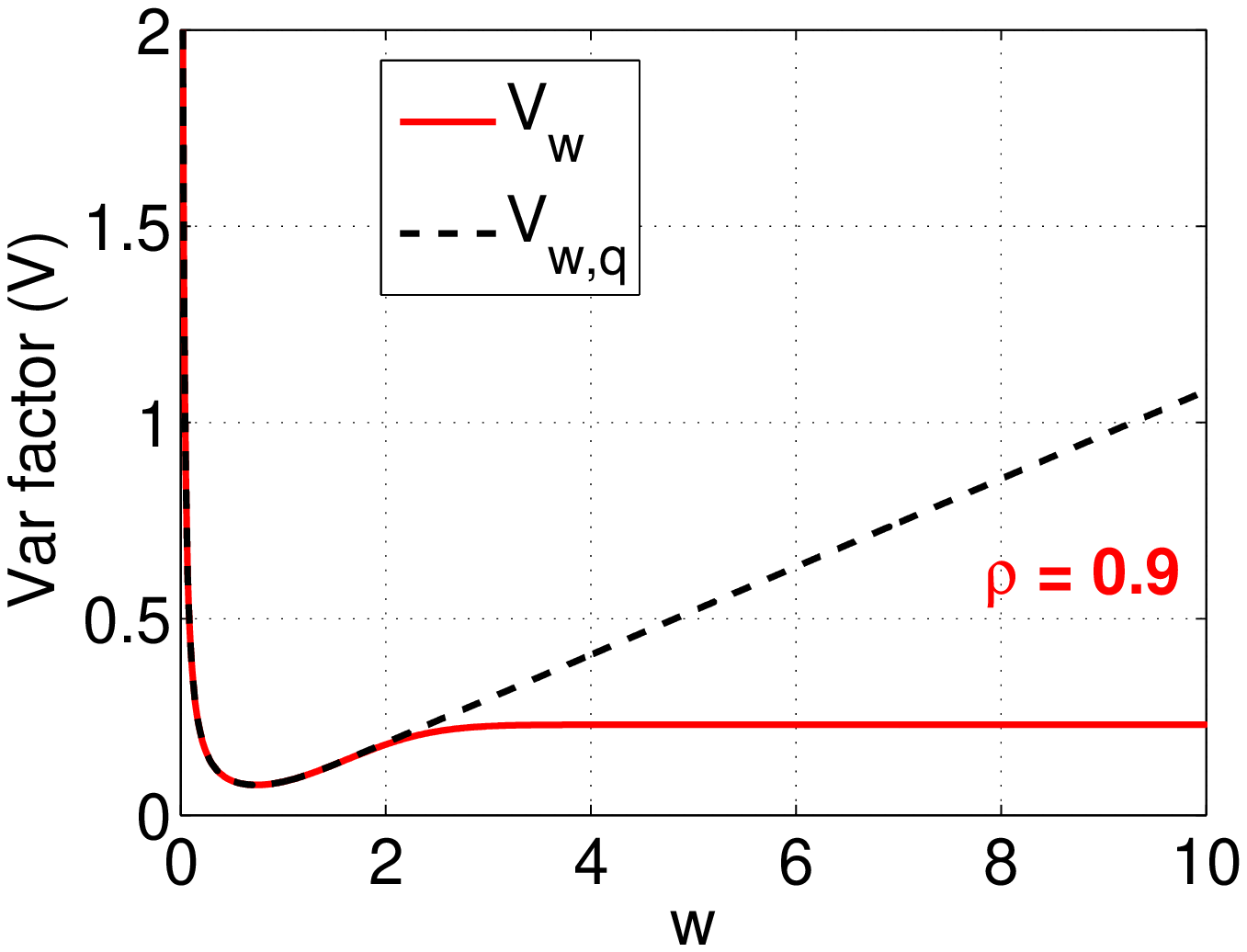}
\includegraphics[width=2.2in]{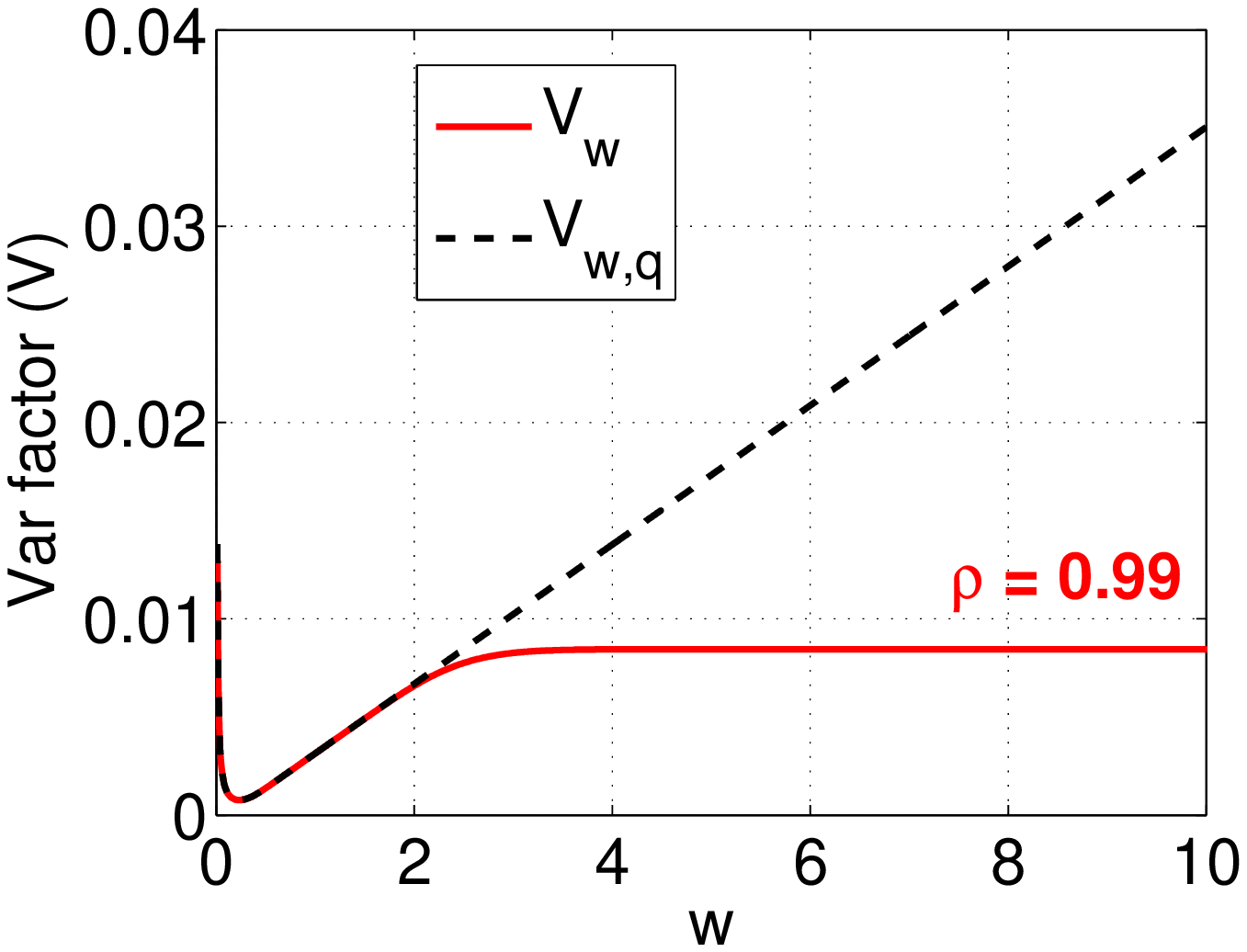}

}

\end{center}
\vspace{-0.22in}
\caption{Comparisons of two coding schemes at fixed bin width $w$, i.e., $V_w$ (\ref{eqn_Vw}) vs $V_{w,q}$ (\ref{eqn_Vwq}). $V_w$ is smaller than $V_{w,q}$ especially when $w>2$ (or even when $w>1$ and $\rho$ is small). For both schemes, at a fixed $\rho$, we can find the optimum $w$ value which minimizes $V_w$ (or $V_{w,q}$). In general, once $w>1\sim2$, $V_w$ is not sensitive to $w$ (unless $\rho$ is very close to 1). This is one significant advantage of the proposed scheme $h_w$.}\label{fig_Vw}
\end{figure}

To compare the variances of the two estimators, $Var\left(\hat{\rho}_{w}\right)$  and $Var\left(\hat{\rho}_{w,q}\right)$, we compare their leading constants, $V_w$ and $V_{w,q}$. Figure~\ref{fig_Vw} plots the $V_w$ and $V_{w,q}$ at selected $\rho$ values, verifying that (i) the variance of the proposed scheme (\ref{eqn_hw}) can be significantly lower than the existing scheme (\ref{eqn_hwq}); and (ii) the performance of the proposed scheme is not as sensitive to the choice of $w$ (e.g., when $w>2$).

\begin{figure}[h!]
\begin{center}

\mbox{
\includegraphics[width=2.5in]{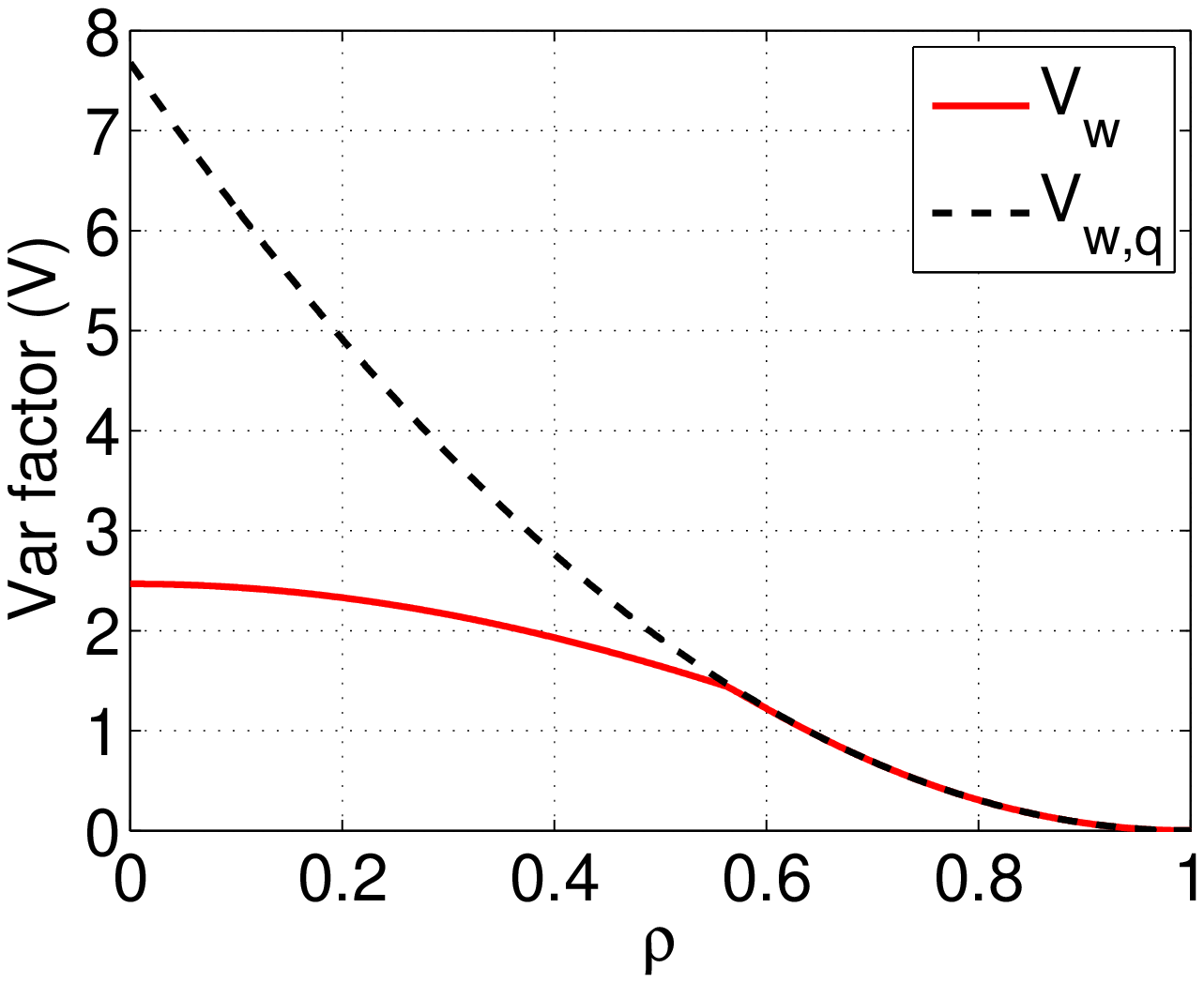}\hspace{0.4in}
\includegraphics[width=2.5in]{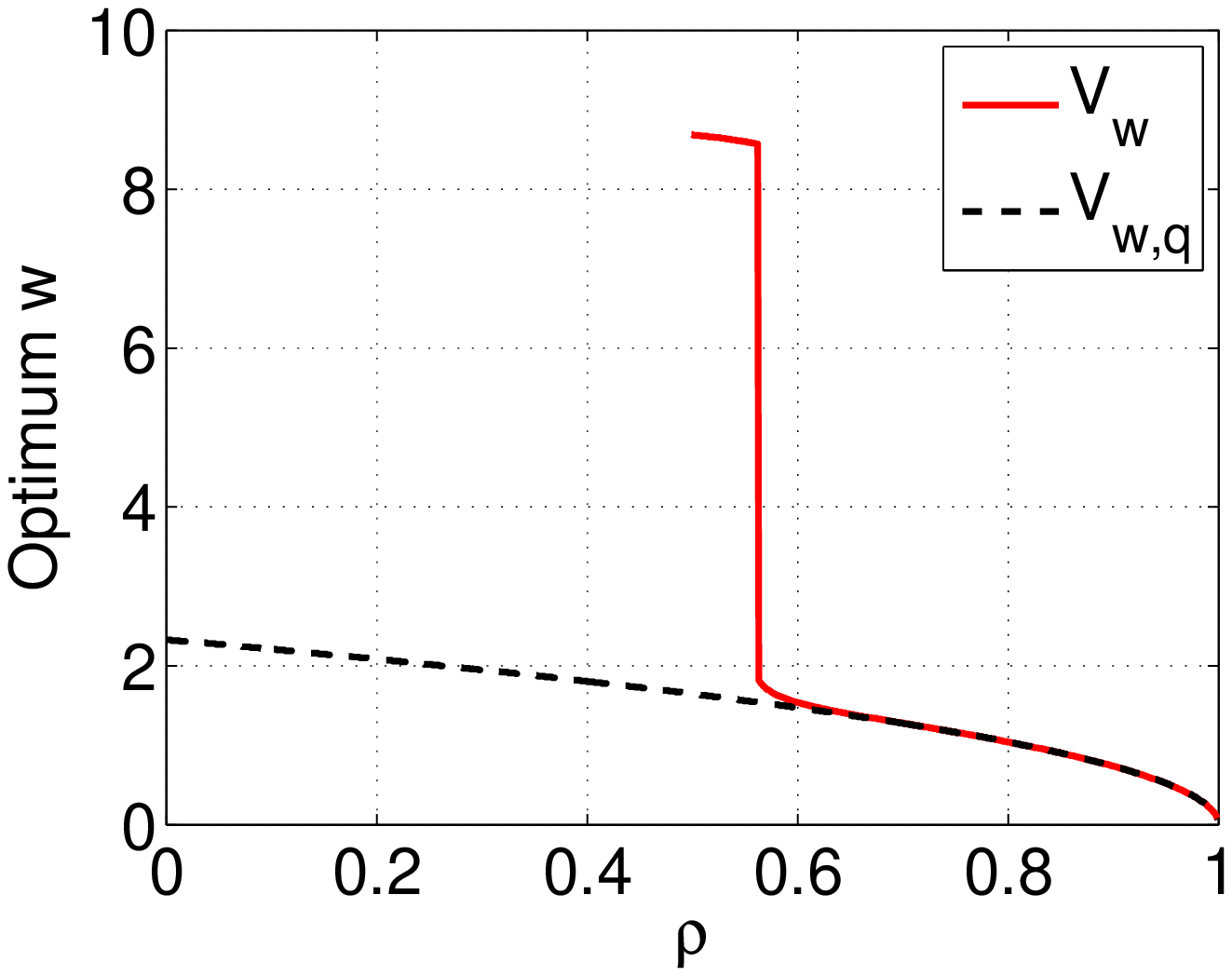}
}
\end{center}
\vspace{-0.2in}
\caption{Comparisons of two coding schemes, $h_w$ and $h_{w,q}$, at optimum bin width $w$.}\label{fig_VwOpt}\vspace{0.2in}
\end{figure}

It is also informative to compare $V_{w}$ and $V_{w,q}$ at their ``optimum'' $w$ values (for fixed $\rho$). Note that  $V_w$ is not  sensitive to $w$ once $w>1\sim2$. The left panel of Figure~\ref{fig_VwOpt} plots the best  values for $V_w$ and $V_{w,q}$, confirming that $V_w$ is significantly lower than $V_{w,q}$ at smaller  $\rho$ values (e.g., $\rho<0.56$).\\

The right panel of Figure~\ref{fig_VwOpt} plots the optimum $w$ values (for fixed $\rho$). Around $\rho = 0.56$, the optimum $w$ for $V_w$ becomes significantly larger than 6 and may not be  reliably evaluated. From the remark for Theorem~\ref{thm_Vw}, we know that  at $\rho=0$ the optimum $w$ grows to $\infty$. Thus, we can conclude that if $\rho<0.56$, it suffices to implement our coding scheme using just 1 bit (i.e., signs of the projected data).  In comparison, for the existing scheme $h_{w,q}$, the optimum $w$ varies much slower. Even at $\rho=0$, the optimum $w$ is around 2. This means $h_{w,q}$ will always need to use more bits than $h_w$, to code the projected data. This is another advantage of our proposed scheme.\\

In practice, we do not know $\rho$ in advance and we often care about data vector pairs of high similarities. When $\rho>0.56$, Figure~\ref{fig_Vw} and Figure~\ref{fig_VwOpt} illustrate that we might want to choose small $w$ values (e.g., $w<1$). However, using a small $w$ value will hurt the performance in pairs of the low similarities. This dilemma motivates us to develop non-uniform coding schemes.

\section{A 2-Bit Non-Uniform Coding Scheme}\label{sec_hw2}

If we quantize the projected data according to the four regions $(-\infty, -w), [-w, 0), [0, w), [w,\infty)$, we  obtain a 2-bit non-uniform scheme. At the risk of abusing  notation, we name this scheme ``$h_{w,2}$'', not to be confused with the name of the existing scheme $h_{w,q}$.

According to Lemma~\ref{lem_Pst}, $h_{w,2}$ is also a valid coding scheme. We can theoretically compute the collision probability, denoted by $P_{w,2}$, which is again a monotonically increasing function of the similarity $\rho$. With $k$ projections, we can estimate $\rho$ from the empirical observation of $P_{w,2}$ and we denote this estimator by $\hat{\rho}_{w,2}$.

Theorem~\ref{thm_hw2} provides the expressions for $P_{w,2}$ and $Var\left(\hat{\rho}_{w,2}\right)$.


\begin{theorem}\label{thm_hw2}
\begin{align}\label{eqn_hw2}
&P_{w,2} = \mathbf{Pr}\left(h_{w,2}^{(j)}(u) = h^{(j)}_{w,2}(v)\right)
 =\left\{1 - \frac{1}{\pi}\cos^{-1}\rho\right\} - 4\int_{0}^{w}\phi(z)\Phi\left(\frac{-w+\rho z}{\sqrt{1-\rho^2}}\right)dz
\end{align}
\begin{align}
Var\left(\hat{\rho}_{w,2}\right) = \frac{V_{w,2}}{k} + O\left(\frac{1}{k^2}\right),\hspace{0.2in}\text{where } \
V_{w,2} = \frac{\pi^2(1-\rho^2)P_{w,2}(1-P_{w,2})}{\left[1-2e^{-\frac{w^2}{2(1-\rho^2)}} + 2e^{-\frac{w^2}{1+\rho}}\right]^2}
\end{align}
\noindent\textbf{Proof}:\ \ See Appendix~\ref{app_thm_hw2}.\hfill$\Box$\\
\end{theorem}

\begin{figure}[h!]
\begin{center}
\mbox{
\includegraphics[width=2.2in]{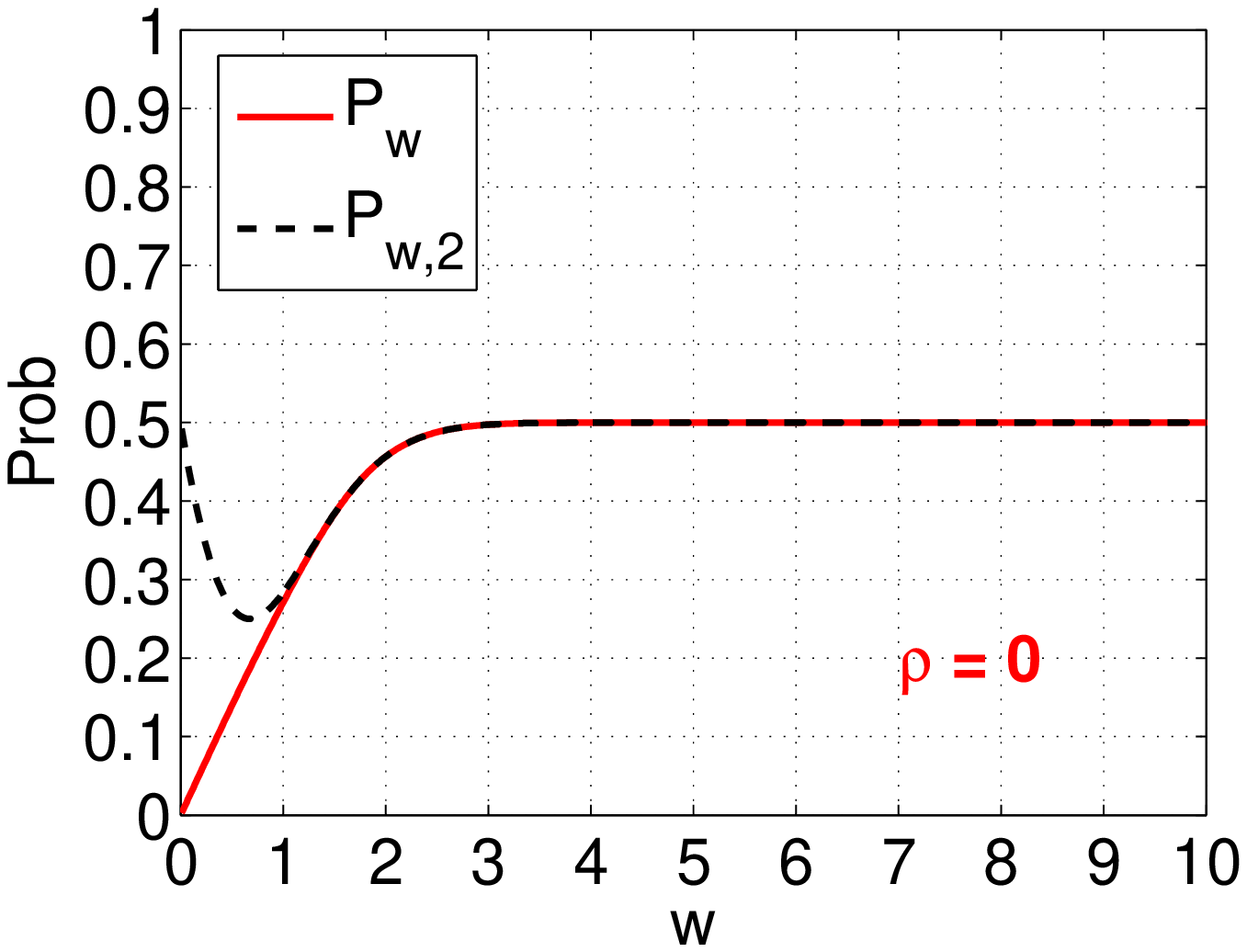}
\includegraphics[width=2.2in]{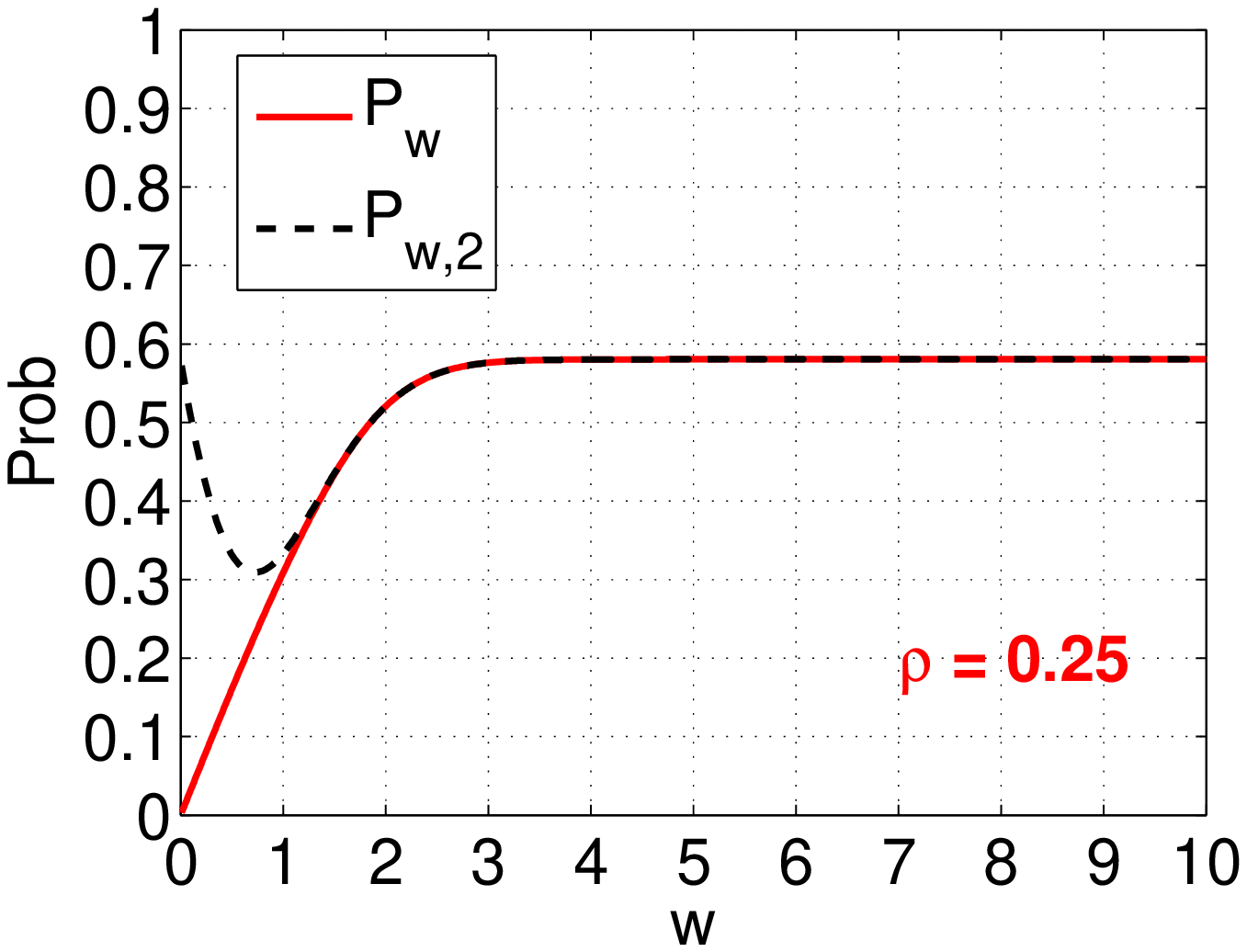}
\includegraphics[width=2.2in]{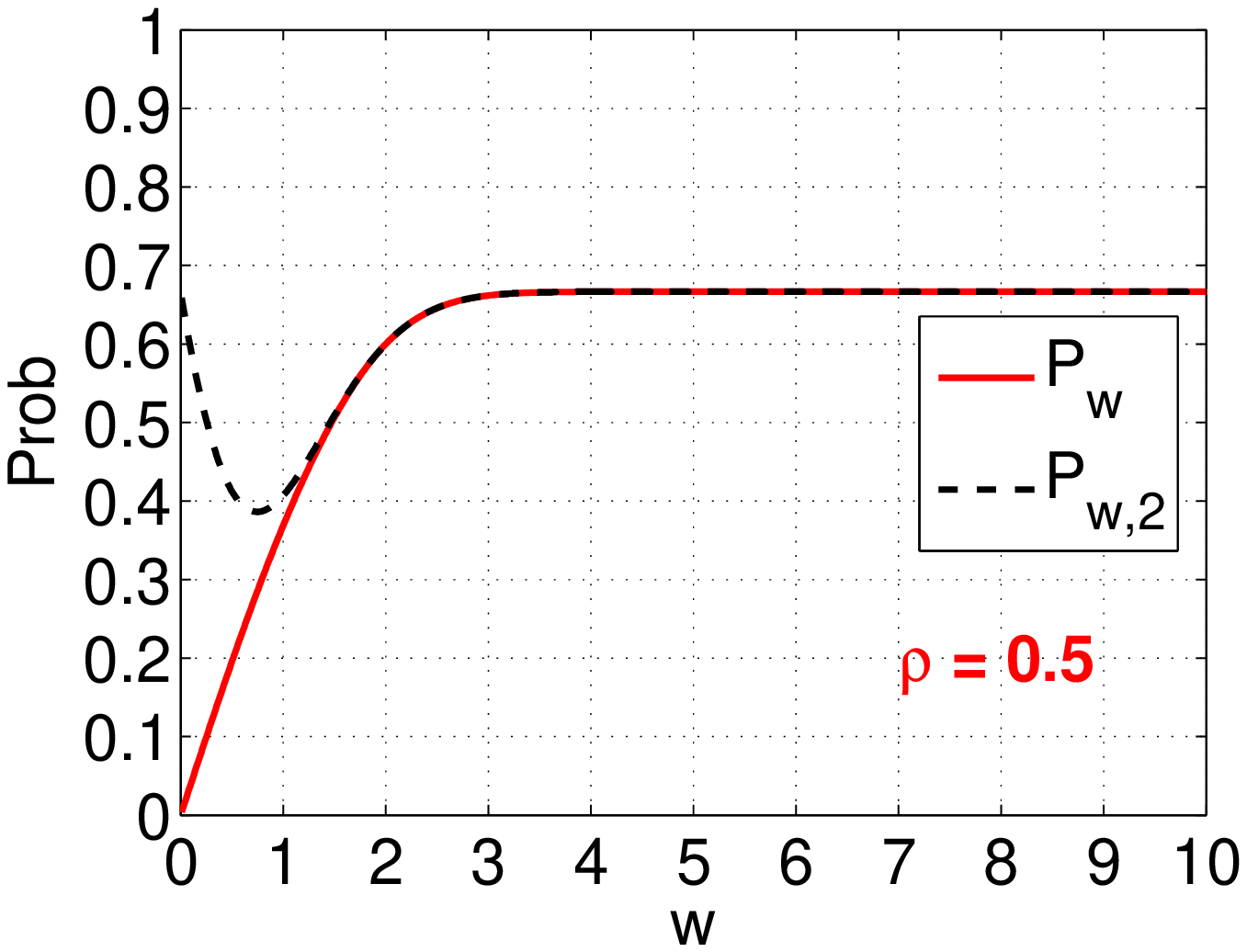}
}
\mbox{
\includegraphics[width=2.2in]{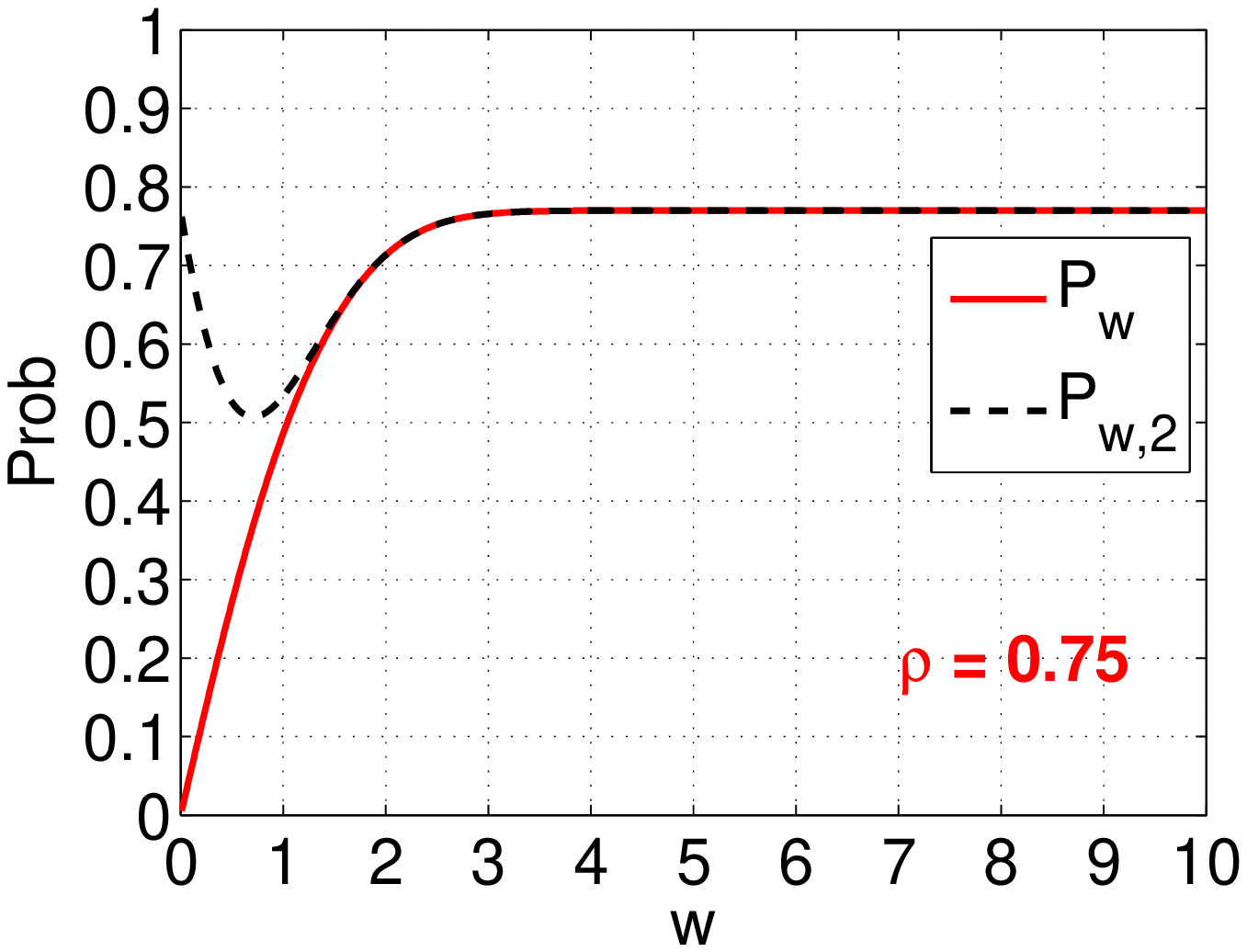}
\includegraphics[width=2.2in]{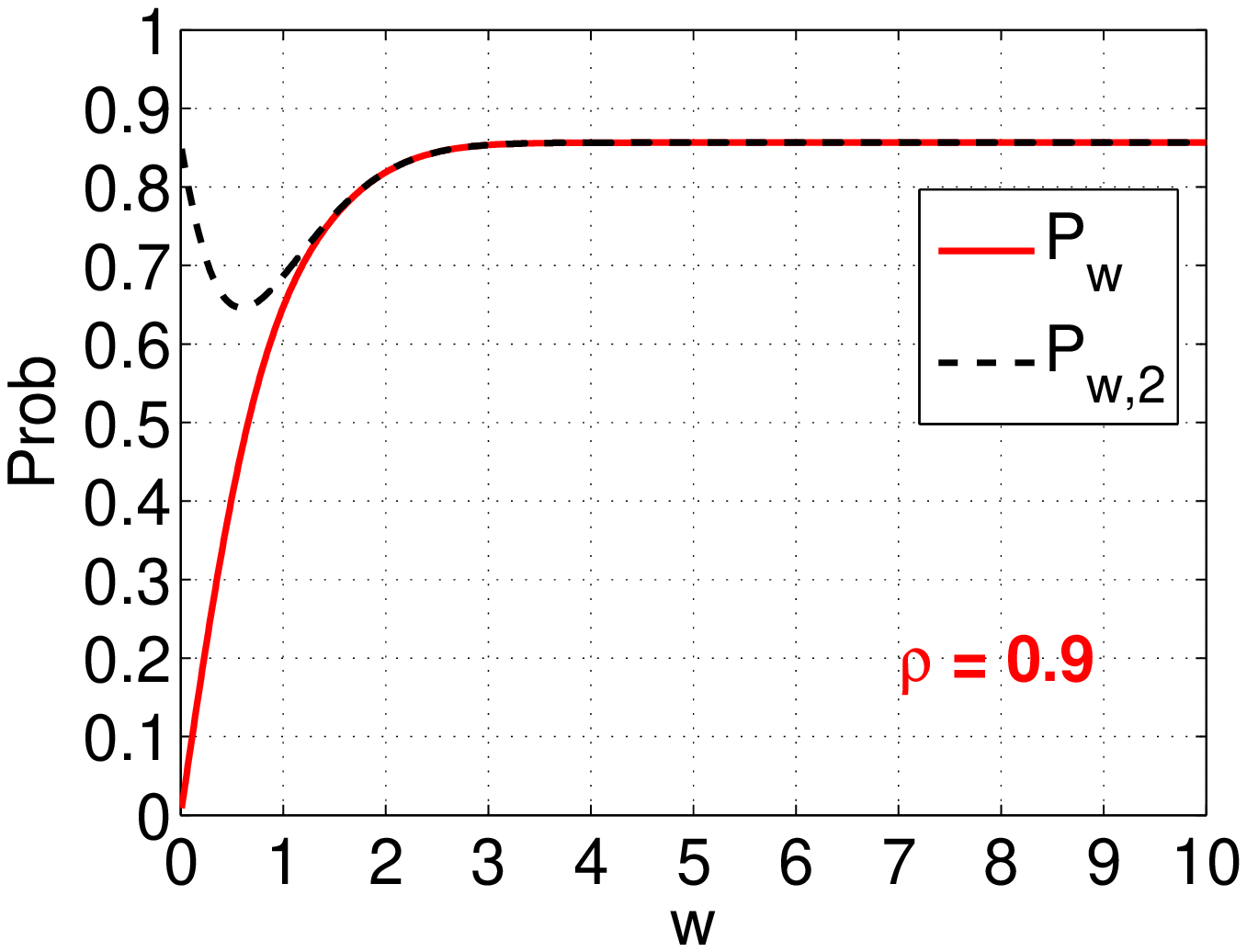}
\includegraphics[width=2.2in]{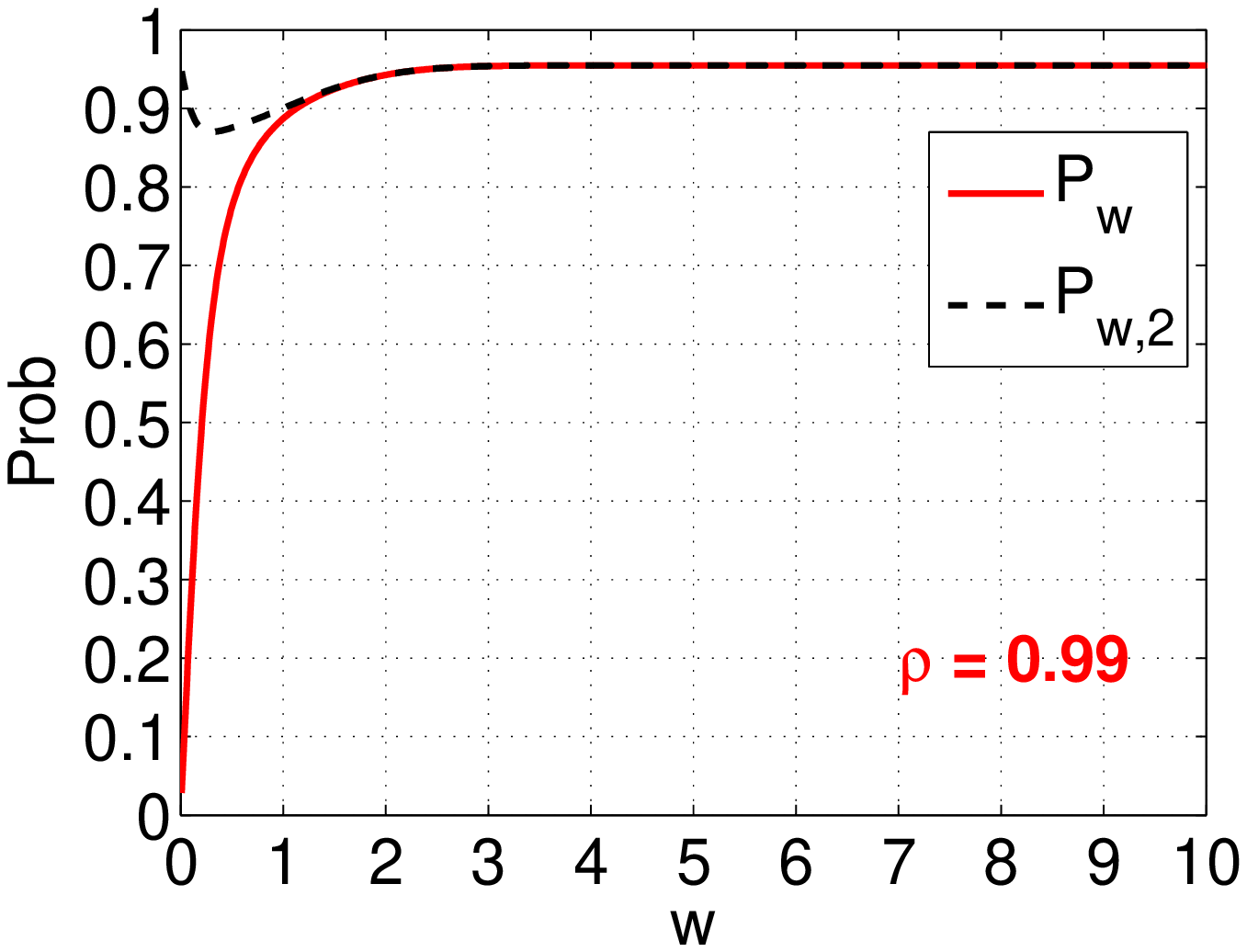}

}

\end{center}
\vspace{-0.2in}
\caption{Collision probabilities for the two proposed coding scheme $h_w$ and $h_{w,2}$. Note that, while $P_{w,2}$ is a monotonically increasing function in $\rho$, it is no longer monotone in $w$. }\label{fig_Pw2}
\end{figure}

Figure~\ref{fig_Pw2} plots $P_{w,2}$ (together with $P_w$) for selected $\rho$ values. When $w>1$, $P_{w,2}$ and $P_{w}$ largely overlap. For small $w$, the two probabilities behave very differently, as expected.  Note $P_{w,2}$ has the same value at $w=0$ and $w=\infty$, and in fact, when $w=0$ or $w=\infty$, we just need one bit (i.e., the signs).
Note that $P_{w,2}$ and $P_{w}$ differ significantly at small $w$. Will this be beneficial? The answer again depends on $\rho$.

\begin{figure}[h!]
\begin{center}

\mbox{
\includegraphics[width=2.2in]{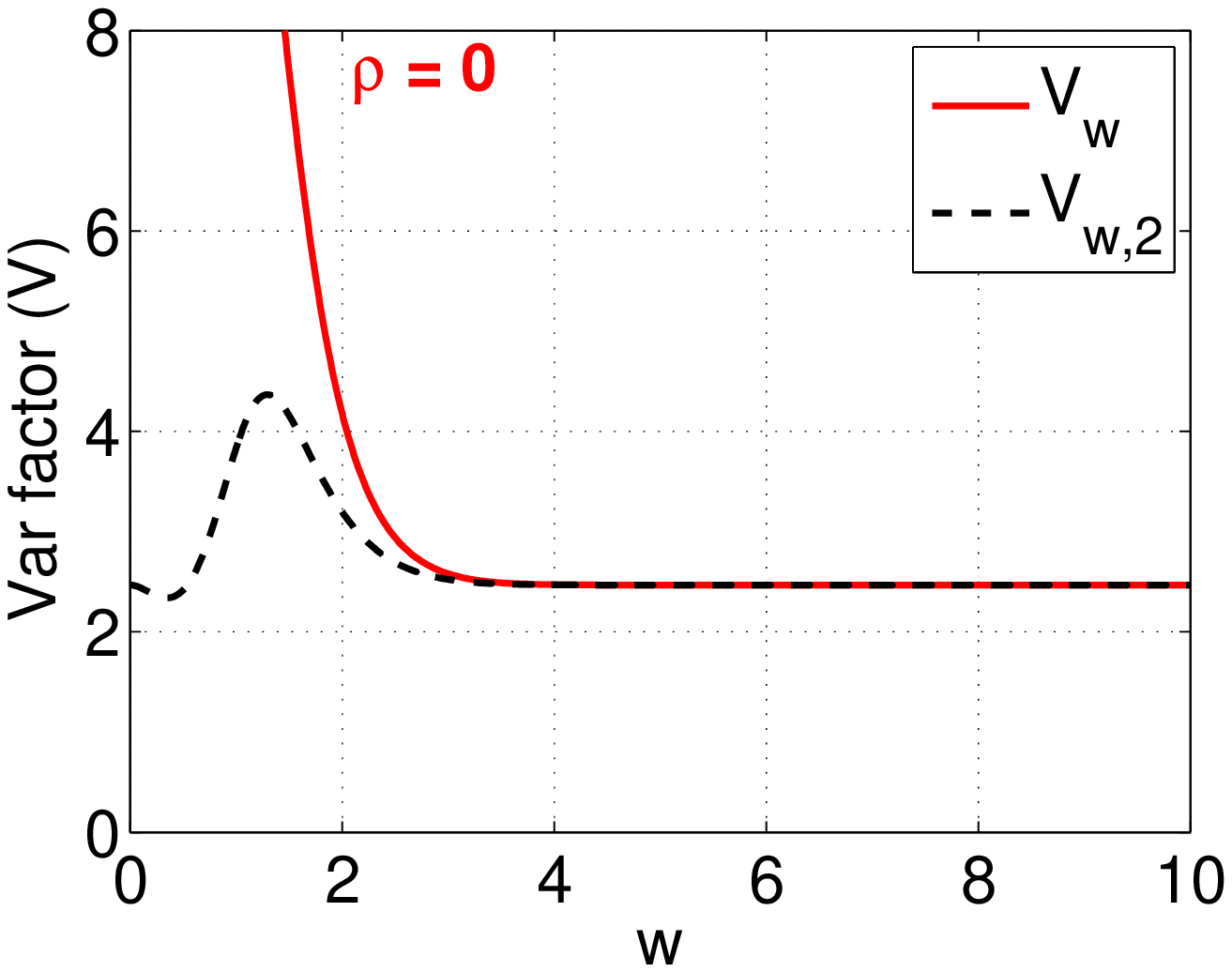}
\includegraphics[width=2.2in]{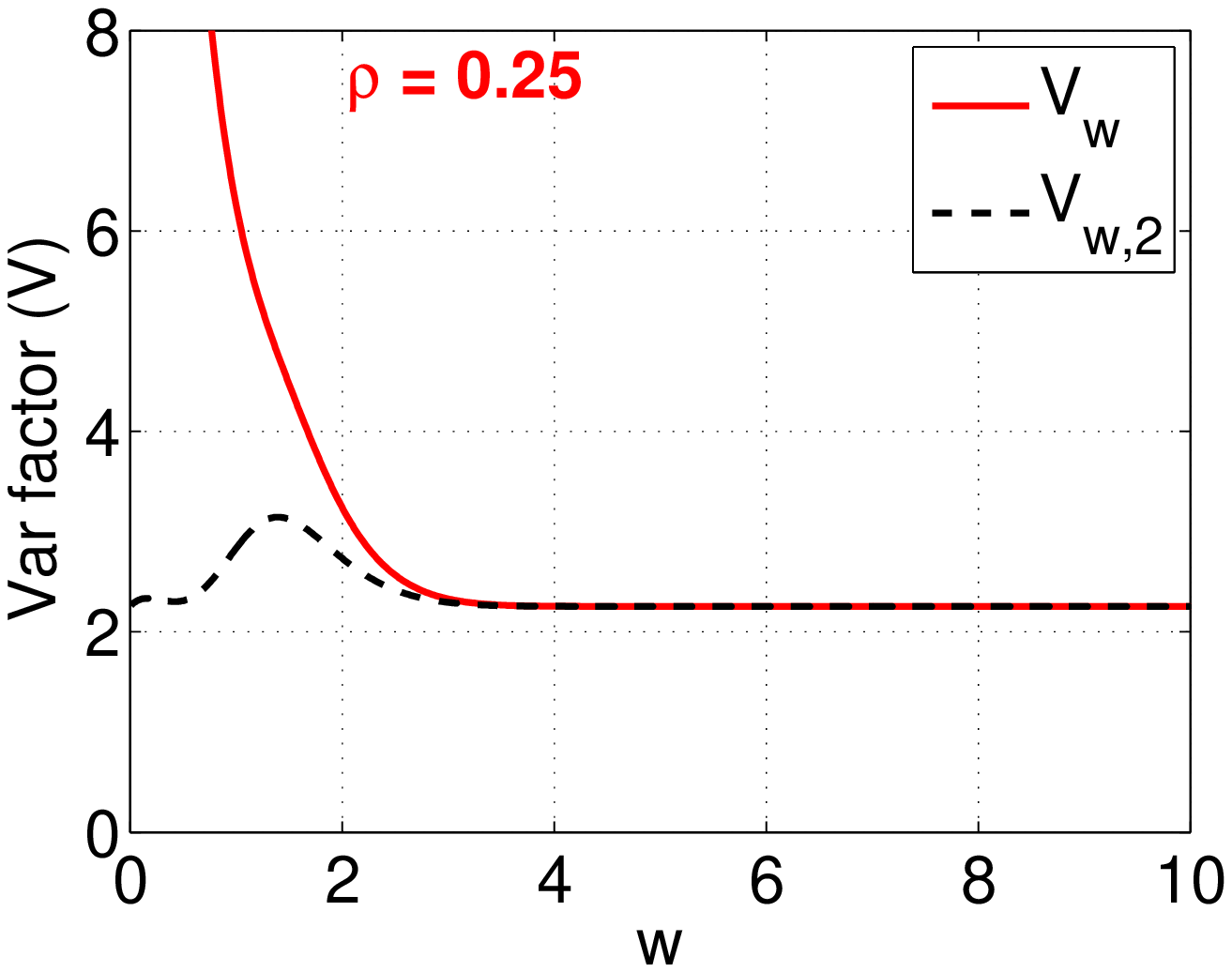}
\includegraphics[width=2.2in]{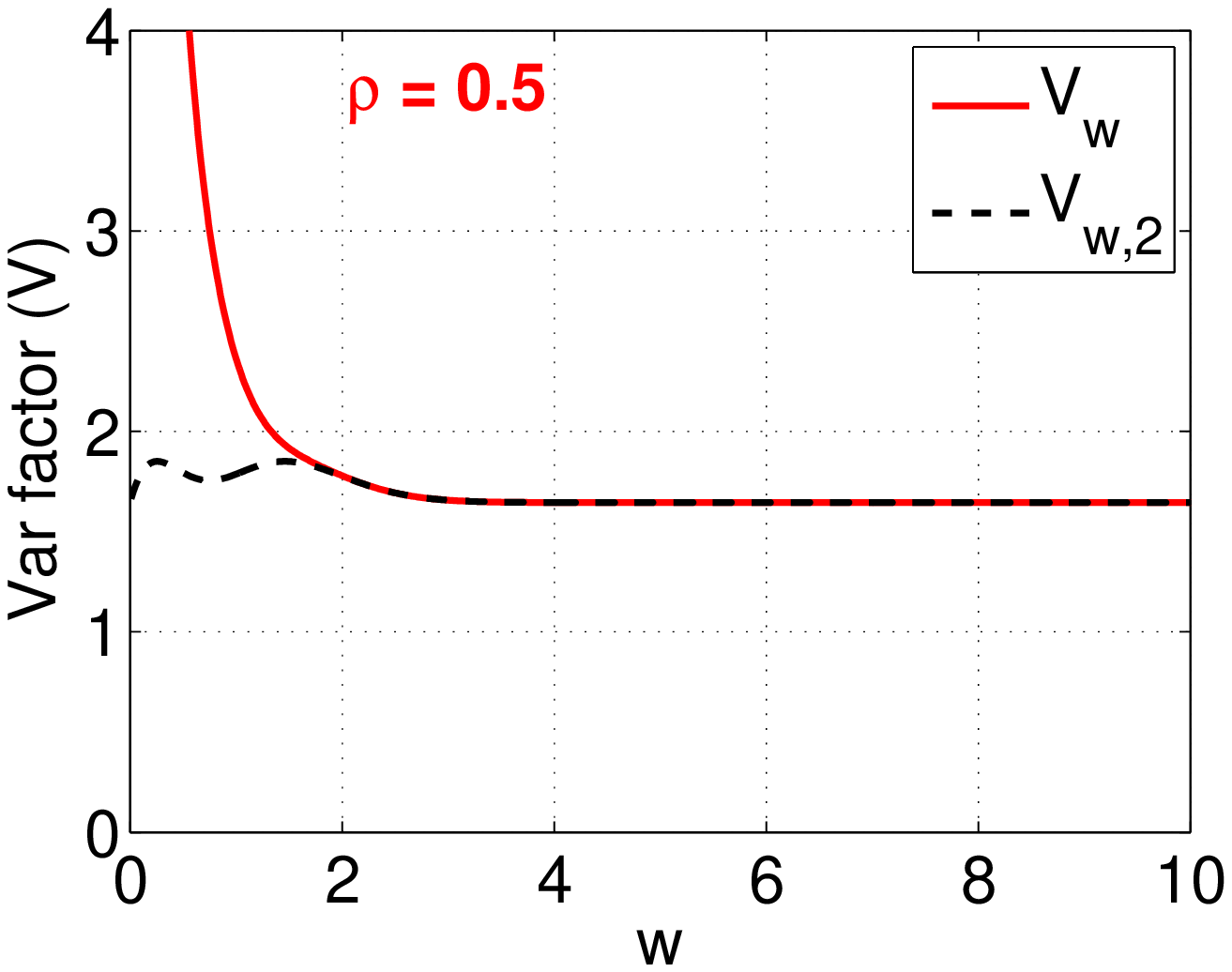}
}
\mbox{
\includegraphics[width=2.2in]{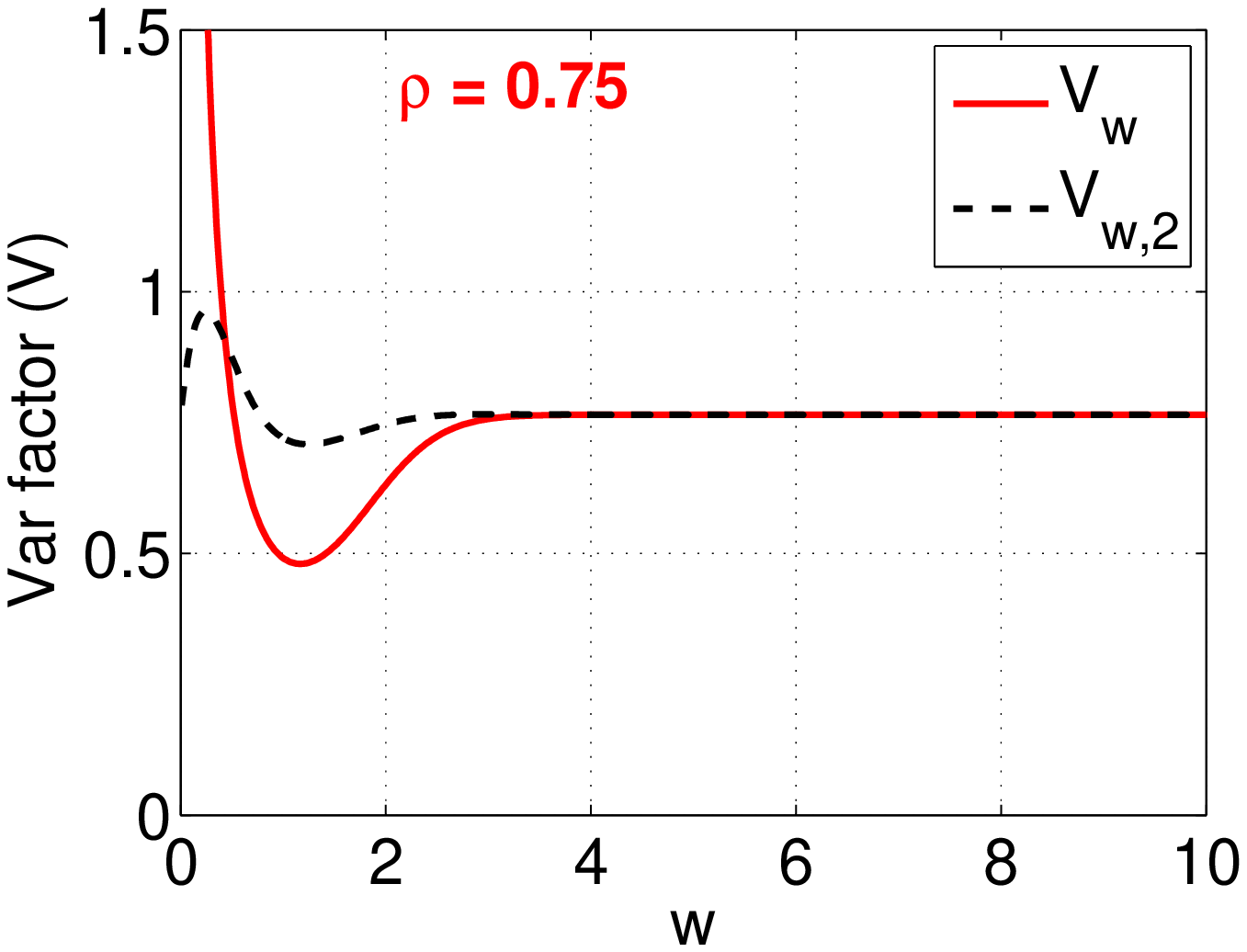}
\includegraphics[width=2.2in]{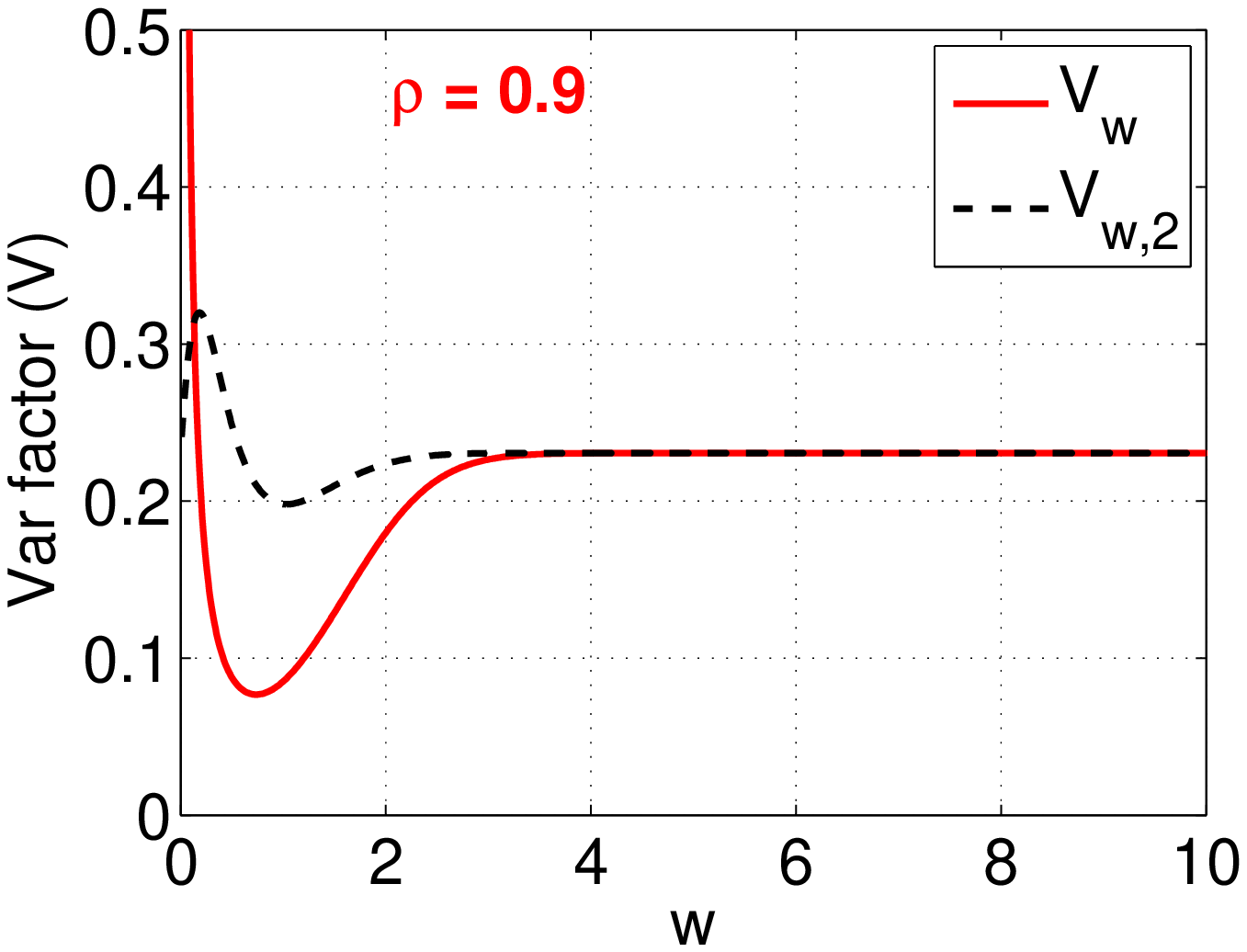}\hspace{-0.1in}
\includegraphics[width=2.2in]{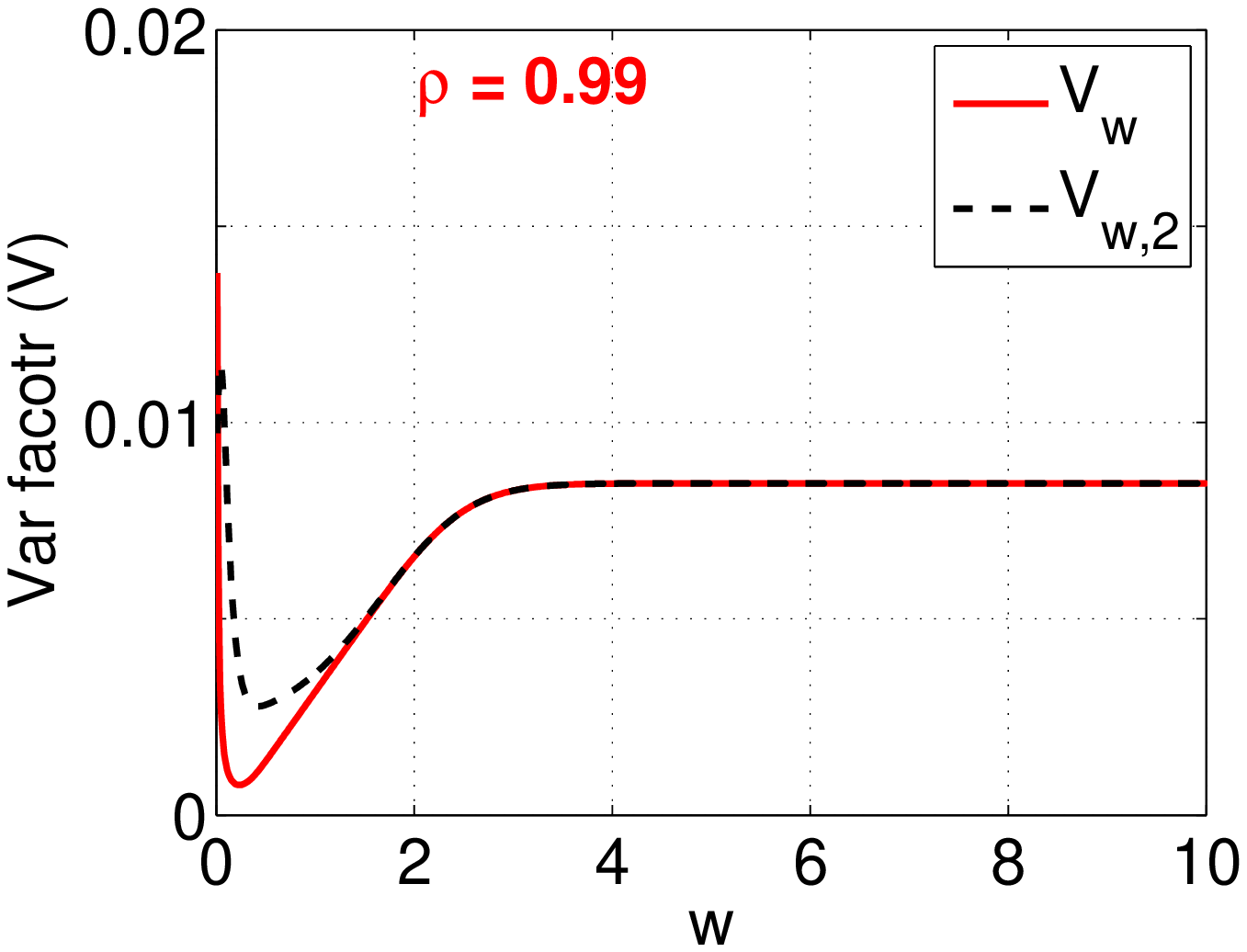}
}
\end{center}
\vspace{-0.2in}
\caption{Comparisons of the estimation variances of  two proposed schemes: $h_w$ and $h_{w,2}$ in terms of $V_w$ and $V_{w,2}$. When $\rho\leq 0.5$, $V_{w,2}$ is significantly lower than $V_{w}$ at small $w$. However, when $\rho$ is high, $V_{w,2}$ will be somewhat higher than $V_w$.  Note that $V_{w,2}$ is not as sensitive to $w$, unlike $V_w$. }\label{fig_Vw2}\vspace{0.2in}
\end{figure}

Figure~\ref{fig_Vw2} plots both $V_{w,2}$ and $V_w$ at selected $\rho$ values, to compare their variances. For $\rho\leq 0.5$, the variance of the estimator using the 2-bit scheme $h_{w,2}$ is significantly lower than that of $h_w$. However, when $\rho$ is high, $V_{w,2}$ might be somewhat higher than $V_{w}$. This means that, in general, we expect the performance of $h_{w,2}$ will be  similar to $h_w$. When applications mainly care about highly similar data pairs, we expect $h_w$ will have (slightly) better performance (at the cost of  more bits).

\vspace{0.2in}

\begin{figure}[h!]
\begin{center}

\mbox{
\includegraphics[width=2.5in]{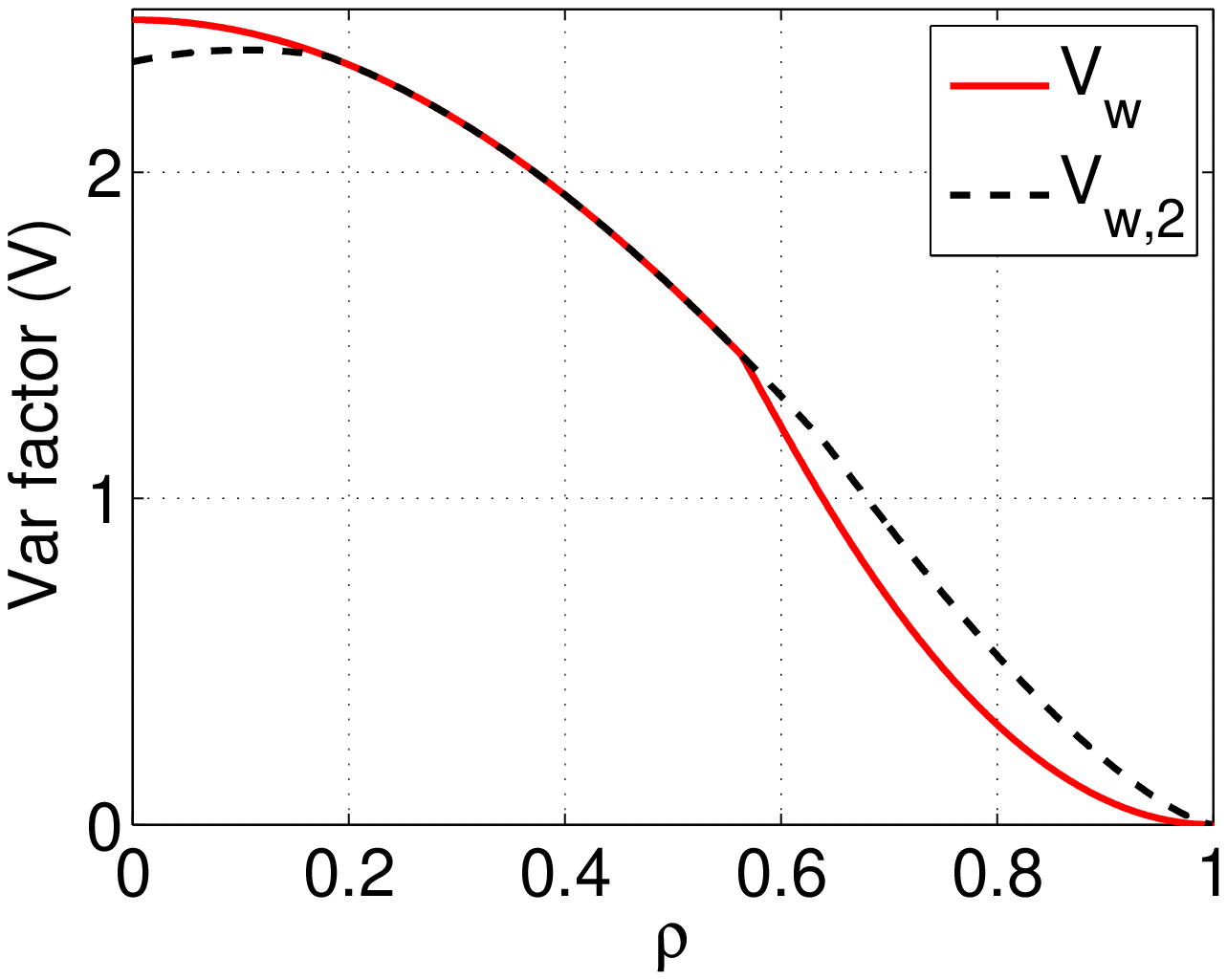}\hspace{0.4in}
\includegraphics[width=2.5in]{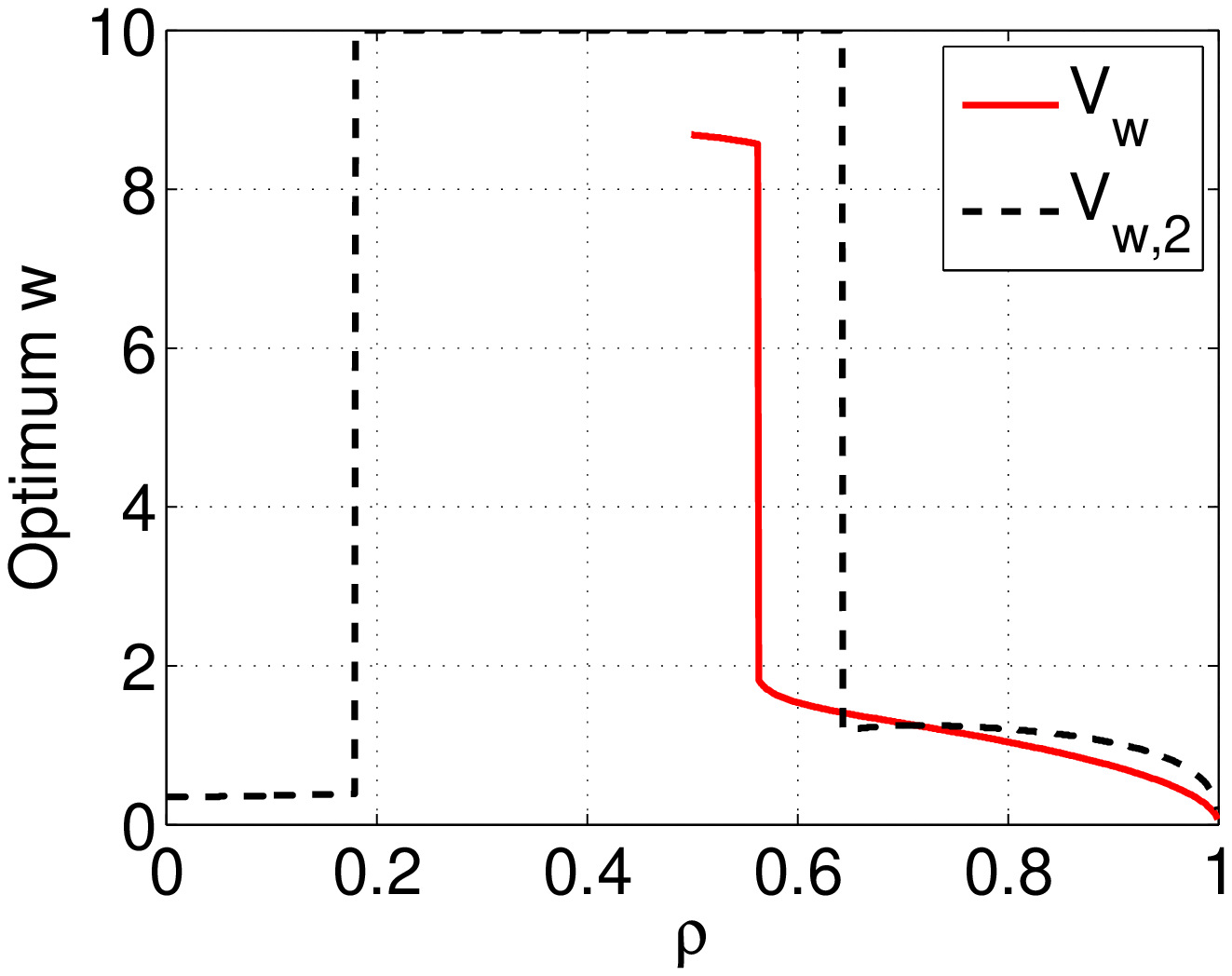}
}

\end{center}
\vspace{-0.2in}
\caption{Left panel: the smallest $V_{w,2}$ (or $V_{w}$) values. Right panel: the optimum $w$ values at which the smallest $V_{w,2}$  (or $V_{w}$) is attained at a fixed $\rho$. }\label{fig_Vw2Opt}
\end{figure}

Finally, Figure~\ref{fig_Vw2Opt}  presents the smallest  $V_{w,2}$ values and the optimum $w$ values at which the smallest $V_{w,2}$ are attained. This plot verifies that $h_w$ and $h_{w,2}$ should perform very similarly, although $h_{w}$ will have better performance at high $\rho$. Also, for a wide range, e.g., $\rho\in[0.2\ 0.62]$, it is preferable to implement $h_{w,2}$ using just 1 bit because the optimum $w$ values are large.

\newpage

\section{The 1-Bit Scheme $h_1$ and Comparisons with $h_{w,2}$ and $h_{w}$}\label{sec_h1}

When $w>6$, it is sufficient to implement $h_w$ or $h_{w,2}$ using just one bit, because the normal probability density decays very rapidly: $1-\Phi(6) = 9.9\times10^{-10}$. Note that we need to consider a very small tail probability because there are many data pairs in a large dataset, not just one pair. With the 1-bit scheme, we simply code the projected data by recording their signs. We denote this scheme by $h_1$, and the corresponding collision probability by $P_1$, and the corresponding estimator by $\hat{\rho}_1$.\\

From Theorem~\ref{thm_hw2}, by setting $w=0$ (or equivalently $w=\infty$), we can directly infer
\begin{align}\label{eqn_h1}
&P_{1} = \mathbf{Pr}\left(h_{1}^{(j)}(u) = h^{(j)}_{1}(v)\right)
 =1 - \frac{1}{\pi}\cos^{-1}\rho
\end{align}
\begin{align}
Var\left(\hat{\rho}_{1}\right) = \frac{V_{1}}{k} + O\left(\frac{1}{k^2}\right),\hspace{0.2in}\text{where } \
V_{1} = \pi^2(1-\rho^2)P_{1}(1-P_{1})
\end{align}

This collision probability is  widely known~\cite{Article:Goemans}. The work of~\cite{Proc:Charikar}  also popularized the use 1-bit coding. The variance was analyzed and compared with a maximum  likelihood estimator in~\cite{Proc:Li_Hastie_Church_COLT06}.\\

\begin{figure}[h!]
\begin{center}

\mbox{
\includegraphics[width=3in]{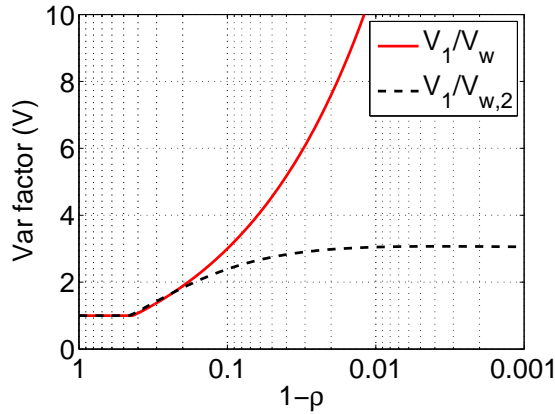}
}

\end{center}
\vspace{-0.2in}
\caption{Variance ratios: $\frac{Var\left(\hat{\rho}_1\right)}{Var\left(\hat{\rho}_{w}\right)}$ and $\frac{Var\left(\hat{\rho}_1\right)}{Var\left(\hat{\rho}_{w,2}\right)}$, to illustrate the reduction of estimation accuracies if the 1-bit coding scheme is used. Here, we plot the maximum ratios (over all choices of bin width $w$) at each $\rho$. To better visualize the high similarity region, we plot $1-\rho$ in log-scale. }\label{fig_VarRatioV1VwVw2}
\end{figure}

\begin{figure}[h!]
\begin{center}

\mbox{
\includegraphics[width=2.7in]{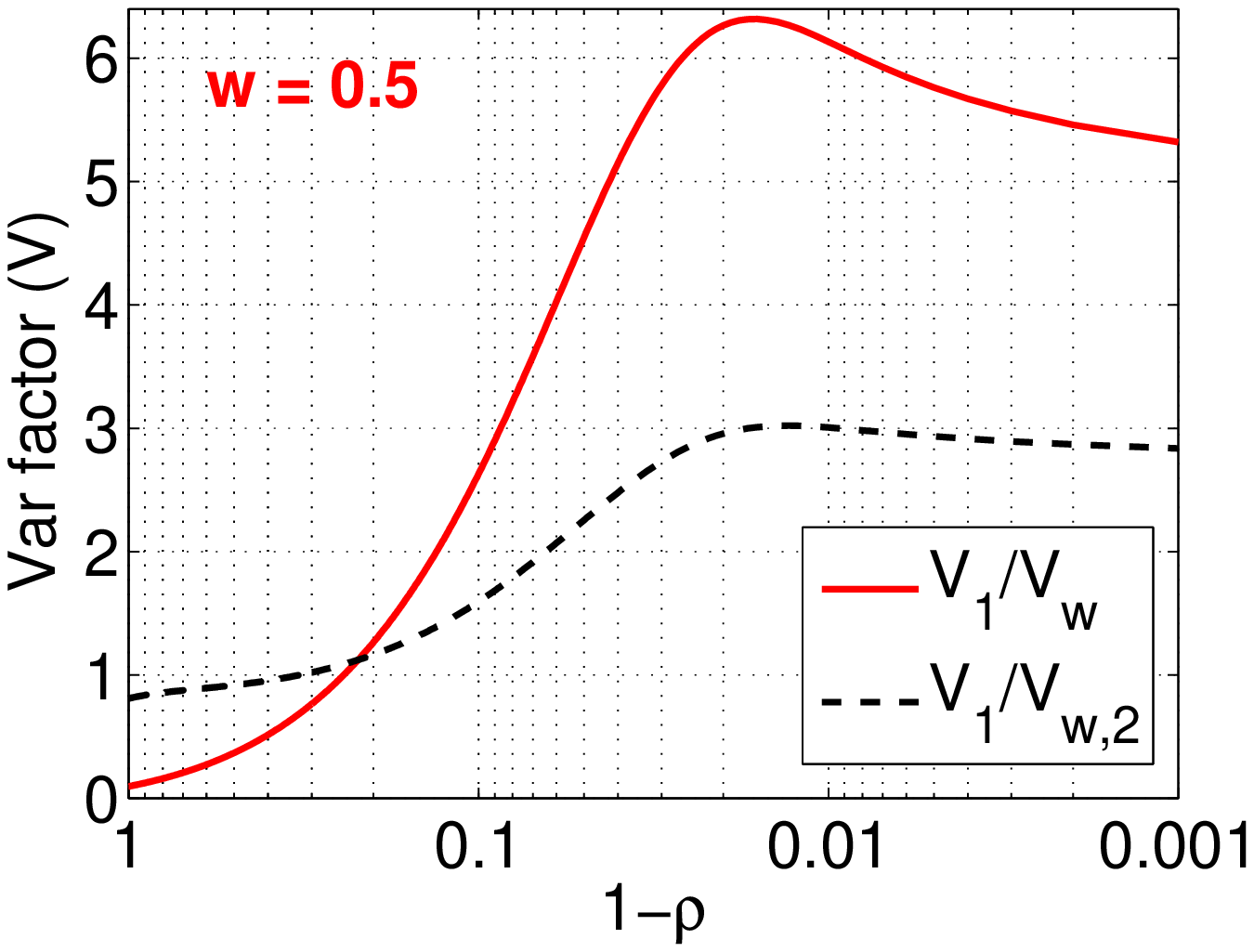}
\includegraphics[width=2.7in]{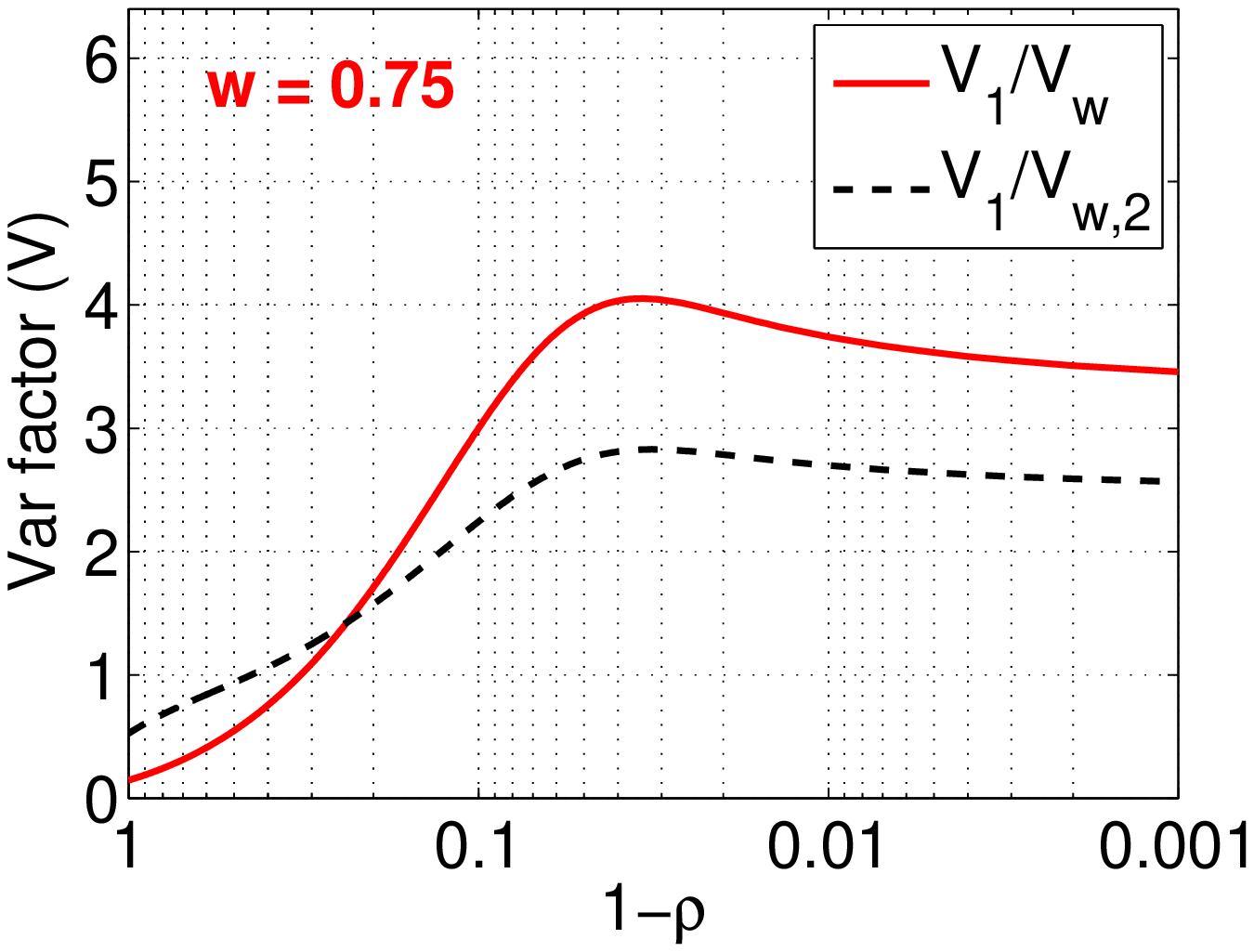}
}

\mbox{
\includegraphics[width=2.7in]{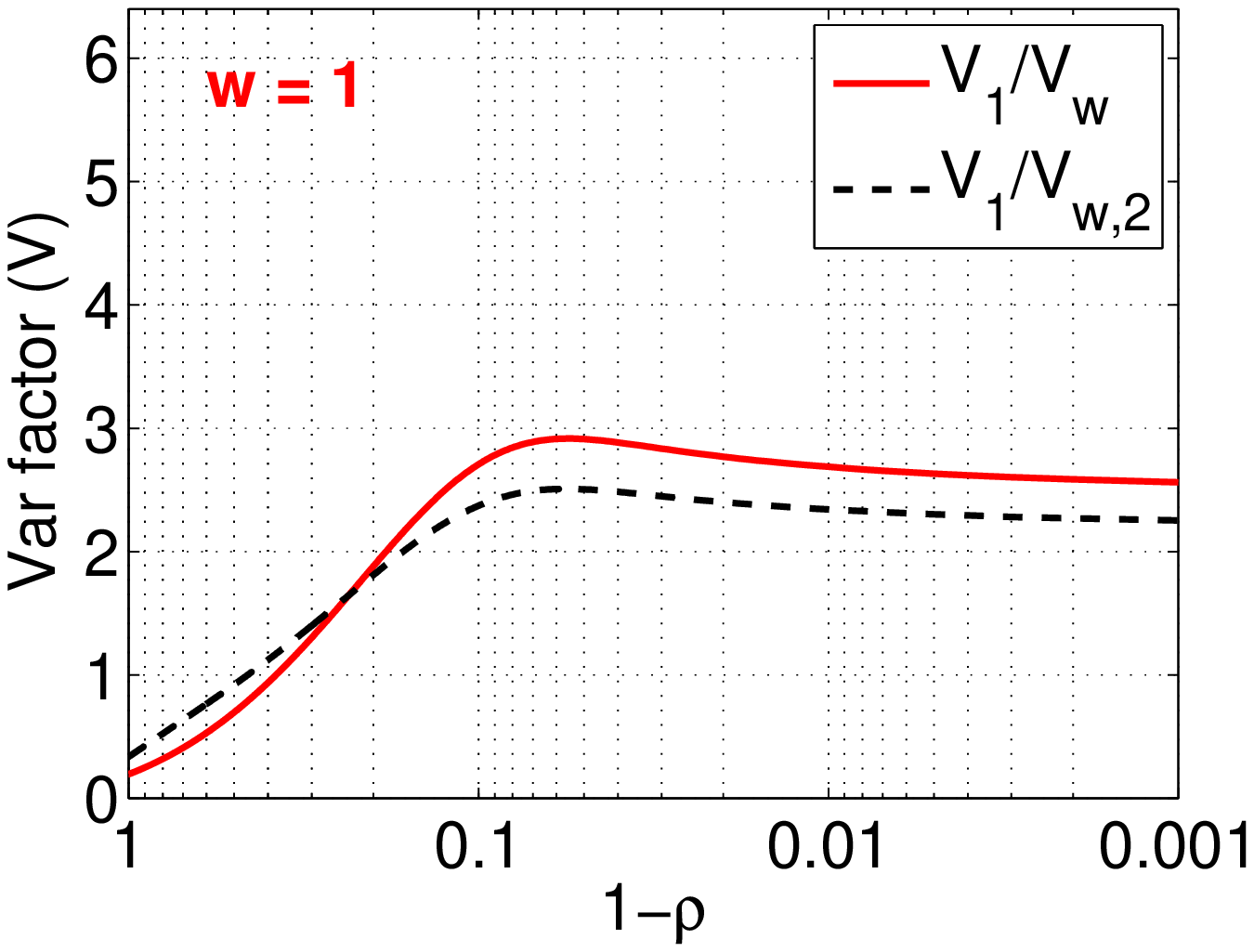}
\includegraphics[width=2.7in]{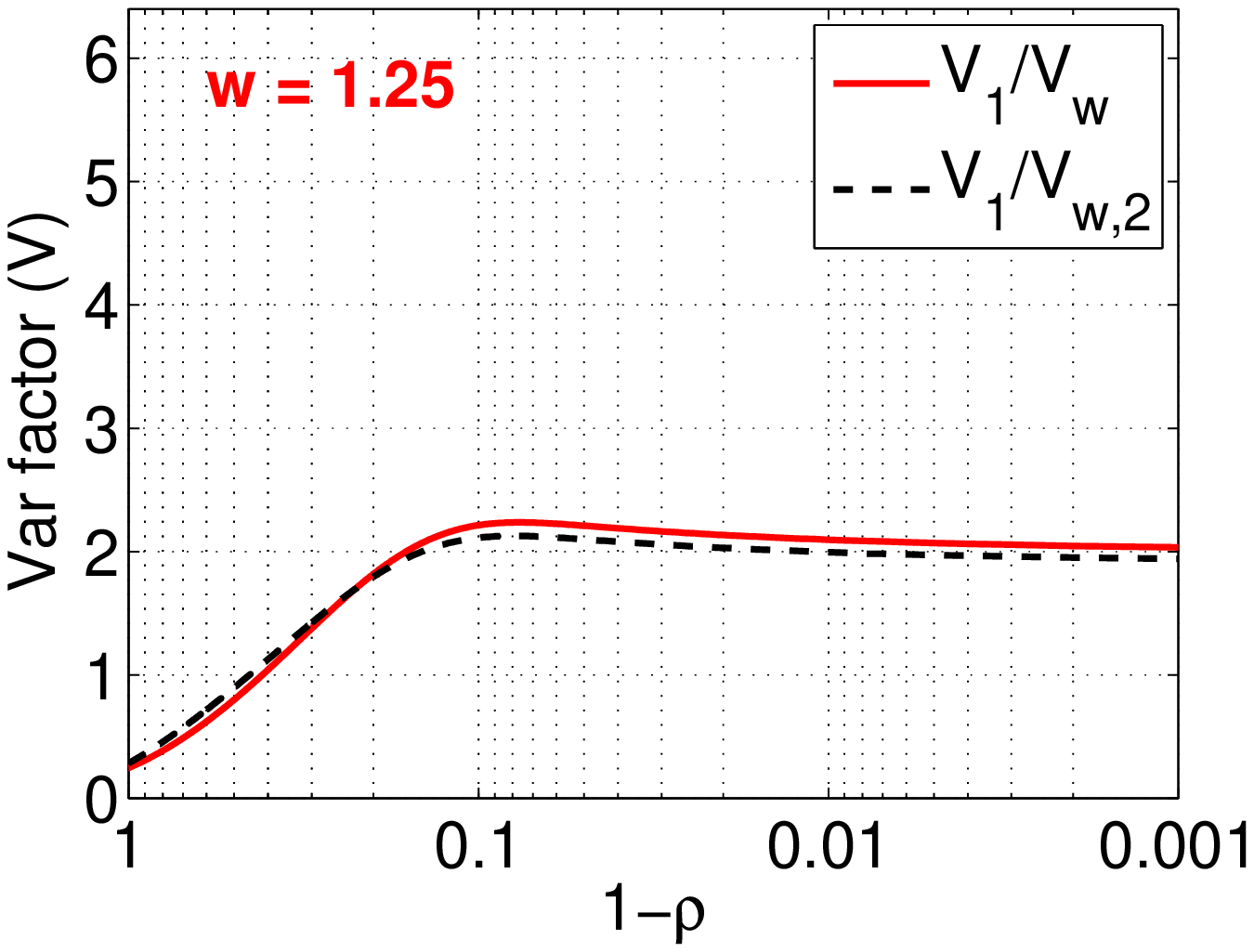}
}

\end{center}
\vspace{-0.2in}
\caption{Variance ratios: $\frac{Var\left(\hat{\rho}_1\right)}{Var\left(\hat{\rho}_{w}\right)}$ and $\frac{Var\left(\hat{\rho}_1\right)}{Var\left(\hat{\rho}_{w,2}\right)}$, for four selected $w$ values. In the high similarity region (e.g., $\rho\geq 0.9$), the 2-bit scheme $h_{w,2}$ significantly outperforms the 1-bit scheme $h_1$. In the low similarity region, $h_{w,2}$ still works reasonably well while the performance of $h_{w}$ can be poor (e.g., when $w\leq 0.75$). This justifies the recommendation of the 2-bit scheme.}\label{fig_VarRatioWV1VwVw2}
\end{figure}

Figure~\ref{fig_VarRatioV1VwVw2}  and Figure~\ref{fig_VarRatioWV1VwVw2}  plot the ratios of the variances: $\frac{Var\left(\hat{\rho}_1\right)}{Var\left(\hat{\rho}_{w}\right)}$ and $\frac{Var\left(\hat{\rho}_1\right)}{Var\left(\hat{\rho}_{w,2}\right)}$, to illustrate how much we lose in accuracy by using only one bit. Note $\hat{\rho}_{1}$ is not related to the bin width $w$ while $Var\left(\hat{\rho}_{w}\right)$ and $Var\left(\hat{\rho}_{w,2}\right)$ are functions of $w$. In Figure~\ref{fig_VarRatioV1VwVw2}, we plot the maximum values of the ratios, i.e., we use the smallest $Var\left(\hat{\rho}_{w}\right)$ and $Var\left(\hat{\rho}_{w,2}\right)$ at each $\rho$. The ratios demonstrate that potentially both $h_{w}$ and $h_{w,2}$ could  substantially outperform $h_1$, the 1-bit scheme.

Note that in Figure~\ref{fig_VarRatioV1VwVw2}, we plot $1-\rho$ in the horizontal axis with log-scale, so that the high similarity region can be visualized better. In practice, many applications are often more interested in the high similarity region, for example, duplicate detections.

\vspace{0.2in}

In practice, we must pre-specify the quantization bin width $w$ in advance. Thus, the improvement of $h_w$ and $h_{w,2}$ over the 1-bit scheme $h_1$ will not be as drastic as shown in Figure~\ref{fig_VarRatioV1VwVw2}. For  more realistic comparisons, Figure~\ref{fig_VarRatioWV1VwVw2} plots $\frac{Var\left(\hat{\rho}_1\right)}{Var\left(\hat{\rho}_{w}\right)}$ and $\frac{Var\left(\hat{\rho}_1\right)}{Var\left(\hat{\rho}_{w,2}\right)}$, for fixed $w$ values.  This figure advocates the recommendation of the 2-bit coding scheme $h_{w,2}$:
\begin{enumerate}
\item In the high similarity region, $h_{w,2}$ significantly outperforms $h_1$. The improvement drops as $w$ becomes larger (e.g., $w>1$). $h_{w}$ also works well, in fact better than $h_{w,2}$ when $w$ is small.
\item In the low similarity region, $h_{w,2}$ still outperforms $h_1$ unless $\rho$ is very low and $w$ is not small. Note that the performance of $h_{w}$ is noticeably worse than $h_{w,2}$ and $h_1$ when $\rho$ is low. \\
\end{enumerate}

Thus, we believe the 2-bit scheme $h_{w,2}$ with $w$ around 0.75 provides an overall good compromise. In fact, this is consistent with our observation in the SVM experiments in Section~\ref{sec_SVM}.\\

\textbf{Can we simply use the 1-bit scheme? } When $w=0.75$, in the high similarity region, the variance ratio $\frac{Var\left(\hat{\rho}_1\right)}{Var\left(\hat{\rho}_{w,2}\right)}$ is between 2 and 3. Note that, per projected data value, the 1-bit scheme requires 1 bit but the 2-bit scheme needs 2 bits. In a sense, the performance of $h_{w,2}$ and $h_1$ is actually similar in terms of the total number bits to store the (coded) projected data, according the analysis in this paper.

For similarity estimation, we believe it is  preferable to use the 2-bit scheme, for the following reasons:
\begin{itemize}
\item The processing cost of the 2-bit scheme would be lower. If we use $k$ projections for the 1-bit scheme and $k/2$ projections for the 2-bit scheme, although they have the same storage cost, the processing cost of $h_{w,2}$  for generating the projections would be only 1/2 of $h_{1}$. For very high-dimensional data, the processing cost can be substantial.
\item As we will show in Section~\ref{sec_SVM}, when we train a linear classifier (e.g., using LIBLINEAR), we need to expand the projected data into a binary vector with exact $k$ 1's if we use $k$ projections for both $h_1$ and $h_{w,2}$. For this application, we observe the training time is mainly determined by the number of nonzero entries and the quality of the input data. Even with the same $k$, we observe the training speed  on the input data generated by $h_{w,2}$ is often slightly faster than using the data generated by $h_1$.
\item In this study, we restrict our attention to linear estimators (which can be written as inner products) by simply using the (overall) collision probability, e.g.,  $P_{w,2} = \mathbf{Pr}\left(h_{w,2}^{(j)}(u) = h^{(j)}_{w,2}(v)\right)$.  There is significant room for improvement by using more refined estimators. For example, we can treat this problem as a contingency table whose cell probabilities are functions of the similarity $\rho$ and hence we can estimate $\rho$ by solving a maximum likelihood equation. Such an estimator is still useful for many applications (e.g., nonlinear kernel SVM). We will report that work separately, to maintain the simplicity of this paper.
\end{itemize}
Note that quantization is a non-reversible process. Once we quantize the data by the 1-bit scheme, there is no hope of recovering any information other than the signs.  Our work provides the necessary theoretical justifications for making practical choices of the coding schemes.

\section{An Experimental Study for Training Linear SVM}\label{sec_SVM}

We conduct experiments  with random projections for training ($L2$-regularized) linear SVM (e.g., LIBLINEAR~\cite{Article:Fan_JMLR08}) on three high-dimensional  datasets: {\em ARCENE, FARM, URL}, which are available from the UCI repository.  The original {\em URL} dataset has about 2.4 million examples (collected in 120 days) in   3231961 dimensions. We only used the data from the first day, with 10000 examples for training and 10000 for testing. The {\em FARM} dataset has 2059 training  and 2084 testing examples in 54877 dimensions. The {\em ARCENE} dataset contains 100 training and 100 testing examples in 10000 dimensions.\\

We implement the four coding schemes studied in this paper: $h_{w,q}$, $h_w$, $h_{w,2}$, and $h_{1}$. Recall $h_{w,q}$~\cite{Proc:Datar_SCG04} was based on uniform quantization plus a random offset,  with bin width $w$. Here, we first illustrate exactly how we utilize the coded data for training  linear SVM.  Suppose we use $h_{w,2}$ and $w=0.75$. We can code an original projected value $x$ into a vector of length 4 (i.e., 2-bit):
\begin{align}\notag
&x\in(-\infty\ -0.75) \Rightarrow [1\ 0\ 0\ 0 ], \hspace{0.5in} x\in[-0.75\ 0) \Rightarrow [0\ 1\ 0\ 0 ],\\\notag
&x\in[0\ 0.75) \Rightarrow [0\ 0\ 1\ 0 ],\hspace{0.9in} x\in[0.75\ \infty) \Rightarrow [0\ 0\ 0\ 1 ]
\end{align}
This way, with $k$ projections, for each feature vector, we obtain a new vector of length $4k$ with exactly $k$ 1's. This new vector is then fed to a solver such as LIBLINEAR. Recently, this strategy was  adopted for linear learning with binary data based on  {\em b-bit minwise hashing}~\cite{Proc:Li_Konig_WWW10,Proc:Li_Owen_Zhang_NIPS12}.

Similarly, when using $h_1$, the dimension of the new vector is $2k$ with exactly $k$ 1's. For $h_w$ and $h_{w,q}$, we must specify a cutoff value such as 6 otherwise they are ``infinite precision'' schemes. Practically speaking, because the normal density decays very rapidly at the tail (e.g., $1-\Phi(6) = 9.9\times 10^{-10}$), we essentially do not suffer from information loss if we choose a large enough cutoff such as 6.\\

\begin{figure}[h!]
\begin{center}

\mbox{
\includegraphics[width=2.2in]{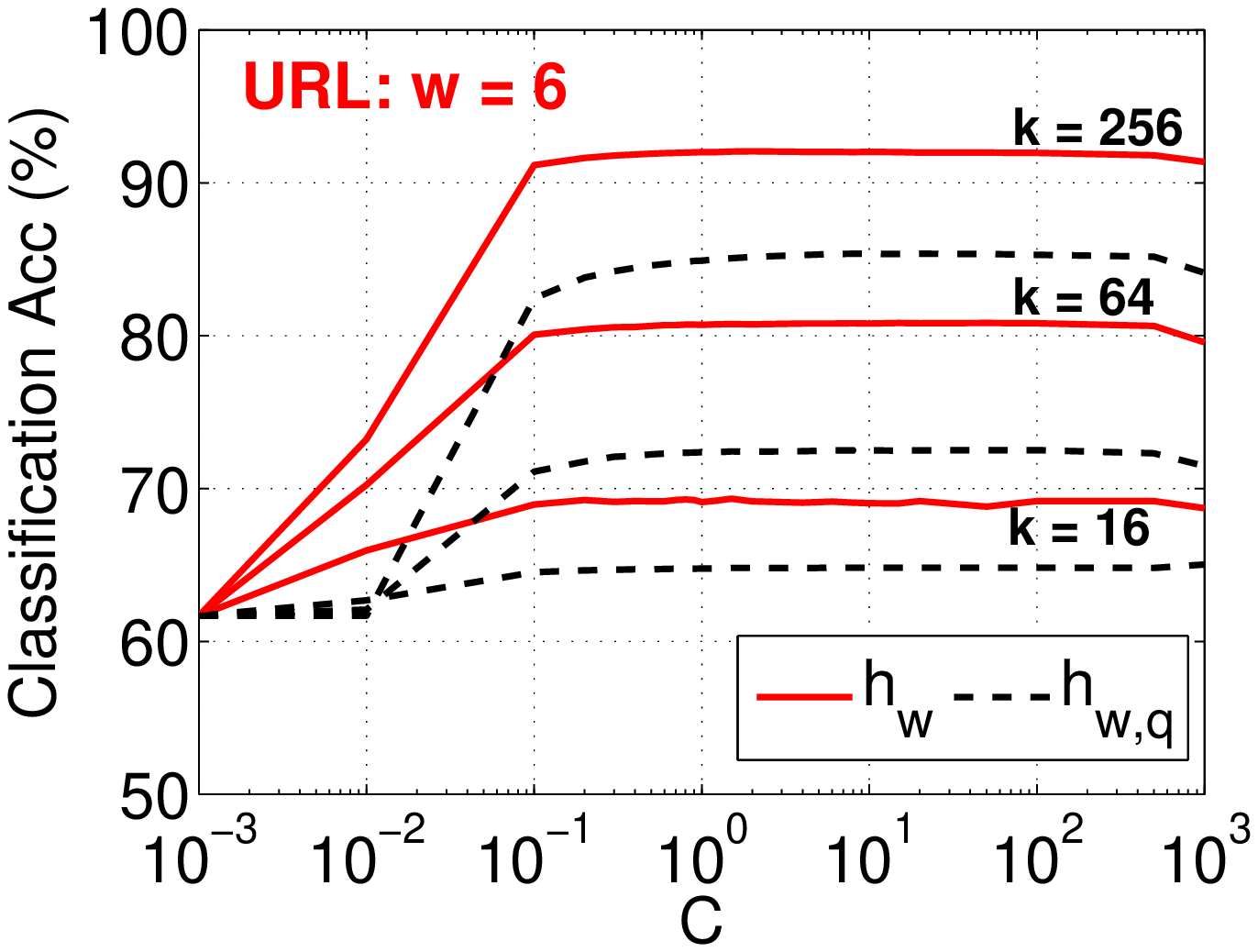}
\includegraphics[width=2.2in]{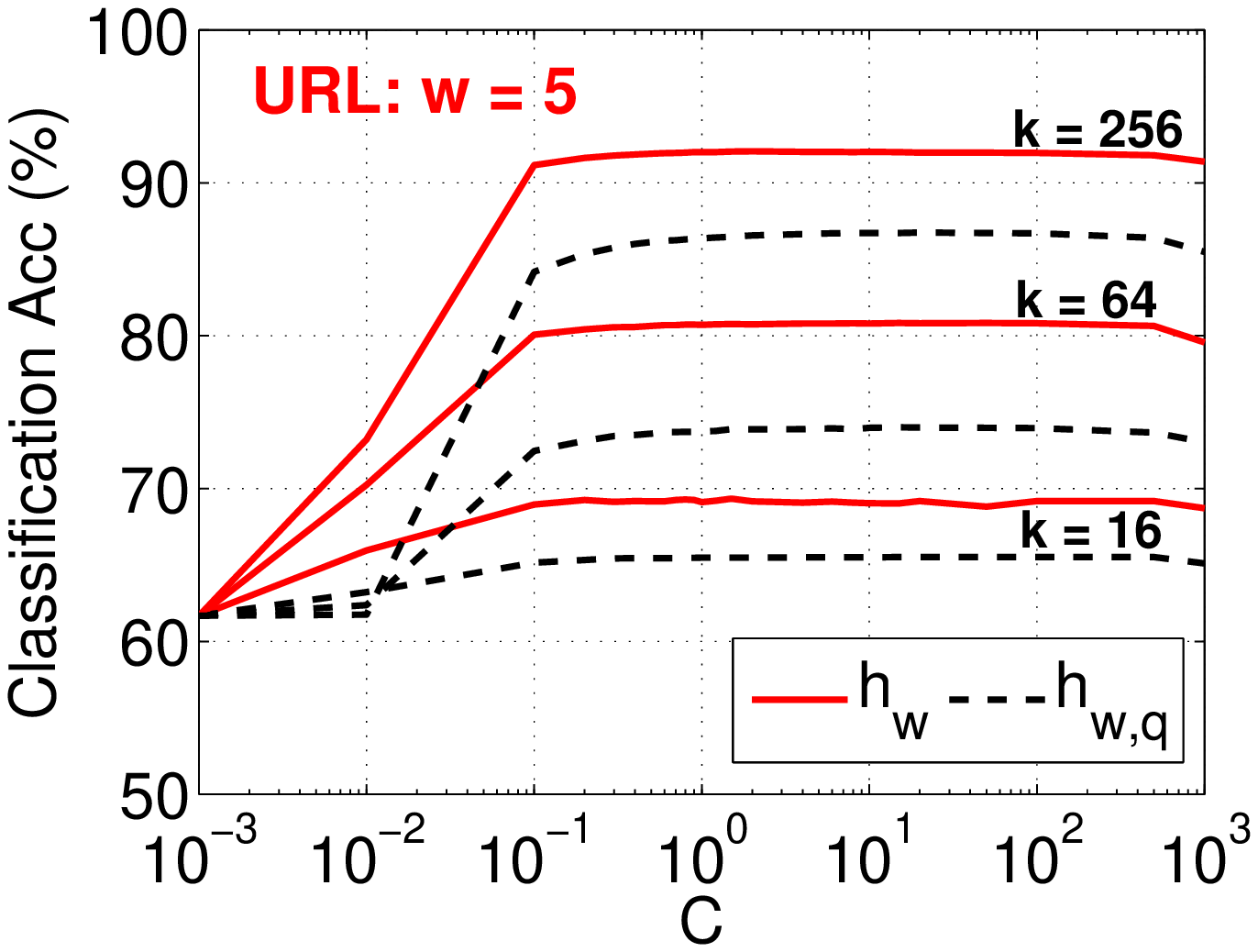}
\includegraphics[width=2.2in]{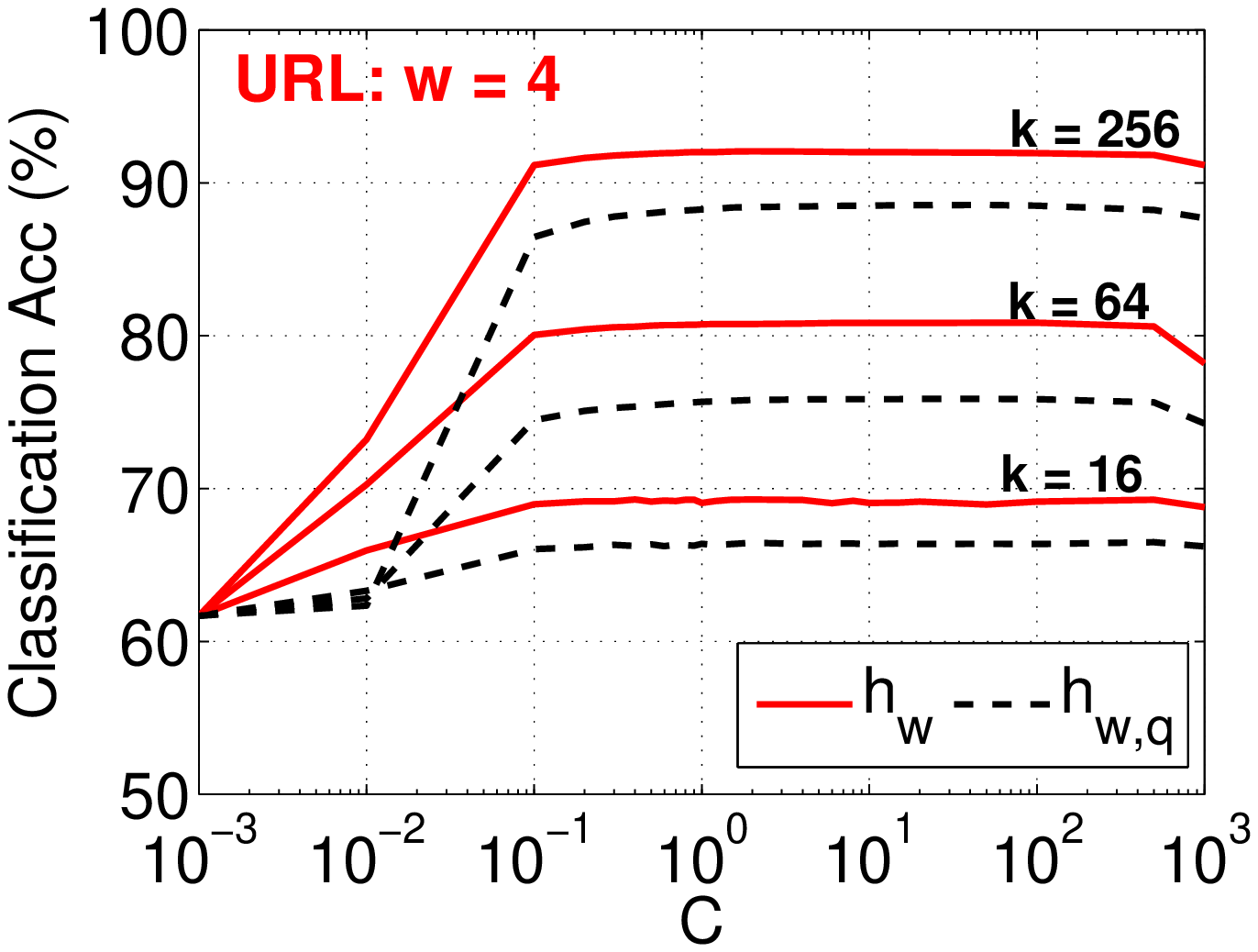}
}

\mbox{
\includegraphics[width=2.2in]{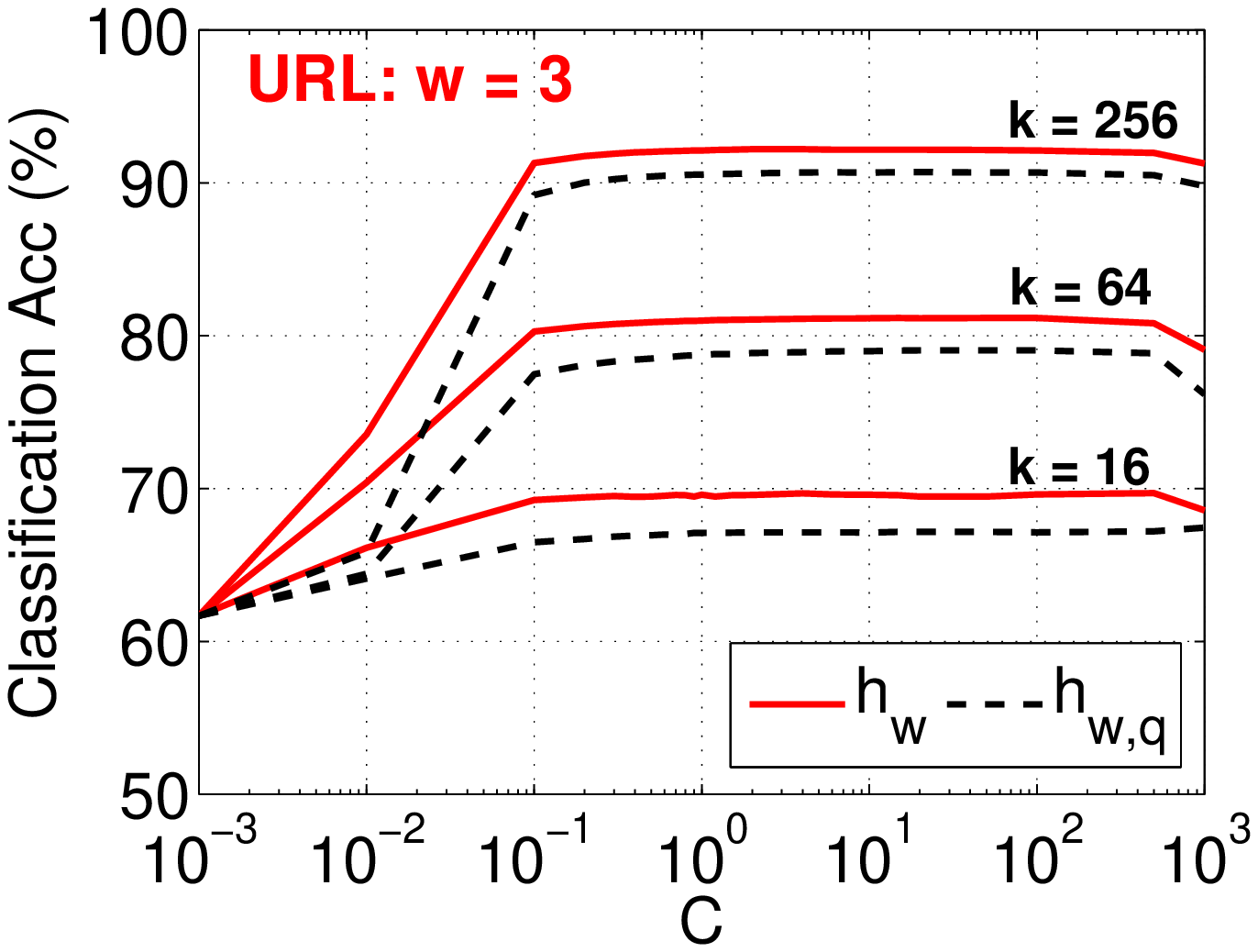}
\includegraphics[width=2.2in]{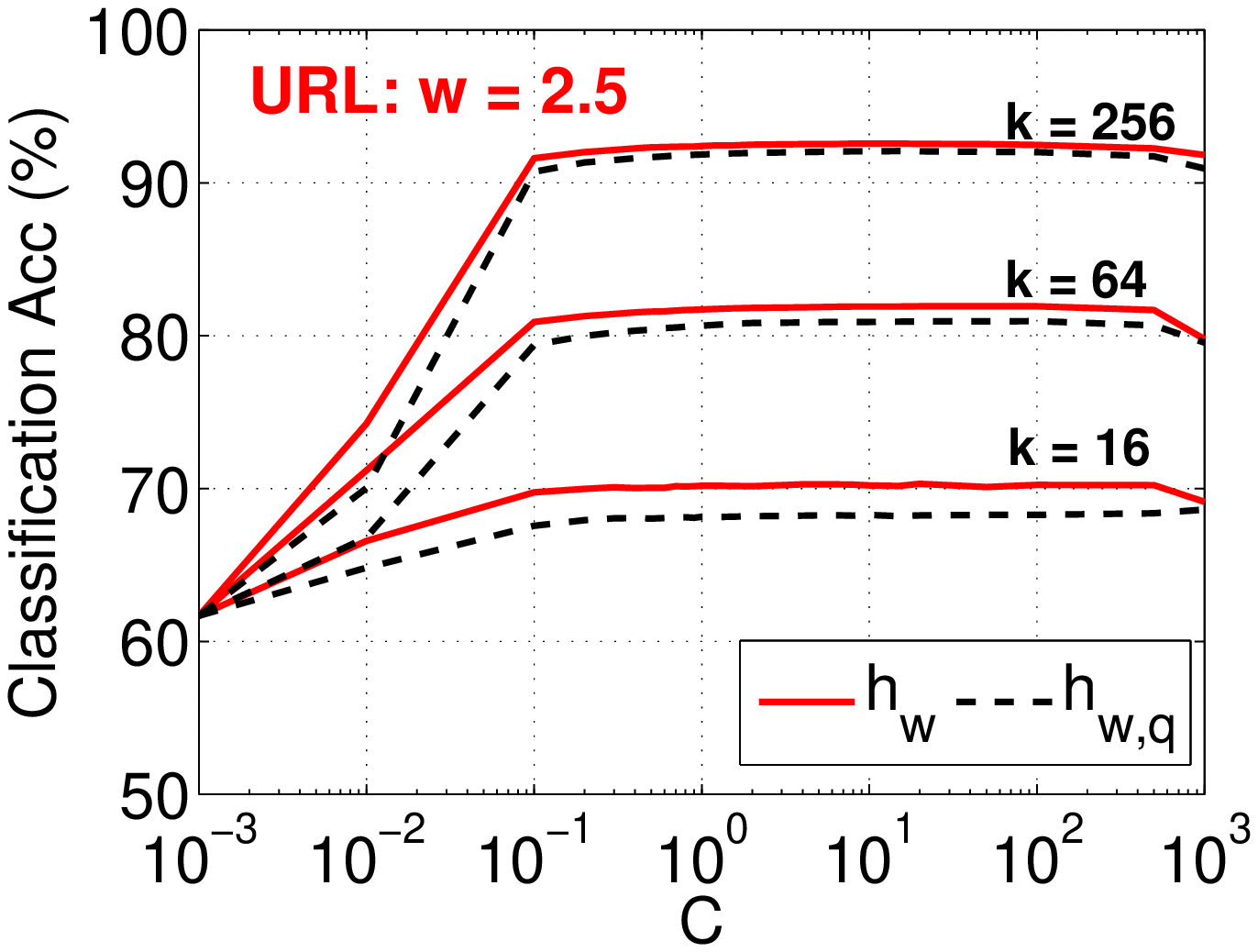}
\includegraphics[width=2.2in]{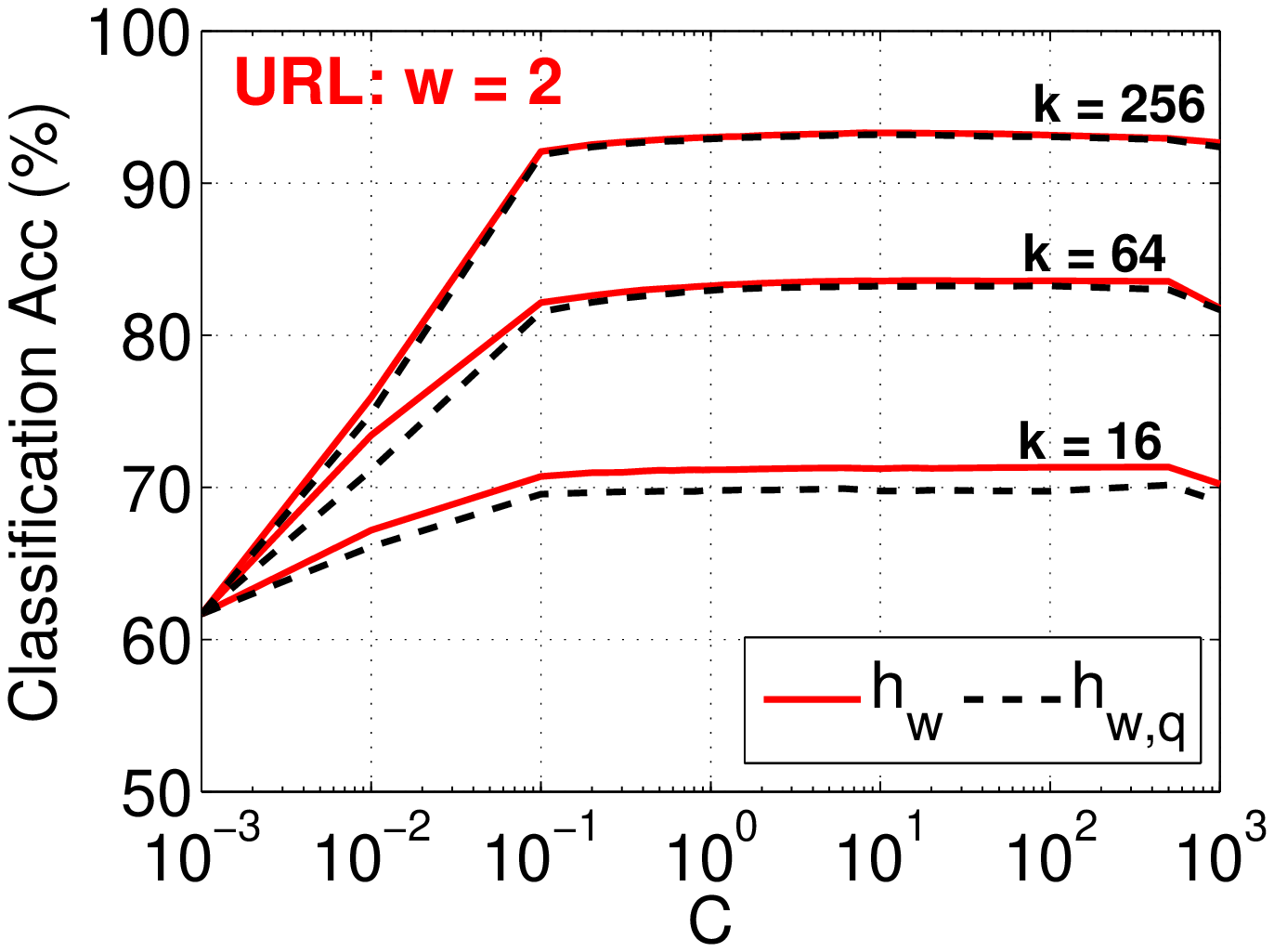}
}

\mbox{
\includegraphics[width=2.2in]{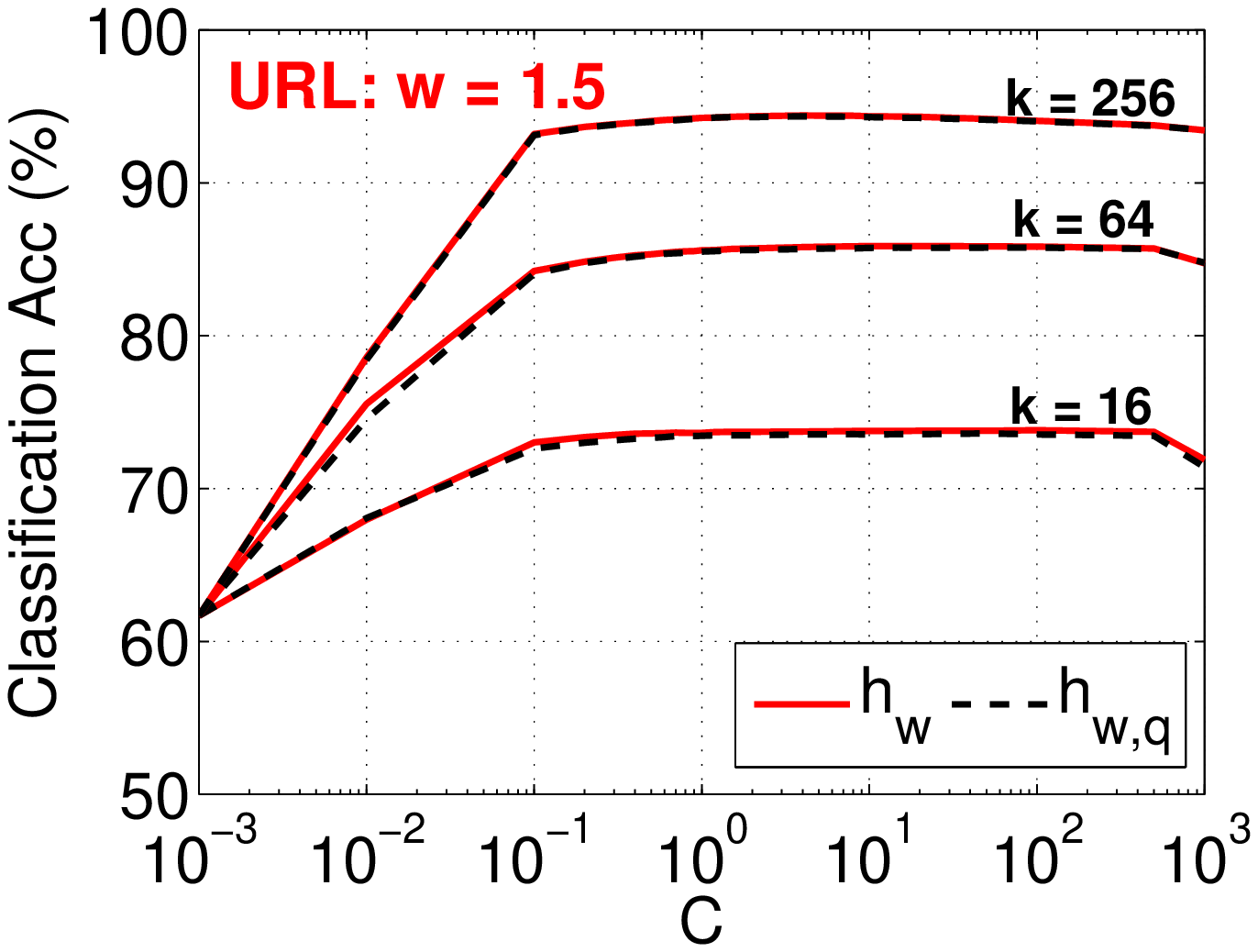}
\includegraphics[width=2.2in]{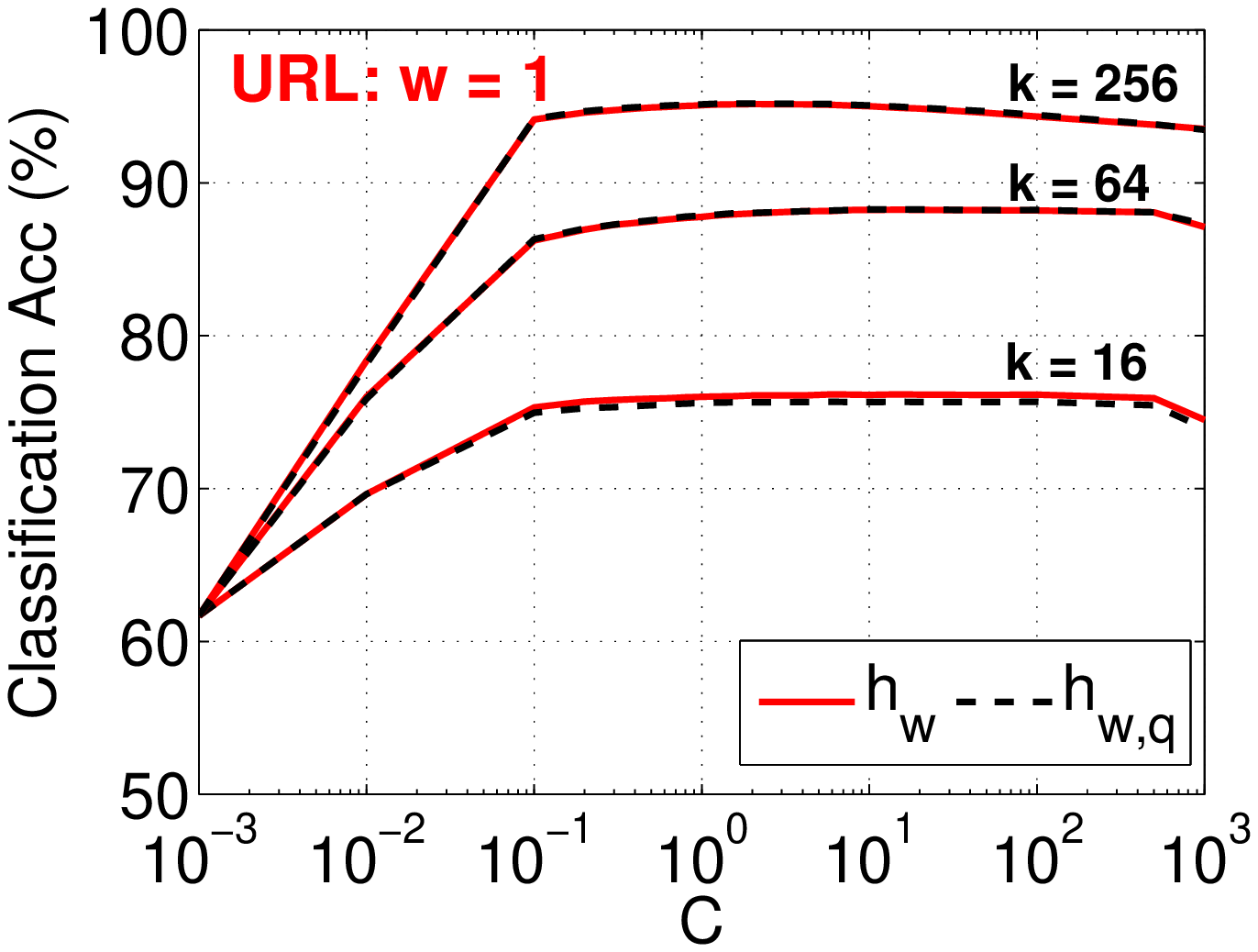}
\includegraphics[width=2.2in]{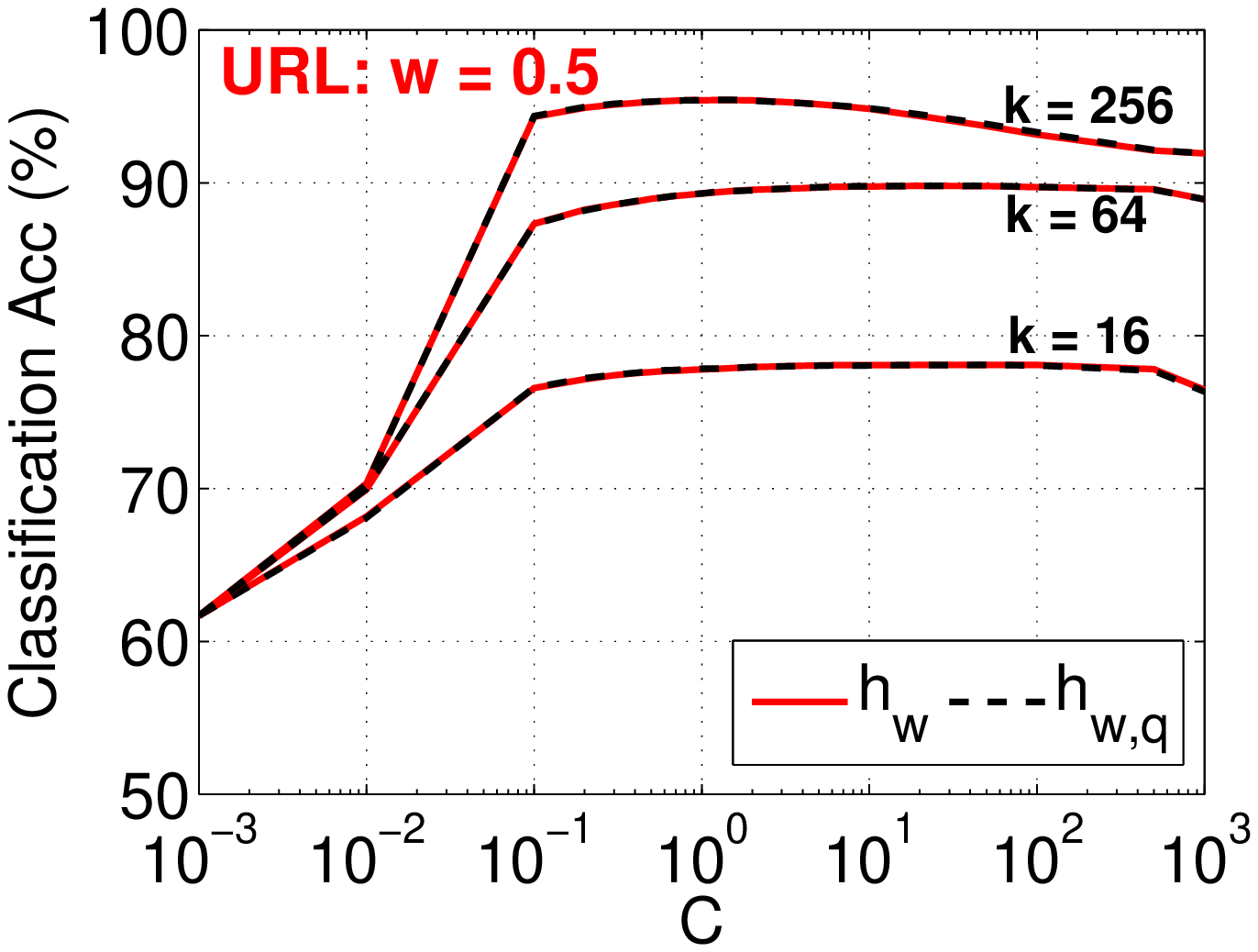}
}

\end{center}
\vspace{-0.20in}
\caption{Test accuracies on the {\em URL} dataset using LIBLINEAR, for comparing two coding schemes:  $h_w$ vs. $h_{w,q}$. Recall $h_{w,q}$~\cite{Proc:Datar_SCG04} was based on uniform quantization plus a random offset,  with bin length $w$. Our proposed scheme $h_w$ removes the random offset.  We report the classification results for $k=16$, $k=64$, and $k=256$. In each panel, there are 3 solid curves for $h_w$ and 3 dashed curves for $h_{w,q}$. We report the results for a wide range of $L_2$-regularization parameter $C$ (i.e., the horizontal axis).  When $w$ is large, $h_{w}$ noticeably outperforms $h_{w,q}$. When $w$ is small, the two schemes perform very similarly. This observation is of course expected from our analysis of the estimation variances. This experiment  confirms that the random offset step of $h_{w,q}$ may not be needed.  }\label{fig_UrlSVMAccQ}
\end{figure}

Figure~\ref{fig_UrlSVMAccQ} reports the test accuracies on the {\em URL} data, for comparing $h_{w,q}$ with $h_w$. The results basically confirm our analysis of the estimation variances. For small bin width $w$, the two schemes perform very similarly. However, when using a relatively large $w$, the scheme $h_{w,q}$ suffers from noticeable reduction of classification accuracies. The experimental results on the other two datasets demonstrate the same phenomenon. This experiment confirms that the step of random offset in $h_{w,q}$ is not  needed, at least for similarity estimation and training linear classifiers.

There is one tuning parameter $C$ in linear SVM. Figure~\ref{fig_UrlSVMAccQ} reports the accuracies for a wide range of $C$ values, from $10^{-3}$ to $10^3$. Before we feed the data to LIBLINEAR, we always normalize them to have unit norm (which is a recommended practice). Our experience is that, with normalized input data, the best accuracies are often attained around $C=1$, as verified in Figure~\ref{fig_UrlSVMAccQ}. For other figures in this section, we will only report $C$ from $10^{-3}$ to 10.

\begin{figure}[h!]
\begin{center}

\mbox{
\includegraphics[width=2.2in]{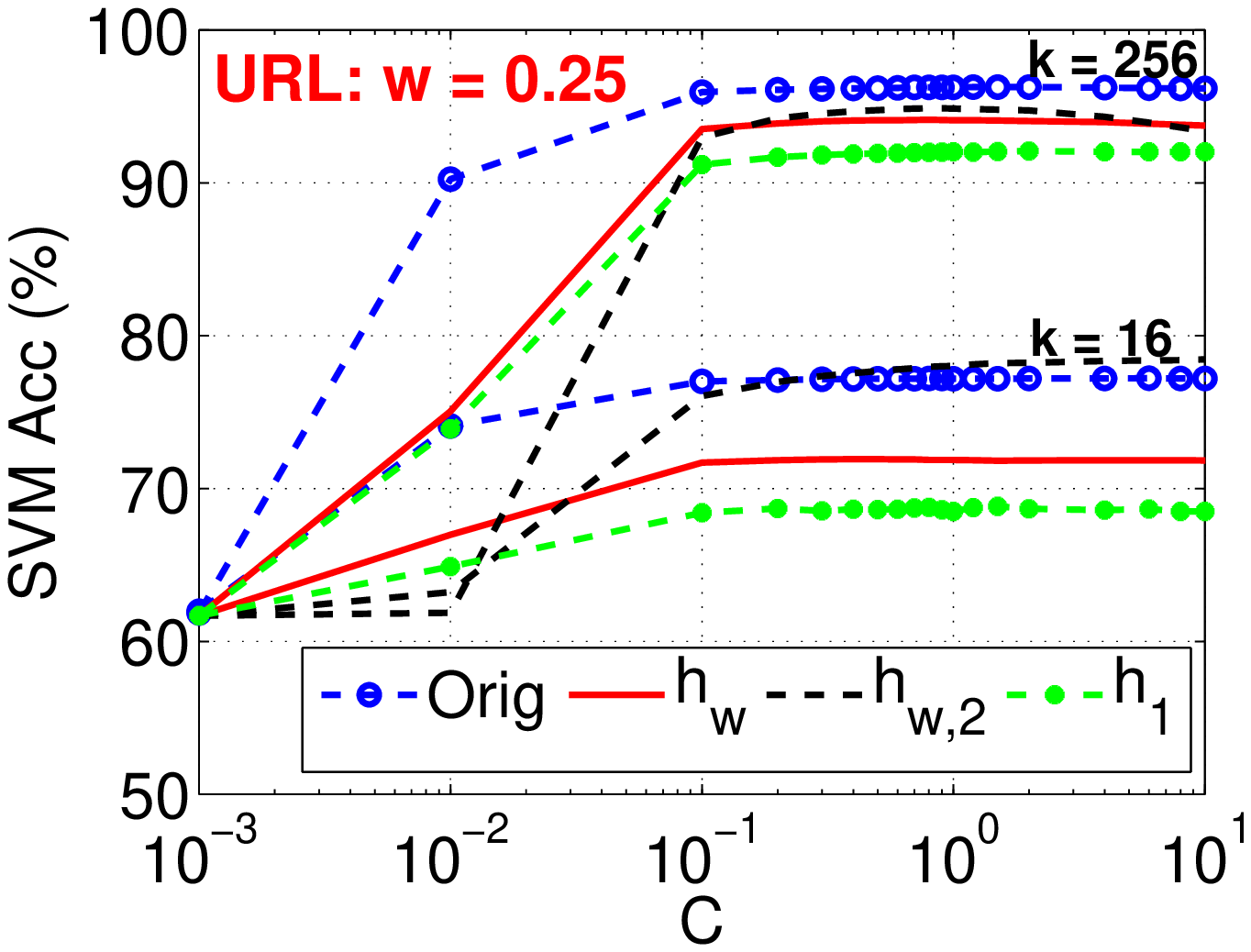}
\includegraphics[width=2.2in]{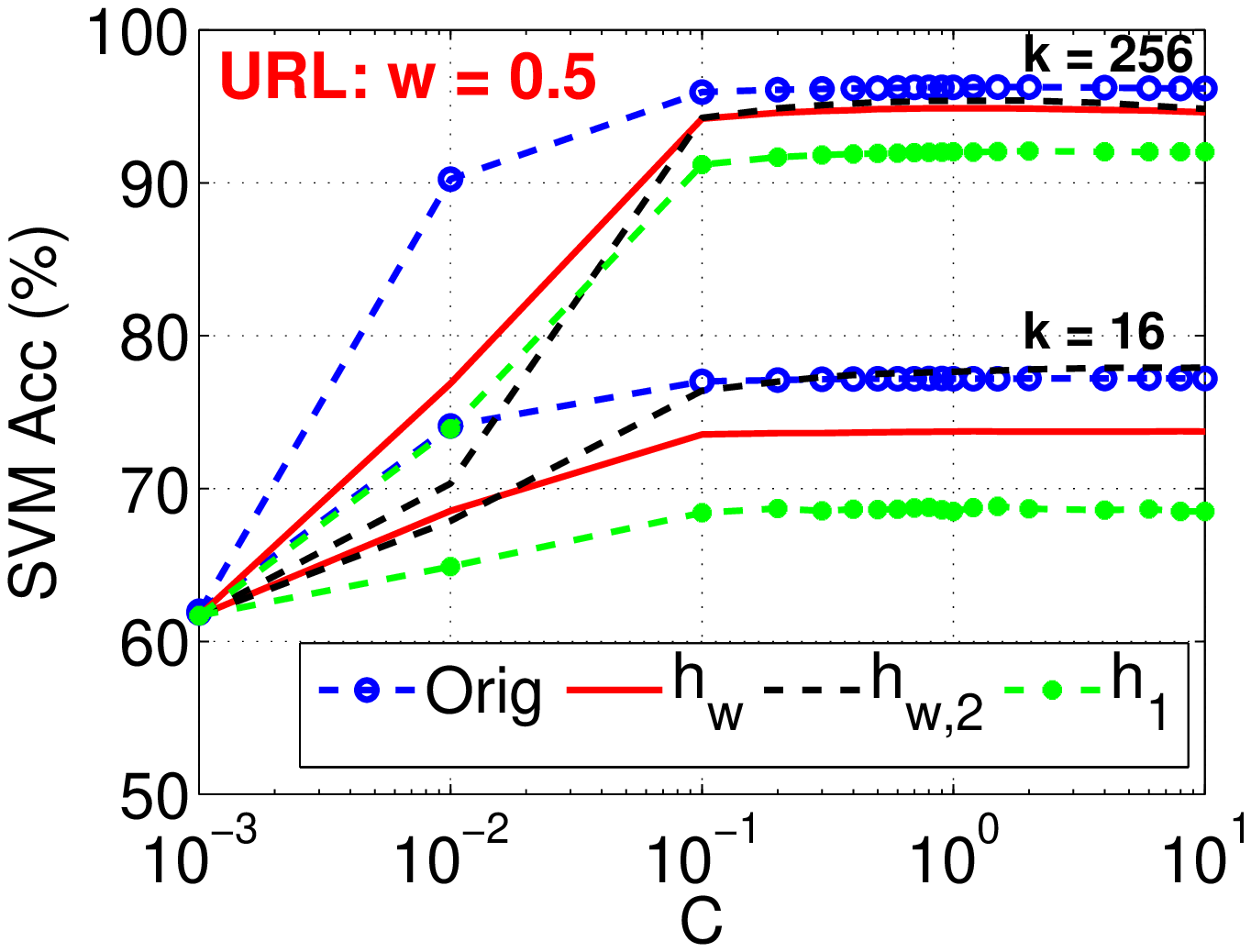}
\includegraphics[width=2.2in]{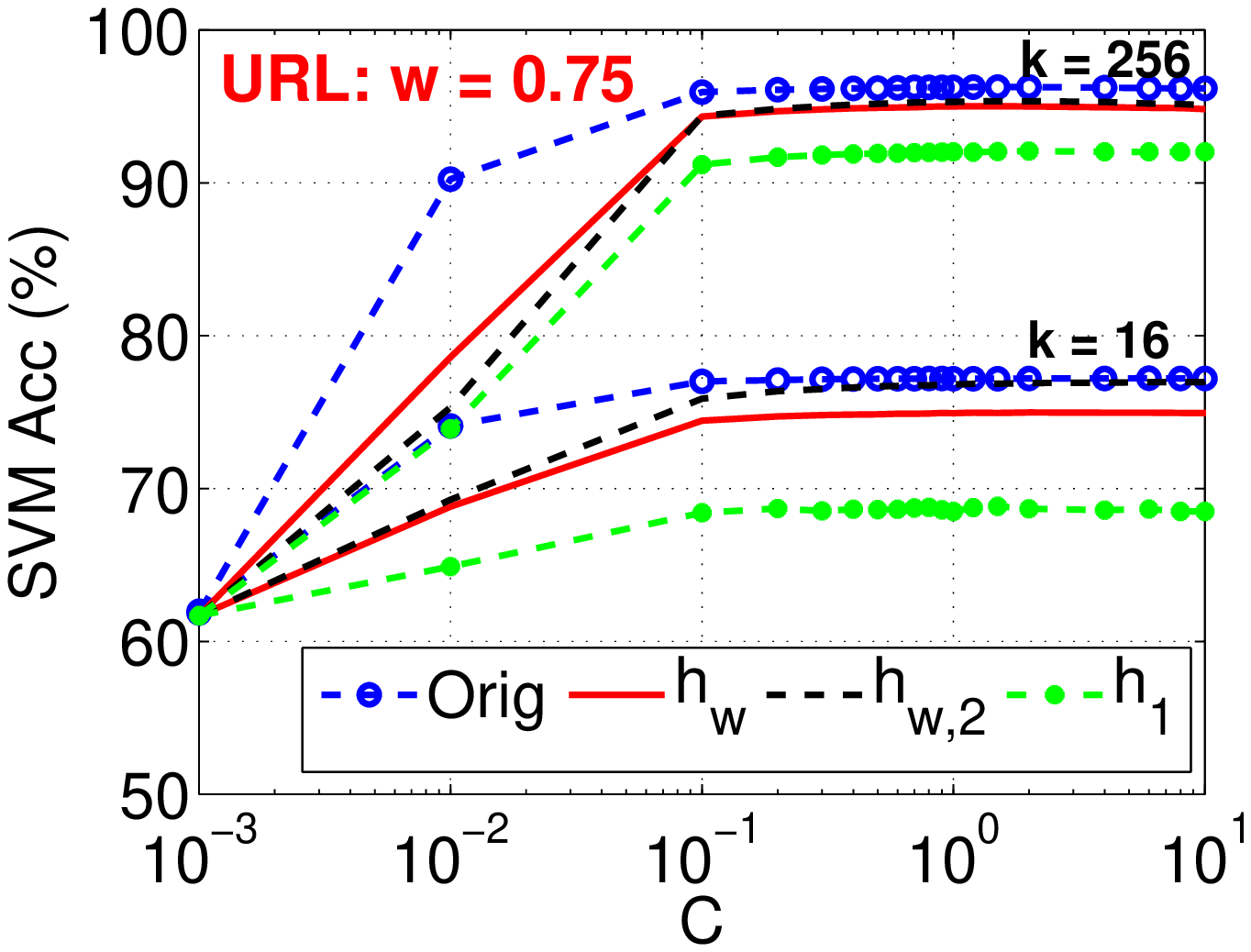}
}

\mbox{
\includegraphics[width=2.2in]{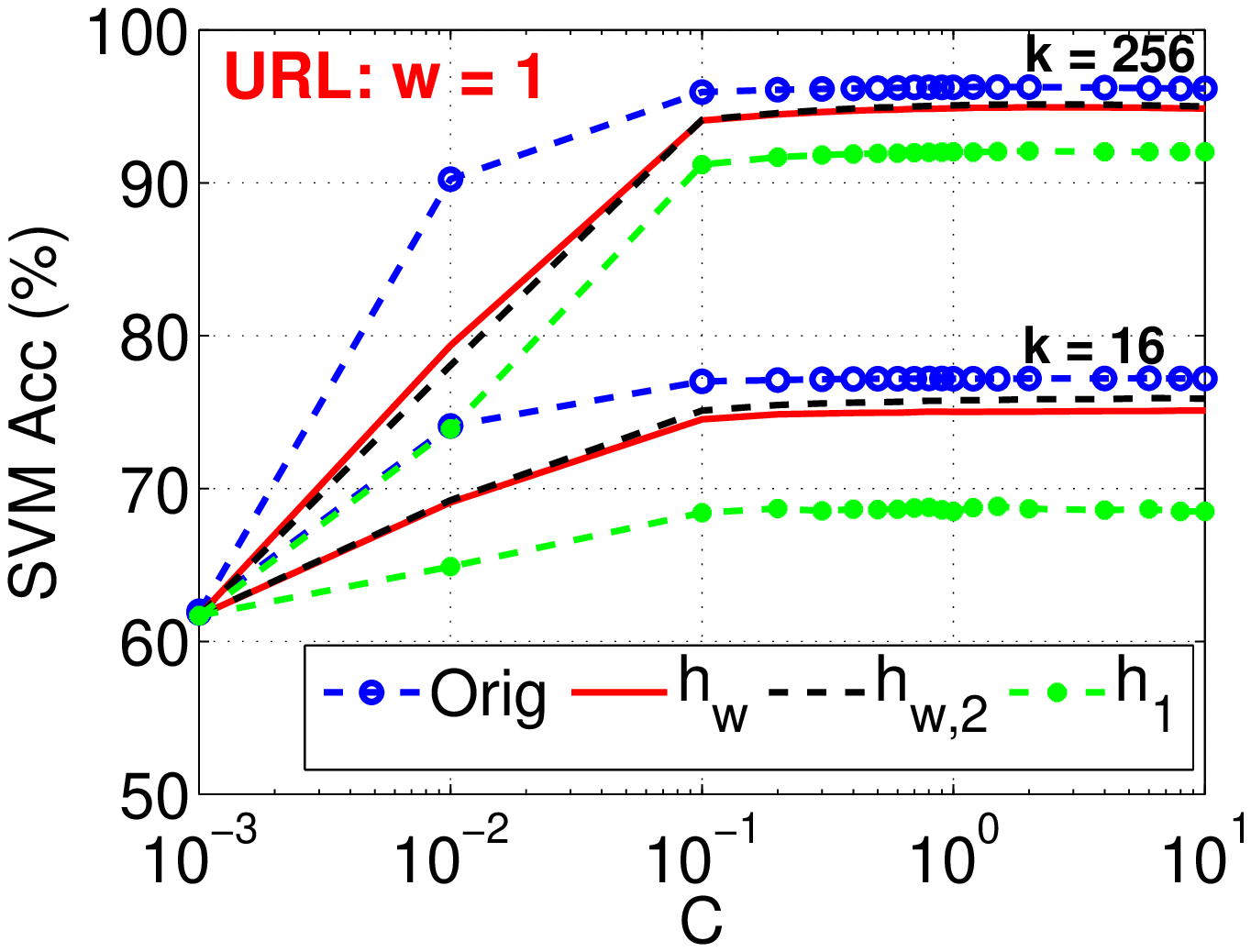}
\includegraphics[width=2.2in]{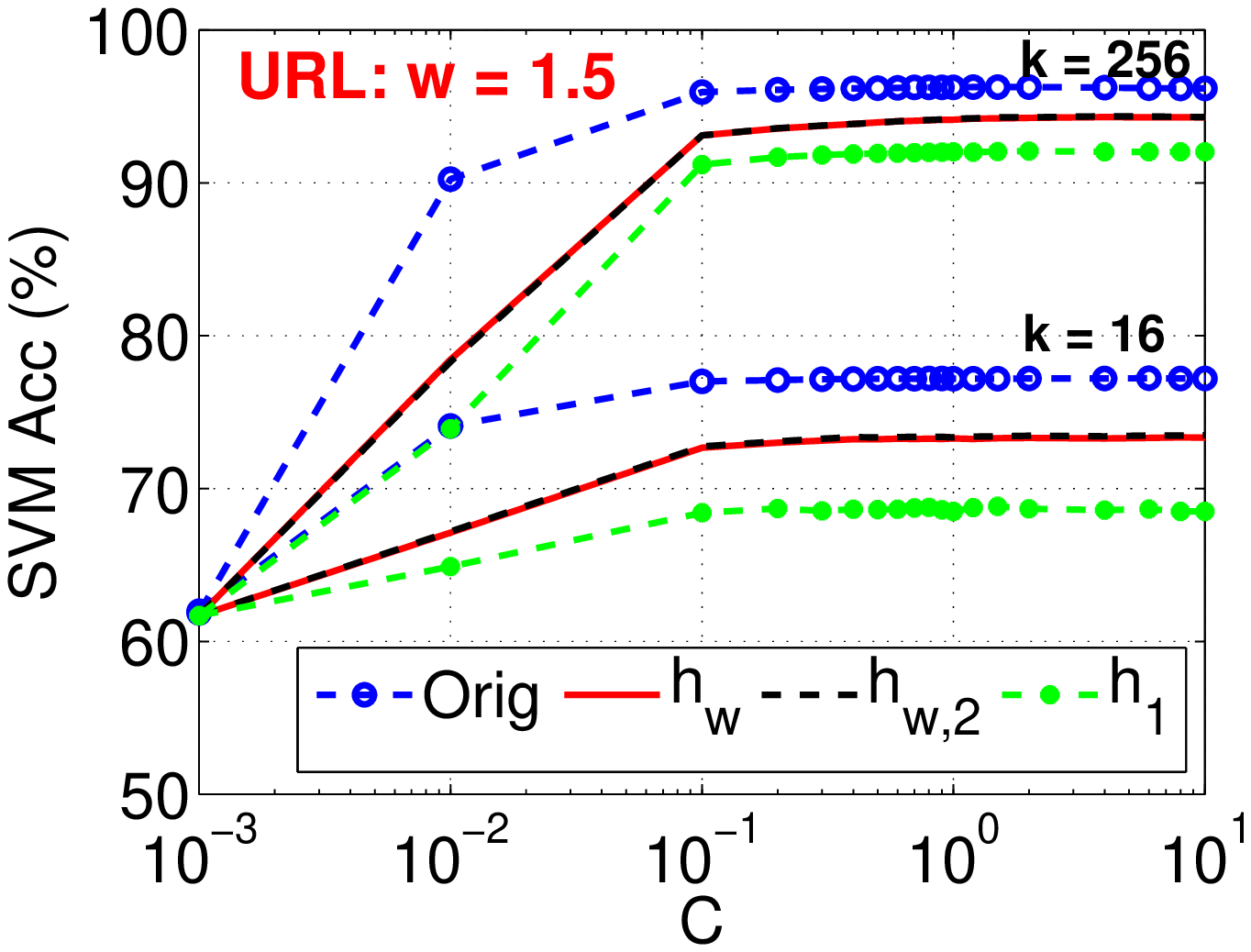}
\includegraphics[width=2.2in]{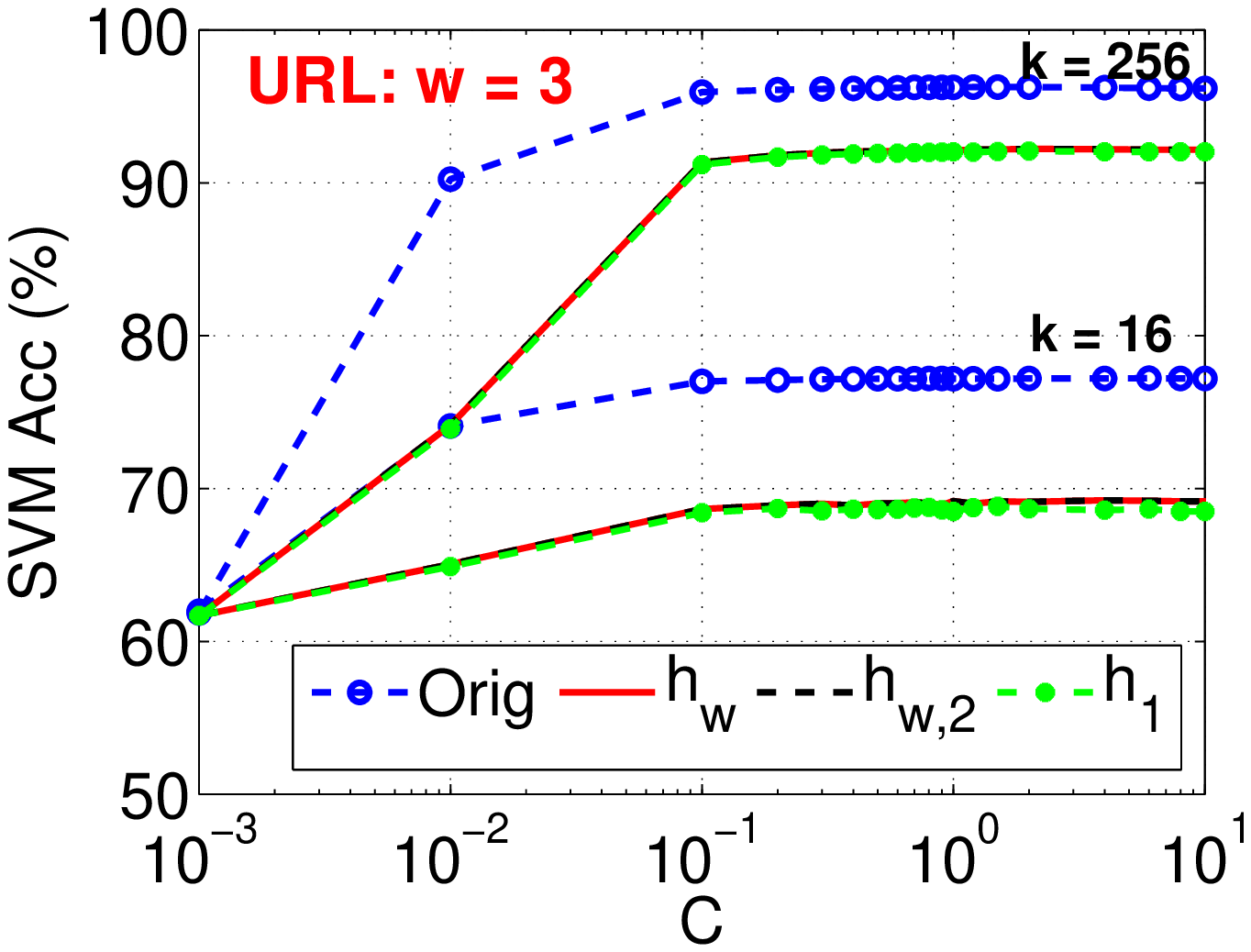}
}

\end{center}
\vspace{-0.20in}
\caption{Test accuracies on the {\em URL} dataset using LIBLINEAR, for comparing four coding schemes: uncoded (``Orig''), $h_w$, $h_{w,2}$, and $h_1$. We report the results for $k=16$ and $k=256$. Thus, in each panel, there are 2 groups of 4 curves. We report the results for a wide range of $L_2$-regularization parameter $C$ (i.e., the horizontal axis).  When $w=0.5\sim 1$, both $h_w$ and $h_{w,2}$ produce similar results as using the original projected data, while the 1-bit scheme $h_1$ is  less competitive. }\label{fig_UrlSVMAcc}
\end{figure}

\begin{figure}[h!]
\begin{center}

\mbox{
\includegraphics[width=2.2in]{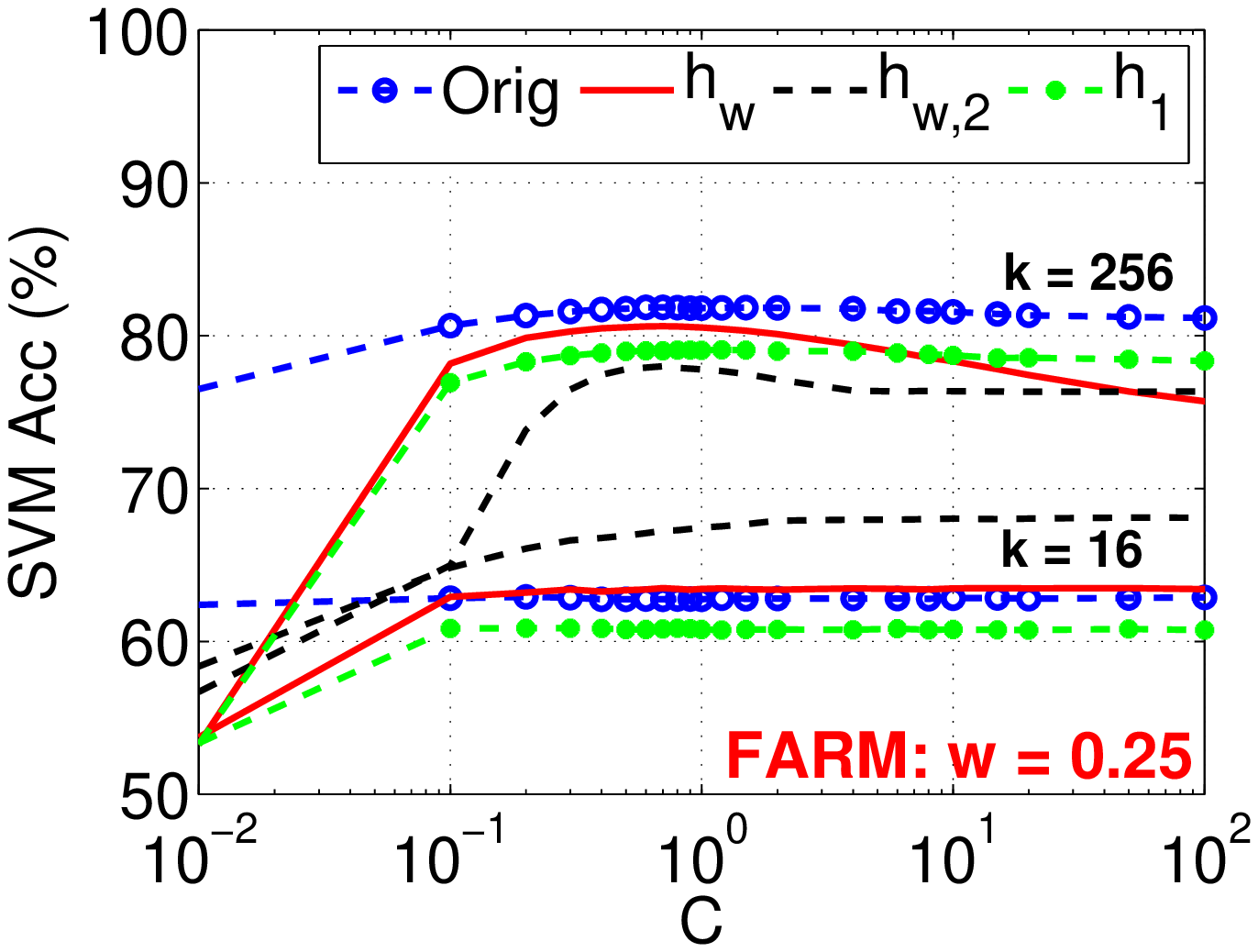}
\includegraphics[width=2.2in]{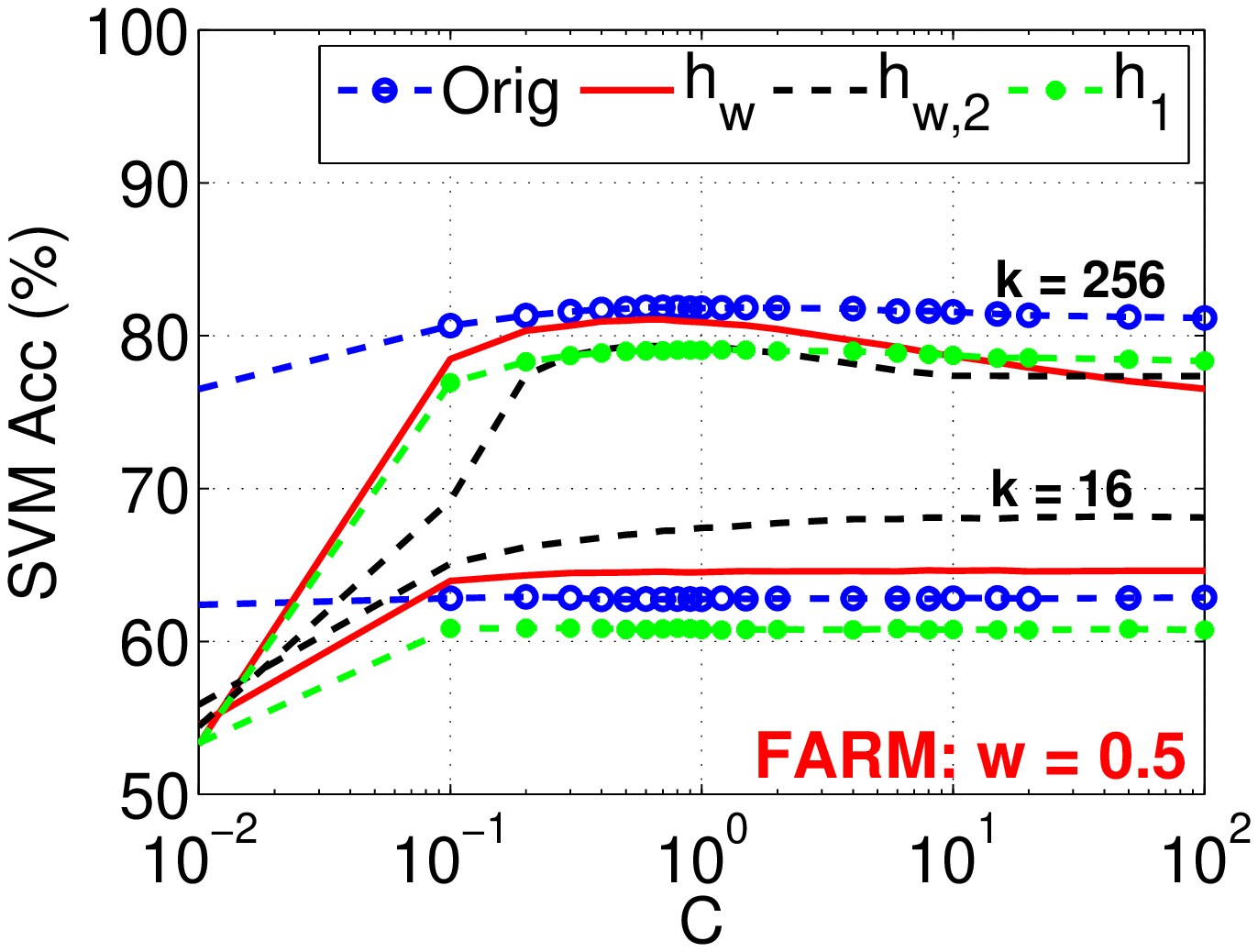}
\includegraphics[width=2.2in]{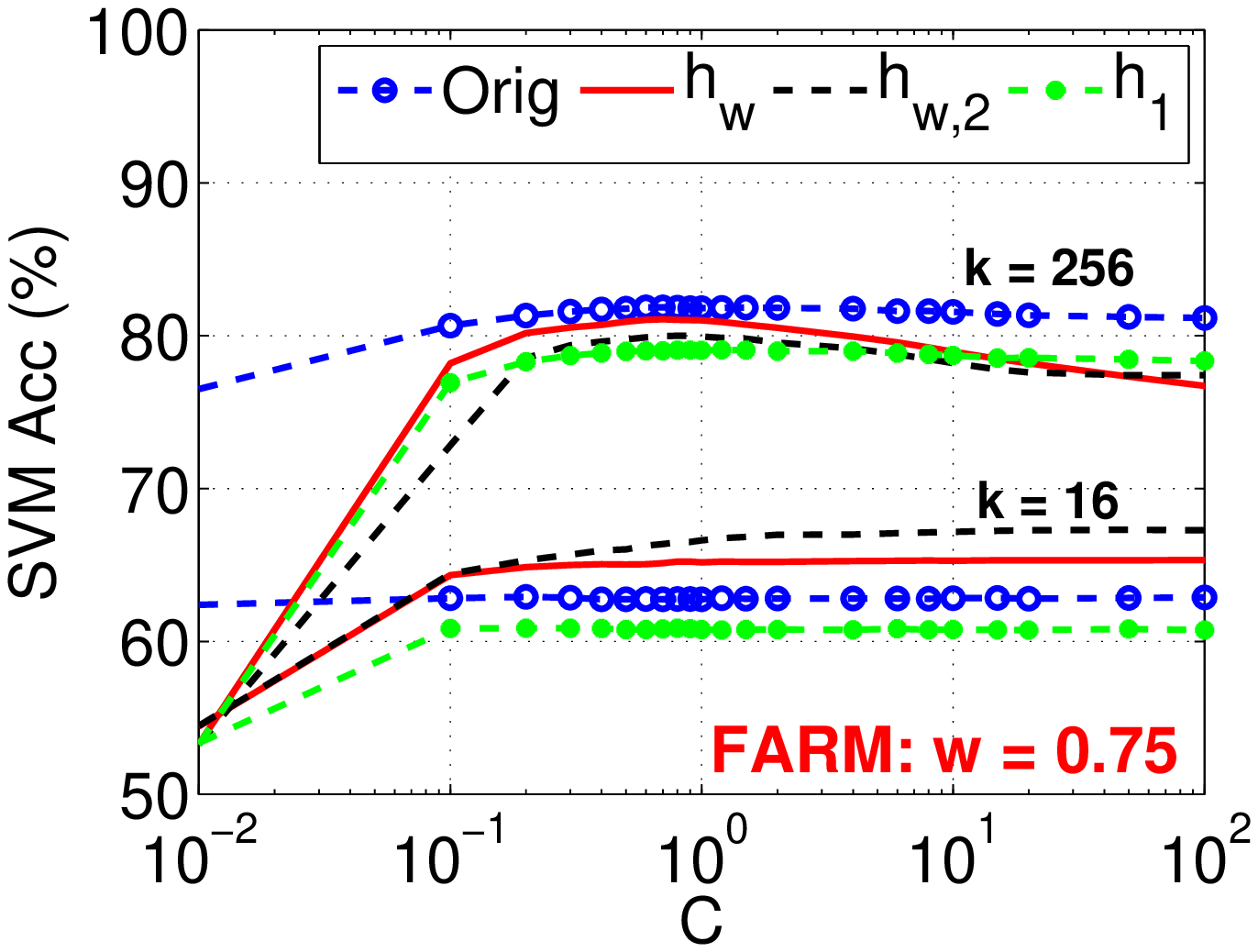}
}

\mbox{
\includegraphics[width=2.2in]{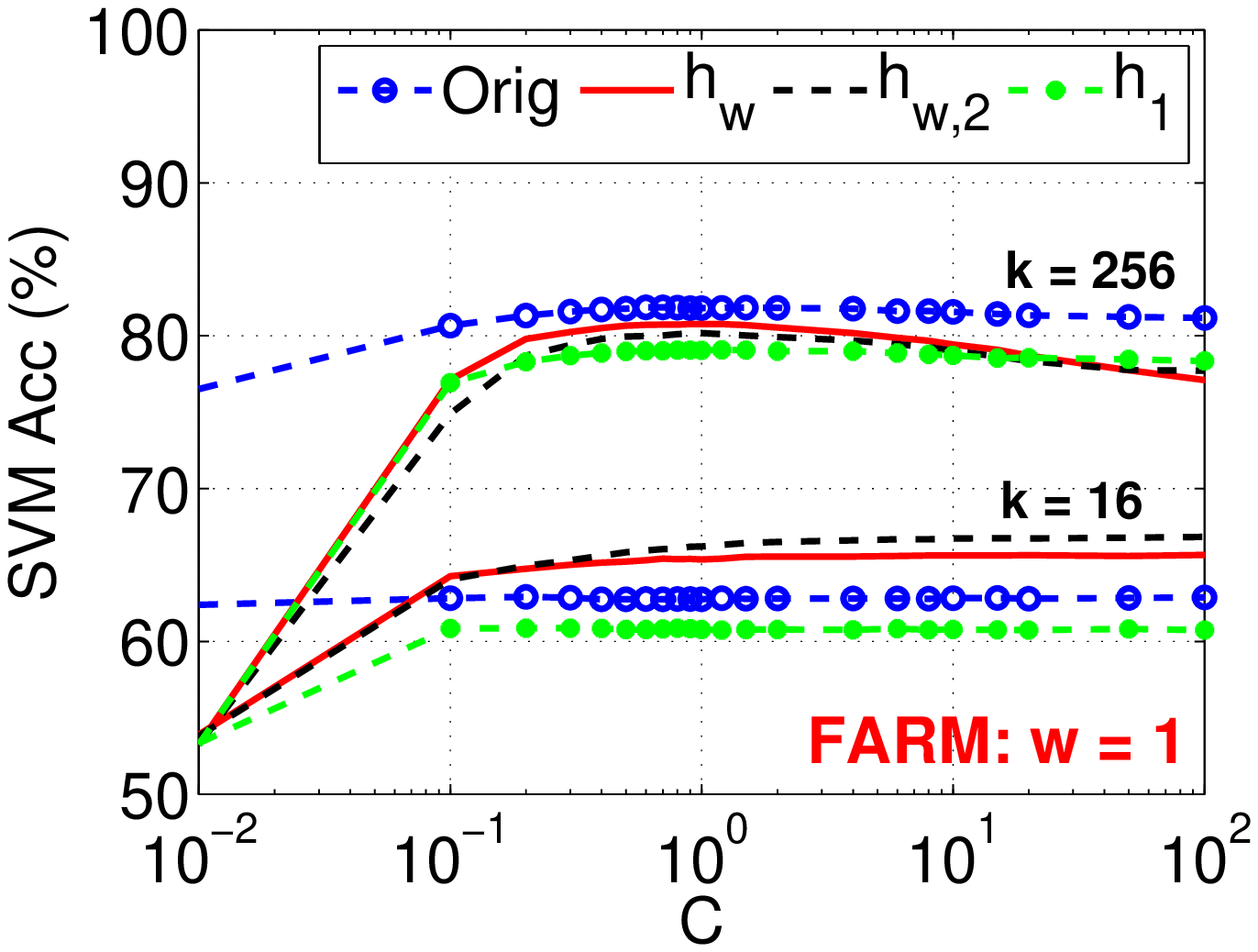}
\includegraphics[width=2.2in]{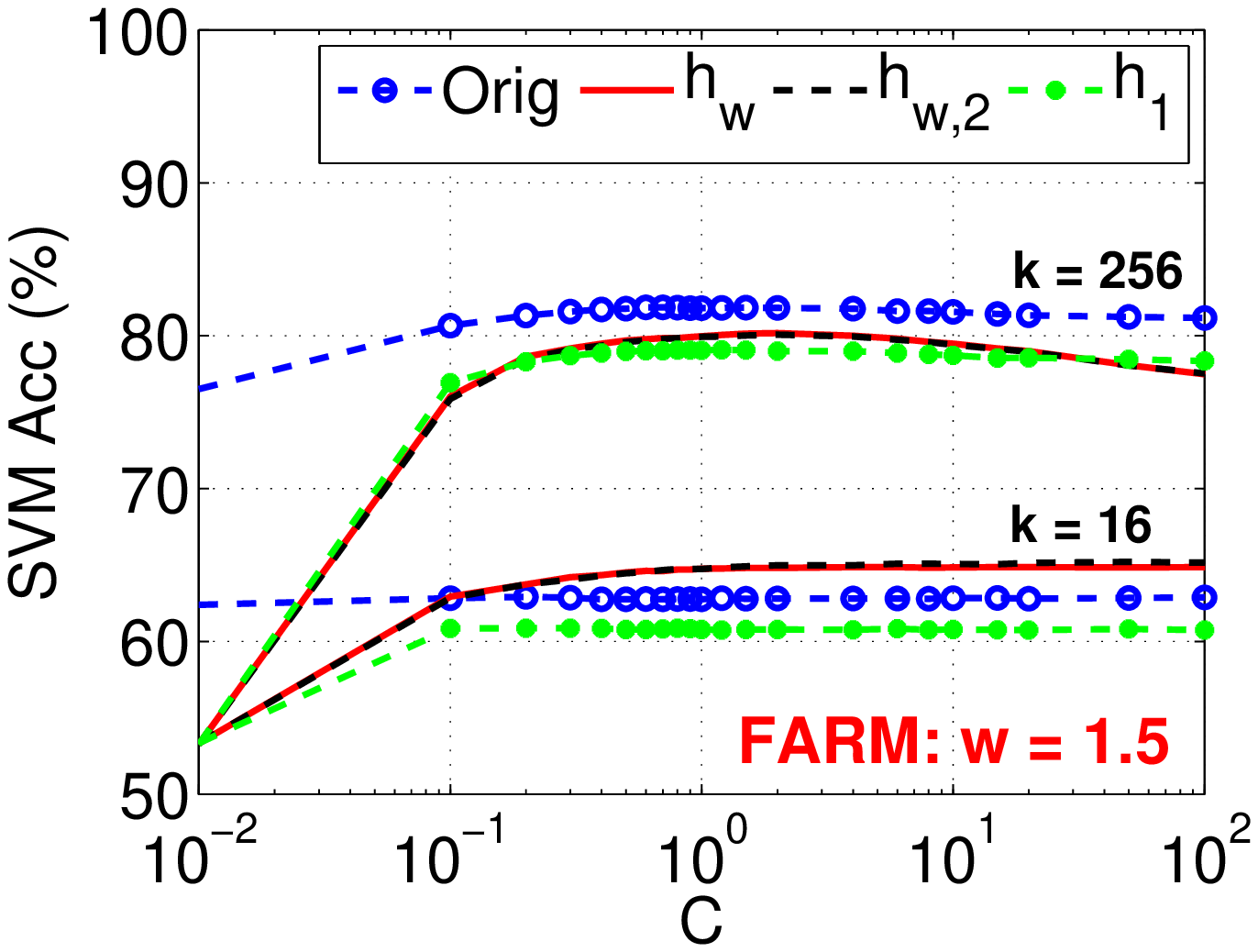}
\includegraphics[width=2.2in]{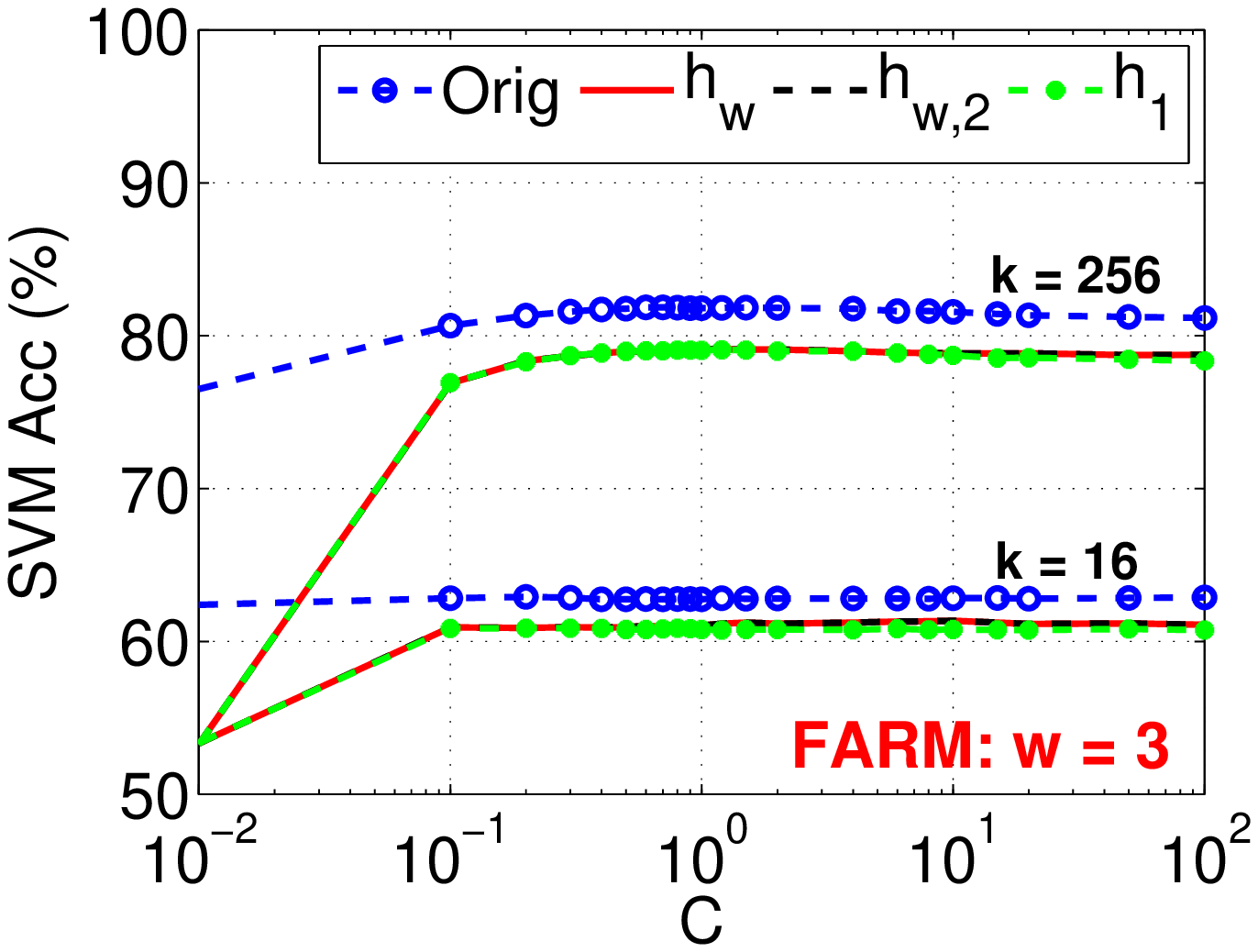}
}

\end{center}
\vspace{-0.2in}
\caption{Test accuracies on the {\em FARM} dataset using LIBLINEAR, for comparing four coding schemes: uncoded (``Orig''), $h_w$, $h_{w,2}$, and $h_1$. We report the results for $k=16$ and $k=256$.  }\label{fig_FarmSVMAcc}
\end{figure}

Figure~\ref{fig_UrlSVMAcc} reports the test classification accuracies (averaged over 20 repetitions) for the {\em URL} dataset.  When $w=0.5\sim 1$, both $h_w$ and $h_{w,2}$ produce similar results as using the original projected data. The 1-bit scheme $h_1$ is obviously less competitive. We provide similar plots (Figure~\ref{fig_FarmSVMAcc}) for the {\em FARM} dataset.
\newpage

\newpage

\begin{figure}[h!]
\begin{center}

\mbox{
\includegraphics[width=2.2in]{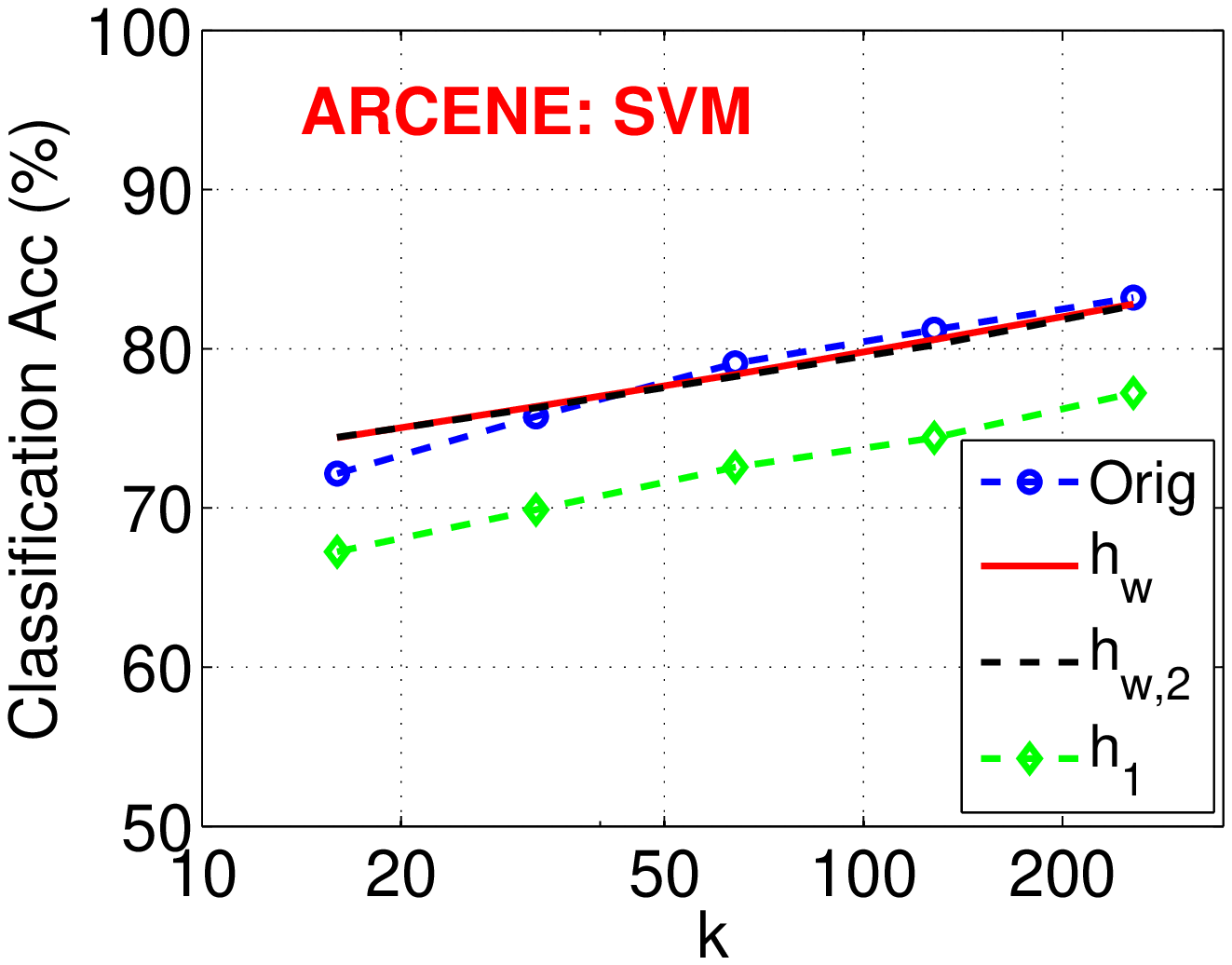}
\includegraphics[width=2.2in]{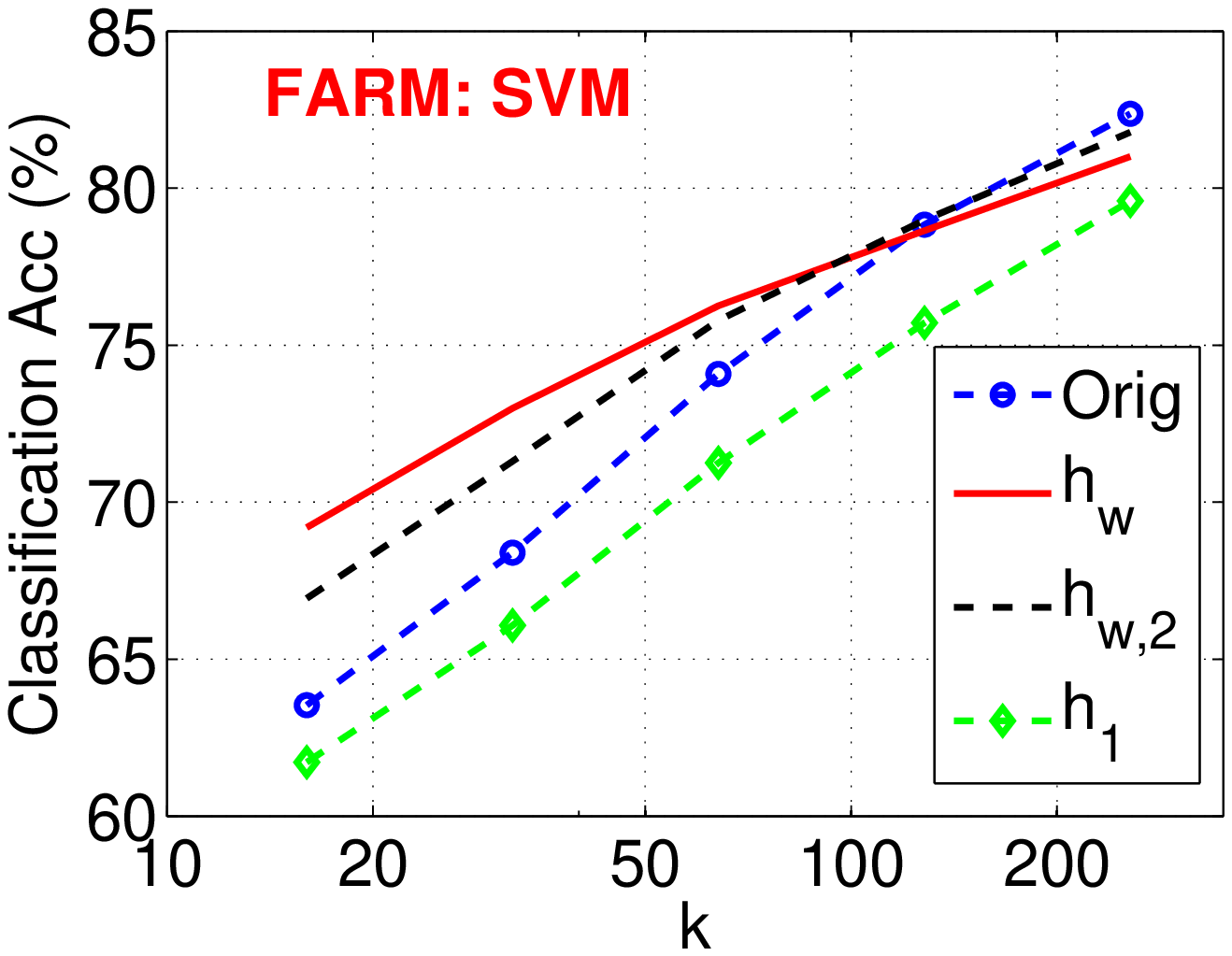}
\includegraphics[width=2.2in]{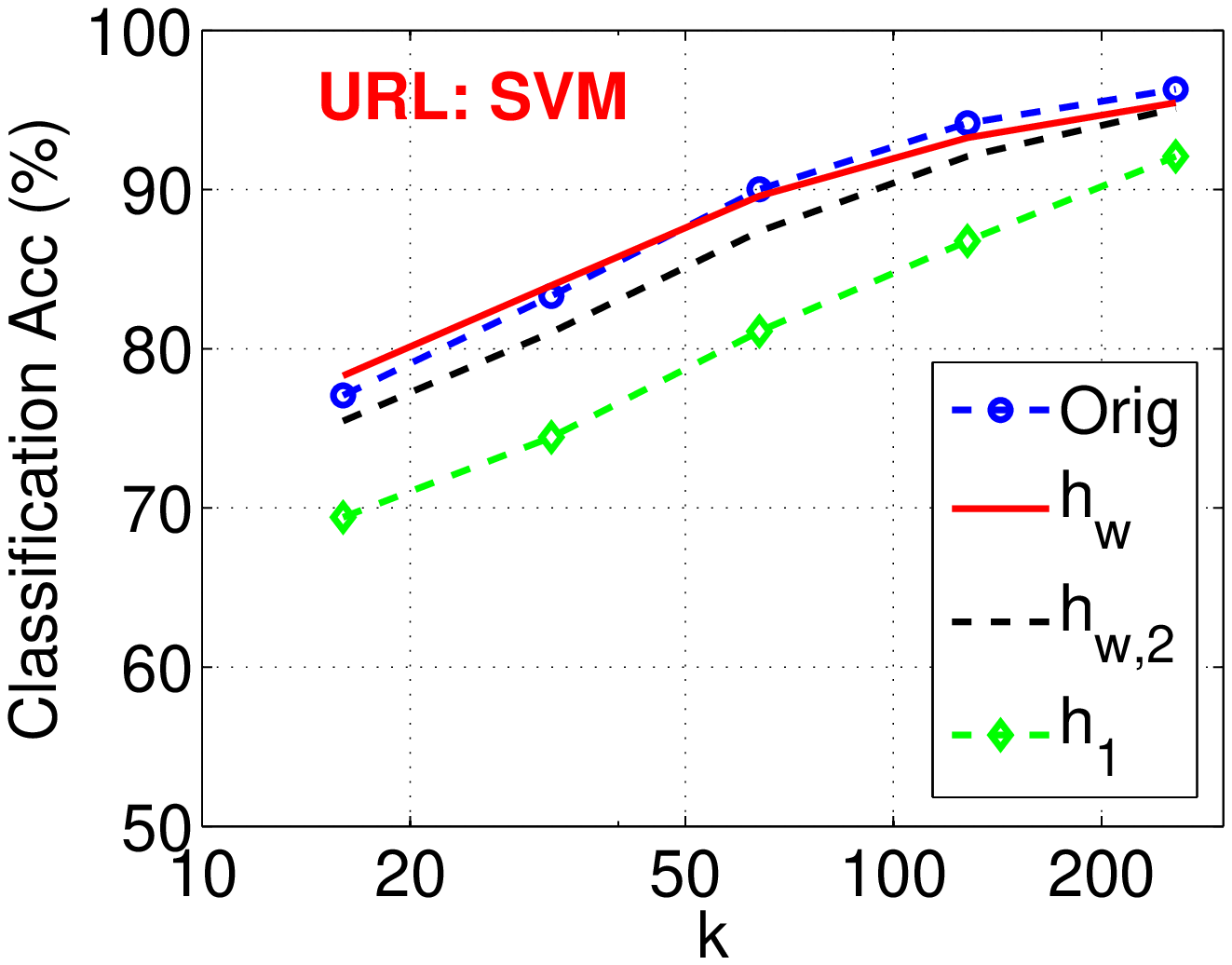}
}
\mbox{
\includegraphics[width=2.2in]{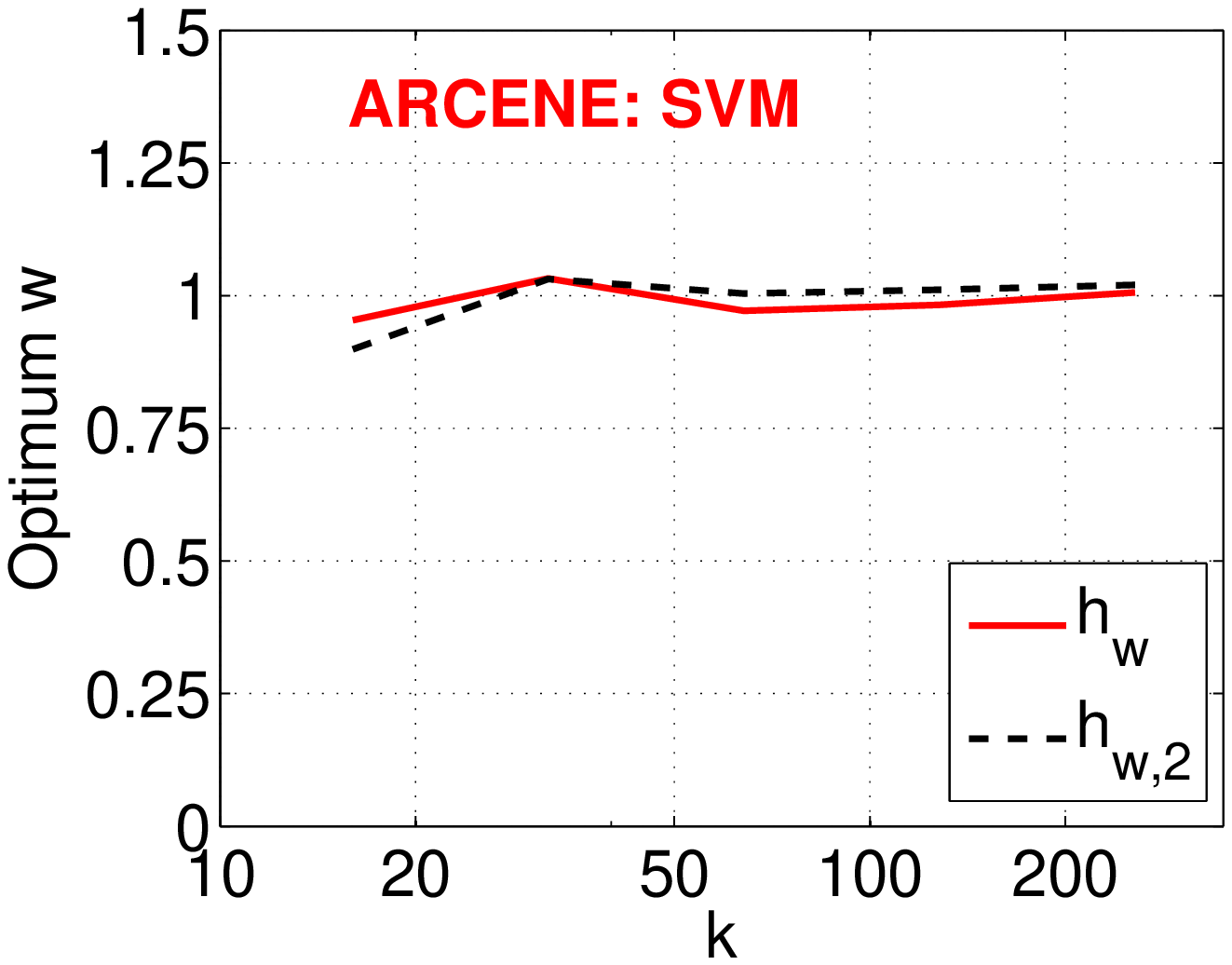}
\includegraphics[width=2.2in]{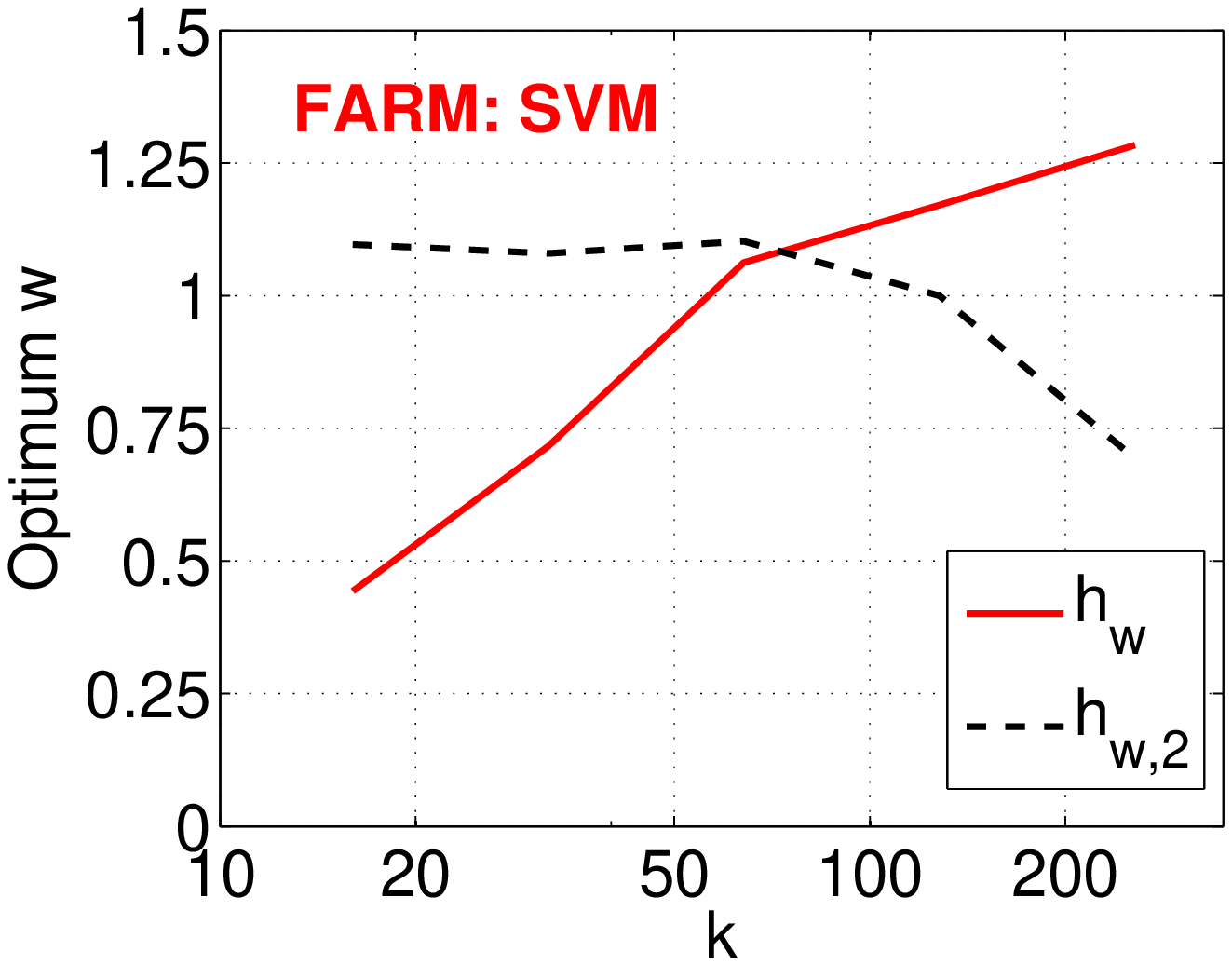}
\includegraphics[width=2.2in]{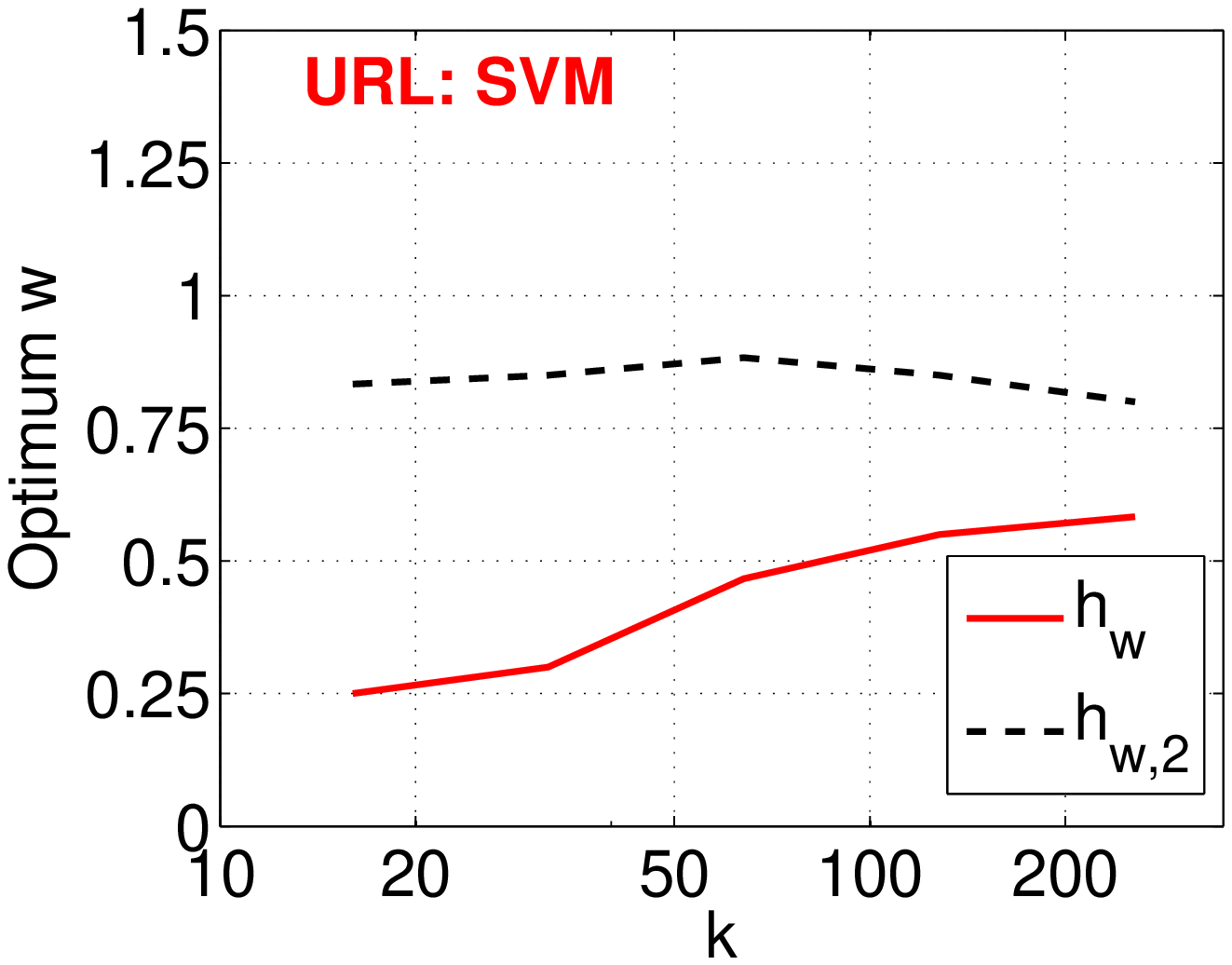}
}

\end{center}
\vspace{-0.20in}
\caption{Summary of the linear SVM results on three datasets. The upper panels  report, for each $k$, the best (highest) classification accuracies among all $C$ values and $w$  values (for $h_{w,2}$ and $h_{w}$). The bottom panels report the $w$ values at which the best accuracies were attained.}\label{fig_SVMmaxAcc}
\end{figure}

We summarize the experiments in Figure~\ref{fig_SVMmaxAcc} for all three datasets.  The upper panels  report, for each $k$, the best (highest) test classification accuracies among all $C$ values and $w$  values (for $h_{w,2}$ and $h_{w}$). The results show a clear trend: (i) the 1-bit ($h_1$)  scheme produces noticeably lower accuracies compared to others;  (ii) the performances of $h_{w,2}$ and $h_w$ are quite similar. The bottom panels of Figure~\ref{fig_SVMmaxAcc} report the $w$ values at which the best accuracies were attained. For $h_{w,2}$, the optimum $w$ values are often close to 1. One interesting observation is that for the {\em FARM} dataset, using the coded data (by $h_w$ or $h_{w,2}$) can actually produce better accuracy than using the original (uncoded) data, when $k$ is not large. This phenomenon may not be too surprising because quantization may be also viewed as some form of regularization and in some cases may help boost the performance.

\section{Future Work}\label{sec_future}

This paper only studies linear estimators, which can be written as inner products. Linear estimators are extremely useful because they allow highly efficient implementation of linear classifiers (e.g., linear SVM) and  near neighbor search methods using hash tables. For applications that allow \textbf{nonlinear estimators} (e.g., nonlinear kernel SVM), we can  substantially improve linear estimators by solving nonlinear MLE (maximum likelihood) equations.  The analysis will be reported separately.

Our work is, to an extent, inspired by the recent work on $b$-bit minwise hashing~\cite{Proc:Li_Konig_WWW10,Proc:Li_Owen_Zhang_NIPS12}, which also proposed a coding scheme for minwise hashing and applied it to learning applications where the data are binary and sparse. Our work is for general data types, as opposed to binary, sparse data.  We expect coding methods will also prove valuable for other variations of random projections, including the count-min sketch \cite{Article:Cormode_05} and related variants~\cite{Proc:Weinberger_ICML2009} and very sparse random projections~\cite{Proc:Li_KDD07}. Another potentially interesting future direction is to develop refined coding schemes for improving {\em sign stable projections}~\cite{Report:SignStable2013} (which are useful for $\chi^2$ similarity estimation, a popular similarity measure in computer vision and NLP).

\section{Conclusion}\label{sec_conclusion}

The method of random projections has become a standard algorithmic approach for computing  distances or correlations in massive, high-dimensional datasets. A compact representation (coding) of the projected data is crucial for efficient transmission, retrieval, and energy consumption.  We have compared a simple scheme based on uniform quantization with the influential coding scheme using windows with a random offset~\cite{Proc:Datar_SCG04}; our scheme appears operationally simpler, more accurate, not as sensitive to parameters (e.g., the widow/bin width $w$), and uses fewer bits.   We furthermore develop a 2-bit non-uniform coding scheme which performs  similarly to uniform quantization. Our experiments with linear SVM on several real-world high-dimensional datasets confirm the efficacy of the two proposed coding schemes.  Based on the theoretical analysis and empirical evidence, we recommend the use of the 2-bit non-uniform coding scheme with the first bin width $w=0.7\sim1$, especially when the target similarity level is high.

\appendix

\section{Proof of Lemma~\ref{lem_Pst}}\label{app_lem_Pst}

The  joint density function of  $(x,y)$  is
$f(x,y;\rho) = \frac{1}{2\pi\sqrt{1-\rho^2}}e^{-\frac{x^2-2\rho xy +y^2}{2(1-\rho^2)}}, \ \ -1\leq \rho\leq 1$.
In this paper we focus on $\rho\geq 0$. We use the usual notation for standard normal pdf and cdf: $\phi(x) = \frac{1}{\sqrt{2\pi}}e^{-x^2/2}$,  $\Phi(x) = \int_{-\infty}^x \phi(x) dx$. The probability $Q_{s,t}$ can be simplified to be
\begin{align}\notag
Q_{s,t}=&  \int_{s}^{t}\int_{s}^{t}\frac{1}{2\pi\sqrt{1-\rho^2}}e^{-\frac{x^2-2\rho xy+y^2}{2(1-\rho^2)}}dx dy\\\notag
=&\int_{s}^{t}\int_{s}^{t}\frac{1}{2\pi\sqrt{1-\rho^2}}e^{-\frac{x^2-2\rho xy+y^2}{2(1-\rho^2)}}dydx\\\notag
=&\int_{s}^{t}\frac{1}{2\pi\sqrt{1-\rho^2}}e^{-\frac{x^2}{2}}\int_{s}^{t}
e^{-\frac{(y-\rho x)^2}{2(1-\rho^2)}}dydx\\\notag
=&\int_{s}^{t}\frac{1}{2\pi\sqrt{1-\rho^2}}e^{-\frac{x^2}{2}}\int_{\frac{s-\rho x}{\sqrt{1-\rho^2}}}^{\frac{t-\rho x}{\sqrt{1-\rho^2}}}
e^{-\frac{u^2}{2}}\sqrt{1-\rho^2}dudx\\\notag
=&\int_{s}^{t}\frac{1}{\sqrt{2\pi}}e^{-\frac{x^2}{2}}\int_{\frac{s-\rho x}{\sqrt{1-\rho^2}}}^{\frac{t-\rho x}{\sqrt{1-\rho^2}}}\frac{1}{\sqrt{2\pi}}
e^{-\frac{u^2}{2}}dudx\\\notag
=&\int_{s}^{t}\frac{1}{\sqrt{2\pi}}e^{-\frac{x^2}{2}}\left[\Phi\left(\frac{t-\rho x}{\sqrt{1-\rho^2}}\right)- \Phi\left(\frac{s-\rho x}{\sqrt{1-\rho^2}}\right)\right]dx
\end{align}
Next we evaluate its derivative $\frac{\partial Q_{s,t}(\rho,s)}{\partial \rho}$.
\begin{align}\notag
\frac{\partial Q_{s,t}(\rho,s)}{\partial \rho} =& \int_{s}^{t}\frac{1}{\sqrt{2\pi}}e^{-\frac{x^2}{2}}\left(\phi\left(\frac{t-\rho x}{\sqrt{1-\rho^2}}\right)\frac{-x+t\rho}{(1-\rho^2)^{3/2}}-\phi\left(\frac{s-\rho x}{\sqrt{1-\rho^2}}\right)\frac{-x+s\rho}{(1-\rho^2)^{3/2}}\right) dx
\end{align}
Note that
\begin{align}\notag
\frac{\partial}{\partial \rho}\left(\Phi\left(\frac{t-\rho x}{\sqrt{1-\rho^2}}\right)\right) = \phi\left(\frac{t-\rho x}{\sqrt{1-\rho^2}}\right)\frac{-x+t\rho}{(1-\rho^2)^{3/2}}
\end{align}
and
\begin{align}\notag
&\int_{s}^{t}\frac{1}{\sqrt{2\pi}}e^{-\frac{x^2}{2}}\phi\left(\frac{t-\rho x}{\sqrt{1-\rho^2}}\right)\frac{-x+t\rho}{(1-\rho^2)^{3/2}}dx\\\notag
 =& \int_{s}^{t}\frac{1}{{2\pi}}e^{-\frac{x^2+t^2-2t\rho x}{2(1-\rho^2)}}\frac{-x+t\rho}{(1-\rho^2)^{3/2}}dx\\\notag
 =& \int_{s}^{t}\frac{1}{{2\pi}}e^{-\frac{(x-t\rho)^2}{2(1-\rho^2)}}e^{-\frac{t^2}{2}}\frac{-x+t\rho}{(1-\rho^2)^{3/2}}dx\\\notag
 =&\frac{1}{2\pi}\frac{1}{(1-\rho^2)^{1/2}}e^{-t^2/2}\left.e^{-\frac{(x-t\rho)^2}{2(1-\rho^2)}}\right|_s^t \\\notag
 =&\frac{1}{2\pi}\frac{1}{(1-\rho^2)^{1/2}}e^{-t^2/2}\left[ e^{-\frac{t^2(1-\rho)}{2(1+\rho)}}-e^{-\frac{\left(s-t\rho\right)^2}{2(1-\rho^2)}} \right]
 \end{align}
 and
\begin{align}\notag
&\int_{s}^{t}\frac{1}{\sqrt{2\pi}}e^{-\frac{x^2}{2}}\phi\left(\frac{s-\rho x}{\sqrt{1-\rho^2}}\right)\frac{-x+s\rho}{(1-\rho^2)^{3/2}}dx\\\notag
  =&\frac{1}{2\pi}\frac{1}{(1-\rho^2)^{1/2}}e^{-s^2/2}\left.e^{-\frac{(x-s\rho)^2}{2(1-\rho^2)}}\right|_s^t\\\notag
 =&\frac{1}{2\pi}\frac{1}{(1-\rho^2)^{1/2}}e^{-s^2/2}\left[ -e^{-\frac{s^2(1-\rho)}{2(1+\rho)}}+e^{-\frac{\left(t-s\rho\right)^2}{2(1-\rho^2)}} \right]
 \end{align}

Combining the results, we obtain
\begin{align}\notag
&\frac{\partial Q_{s,t}(\rho,s)}{\partial \rho}\\\notag
 =& \int_{s}^{t}\frac{1}{\sqrt{2\pi}}e^{-\frac{x^2}{2}}\left(\phi\left(\frac{t-\rho x}{\sqrt{1-\rho^2}}\right)\frac{-x+t\rho}{(1-\rho^2)^{3/2}}-\phi\left(\frac{s-\rho x}{\sqrt{1-\rho^2}}\right)\frac{-x+s\rho}{(1-\rho^2)^{3/2}}\right) dx\\\notag
=&\frac{1}{2\pi}\frac{1}{(1-\rho^2)^{1/2}}e^{-t^2/2}\left[ e^{-\frac{t^2(1-\rho)}{2(1+\rho)}}-e^{-\frac{\left(s-t\rho\right)^2}{2(1-\rho^2)}} \right]
-\frac{1}{2\pi}\frac{1}{(1-\rho^2)^{1/2}}e^{-s^2/2}\left[ -e^{-\frac{s^2(1-\rho)}{2(1+\rho)}}+e^{-\frac{\left(t-s\rho\right)^2}{2(1-\rho^2)}} \right]\\\notag
=&\frac{1}{2\pi}\frac{1}{(1-\rho^2)^{1/2}}\left(e^{-\frac{t^2}{(1+\rho)}} -e^{-\frac{t^2+s^2-2st\rho}{2(1-\rho^2)}}
+e^{-\frac{s^2}{(1+\rho)}} -e^{-\frac{t^2+s^2-2st\rho}{2(1-\rho^2)}}
\right)\\\notag
=&\frac{1}{2\pi}\frac{1}{(1-\rho^2)^{1/2}}\left(e^{-\frac{t^2}{(1+\rho)}}+e^{-\frac{s^2}{(1+\rho)}} -2e^{-\frac{t^2+s^2-2st\rho}{2(1-\rho^2)}}\right)\\\notag
=&\frac{1}{2\pi}\frac{1}{(1-\rho^2)^{1/2}}\left[\left(e^{-\frac{t^2}{2(1+\rho)}}-e^{-\frac{s^2}{2(1+\rho)}} \right)^2 +2e^{-\frac{t^2+s^2}{2(1+\rho)}}-2e^{-\frac{t^2+s^2-2st\rho}{2(1-\rho^2)}}\right]\\\notag
 \geq& 0
\end{align}
The last inequality holds because
\begin{align}\notag
&\left[-\frac{t^2+s^2}{2(1+\rho)}\right]-\left[-\frac{t^2+s^2-2st\rho}{2(1-\rho^2)}\right]\\\notag
=&\frac{1}{2(1-\rho^2)}\left[-(t^2+s^2)(1-\rho) +t^2+s^2-2st\rho\right]\\\notag
=&\frac{\rho}{2(1-\rho^2)}\left[s-t\right]^2\\\notag
\geq& 0
\end{align}

This completes the proof.

\section{Proof of Theorem~\ref{thm_Vwq}}\label{app_thm_Vwq}

From the collision probability,
$P_{w,q} = 2\Phi\left(\frac{w}{\sqrt{d}}\right)-1-\frac{2}{\sqrt{2\pi}w/\sqrt{d}}+\frac{2}{w/\sqrt{d}}\phi\left(\frac{w}{\sqrt{d}}\right)$,
we can estimate $d$ (and $\rho$). Recall $d = 2(1-\rho)$.  We denote the estimator by $\hat{d}_{w,q}$ (and $\hat{\rho}_{w,q}$), from the empirical probability $\hat{P}_{w,q}$, which is estimated without bias from $k$ projections.

Note that $\hat{d}_{w,q} = g(\hat{P}_{w,q})$ for a nonlinear function $g$. As $k\rightarrow\infty$, the estimator  $\hat{d}_{w,q}$ is asymptotically unbiased. The variance can be determined by the ``delta'' method:
\begin{align}\notag
Var\left(\hat{d}_{w,q}\right) = Var\left(\hat{P}_{w,q}\right) \left[g^\prime\left(P_{w,q}\right)\right]^2 + O\left(\frac{1}{k^2}\right)
 = \frac{1}{k}P_{w,q}\left(1-{P}_{w,q}\right) \left[g^\prime\left(P_{w,q}\right)\right]^2 + O\left(\frac{1}{k^2}\right)
\end{align}
Since
\begin{align}\notag
\frac{\partial {P}_{w,q}} {\partial {d}}=& -\phi\left(\frac{w}{\sqrt{d}}\right)wd^{-3/2}-\frac{1}{\sqrt{2\pi}w}d^{-1/2}+\frac{1}{w}d^{-1/2}\phi\left(\frac{w}{\sqrt{d}}\right)
+\frac{1}{w/\sqrt{d}}\phi\left(\frac{w}{\sqrt{d}}\right)\left(\frac{w}{\sqrt{d}}\right)wd^{-3/2}\\\notag
=&\frac{1}{w/\sqrt{d}}d^{-1}\phi\left(\frac{w}{\sqrt{d}}\right)-\frac{1}{\sqrt{2\pi}w/\sqrt{d}}d^{-1}
\end{align}
we have
\begin{align}\notag
g^\prime\left(P_{w,q}\right) = \frac{1}{\frac{\partial {P}_{w,q}} {\partial {d}}}= \frac{w/\sqrt{d}}{\phi\left(w/\sqrt{d}\right)-1/\sqrt{2\pi}}d
\end{align}
and
\begin{align}\notag
Var\left(\hat{d}_{w,q}\right) =  \frac{d^2}{k}\left(\frac{w/\sqrt{d}}{\phi\left(w/\sqrt{d}\right)-1/\sqrt{2\pi}}\right)^2P_{w,q}(1-P_{w,q})
 + O\left(\frac{1}{k^2}\right)
 \end{align}
Because $\rho = 1-d/2$, we know that
\begin{align}\notag
Var\left(\hat{\rho}_{w,q}\right) =  \frac{1}{4}Var\left(\hat{d}_{w,q}\right) = \frac{V_{w,q}}{k} + O\left(\frac{1}{k^2}\right)
 \end{align}
where
\begin{align}\notag
V_{w,q} = d^2/4\left(\frac{w/\sqrt{d}}{\phi\left(w/\sqrt{d}\right)-1/\sqrt{2\pi}}\right)^2P_{w,q}(1-P_{w,q})
\end{align}
This completes the proof.

\section{Proof of Theorem~\ref{thm_Vw}}\label{app_thm_Vw}
This proof is similar to the proof of Theorem~\ref{thm_Vwq}. To evaluate the asymptotic variance, we need to compute $\frac{\partial P_w}{\partial \rho}$:
\begin{align}\notag
\frac{\partial P_w }{\partial \rho}=&\frac{1}{\pi}\frac{1}{(1-\rho^2)^{1/2}}\sum_{i=0}^\infty \left(e^{-\frac{(i+1)^2w^2}{(1+\rho)}}
+e^{-\frac{i^2w^2}{(1+\rho)}} -2e^{-\frac{(i+1)^2w^2+i^2w^2-2i(i+1)w^2\rho}{2(1-\rho^2)}}\right)\\\notag
=&\frac{1}{\pi}\frac{1}{(1-\rho^2)^{1/2}}\sum_{i=0}^\infty \left(e^{-\frac{(i+1)^2w^2}{(1+\rho)}}
+e^{-\frac{i^2w^2}{(1+\rho)}}-2e^{-\frac{w^2}{2(1-\rho^2)}}e^{-\frac{i(i+1)w^2}{1+\rho}}\right)
\end{align}
Thus,
\begin{align}\notag
&Var\left(\hat{\rho}_{w}\right) = \frac{V_{w}}{k} + O\left(\frac{1}{k^2}\right),\hspace{0.2in}\text{where} \\\notag
&V_{w} = \frac{\pi^2(1-\rho^2)P_w(1-P_w)}{\left[\sum_{i=0}^\infty \left(e^{-\frac{(i+1)^2w^2}{(1+\rho)}}
+e^{-\frac{i^2w^2}{(1+\rho)}}-2e^{-\frac{w^2}{2(1-\rho^2)}}e^{-\frac{i(i+1)w^2}{1+\rho}}\right)\right]^2}
\end{align}
Next, we consider the special  case with $\rho\rightarrow0$.
\begin{align}\notag
\left.P_{w}\right|_{\rho=0} =& 2\sum_{i=0}^\infty \left(\Phi((i+1)w) - \Phi(iw)\right)^2
 = 2\sum_{i=0}^\infty \left(\int_{iw}^{(i+1)w} \phi(x) dx \right)^2
 =2w^2\sum_{i=0}^\infty \left(\int_{i}^{i+1} \phi(wx) dx \right)^2
\end{align}
\begin{align}\notag
\left.\frac{\partial P_w }{\partial \rho}\right|_{\rho=0}=\frac{1}{\pi}\sum_{i=0}^\infty \left(e^{-(i+1)^2w^2/2}- e^{-i^2w^2/2}\right)^2
=2\sum_{i=0}^\infty \left(\phi((i+1)w) - \phi(iw)\right)^2
\end{align}
Combining the results, we obtain
\begin{align}\notag
\left.V_{w}\right|_{\rho=0} =& \frac{2w^2\sum_{i=0}^\infty \left(\int_{i}^{i+1} \phi(wx) dx \right)^2\left(1-2w^2\sum_{i=0}^\infty \left(\int_{i}^{i+1} \phi(wx) dx \right)^2\right) }{\left(\frac{1}{\pi}\sum_{i=0}^\infty \left(e^{-(i+1)^2w^2/2}- e^{-i^2w^2/2}\right)^2\right)^2}\\\notag
=&\frac{\sum_{i=0}^\infty \left(\Phi((i+1)w) - \Phi(iw)\right)^2\left(1/2-\sum_{i=0}^\infty \left(\Phi((i+1)w) - \Phi(iw)\right)^2\right)}{\left(\sum_{i=0}^\infty \left(\phi((i+1)w) - \phi(iw)\right)^2\right)^2}
 \end{align}

\section{Proof of Theorem~\ref{thm_hw2}}\label{app_thm_hw2}

\begin{align}\notag
&P_{w,2} = \mathbf{Pr}\left(h_{w,2}^{(j)}(u) = h^{(j)}_{w,2}(v)\right)\\\notag
=&2\int_{0}^{w}\phi(x)\left[\Phi\left(\frac{w-\rho x}{\sqrt{1-\rho^2}}\right)- \Phi\left(\frac{-\rho x}{\sqrt{1-\rho^2}}\right)\right]dx
 + 2\int_{w}^{\infty}\phi(x)\left[1- \Phi\left(\frac{w-\rho x}{\sqrt{1-\rho^2}}\right)\right]dx\\\notag
 =&2\int_{0}^{w}\phi(x)\left[\Phi\left(\frac{w-\rho x}{\sqrt{1-\rho^2}}\right)- 1+\Phi\left(\frac{\rho x}{\sqrt{1-\rho^2}}\right)\right]dx
 + 2\int_{w}^{\infty}\phi(x)\left[1- \Phi\left(\frac{w-\rho x}{\sqrt{1-\rho^2}}\right)\right]dx\\\notag
  =&4\int_{0}^{w}\phi(x)\left[\Phi\left(\frac{w-\rho x}{\sqrt{1-\rho^2}}\right)- 1\right]dx+2\int_{0}^{w}\phi(x)\Phi\left(\frac{\rho x}{\sqrt{1-\rho^2}}\right)dx
 + 2\int_{0}^{\infty}\phi(x)\left[1- \Phi\left(\frac{w-\rho x}{\sqrt{1-\rho^2}}\right)\right]dx\\\notag
  =&-4\int_{0}^{w}\phi(x)\Phi\left(\frac{-w+\rho x}{\sqrt{1-\rho^2}}\right)dx+2\int_{0}^{w}\phi(x)\Phi\left(\frac{\rho x}{\sqrt{1-\rho^2}}\right)dx
 + 2\int_{0}^{\infty}\phi(x)\left[1- \Phi\left(\frac{w-\rho x}{\sqrt{1-\rho^2}}\right)\right]dx\\\notag
 =&1 - \frac{1}{\pi}\cos^{-1}\rho - 4\int_{0}^{w}\phi(x)\Phi\left(\frac{-w+\rho x}{\sqrt{1-\rho^2}}\right)dx
\end{align}
We need to show
\begin{align}\notag
g(\rho) = \int_{0}^{w}\phi(x)\Phi\left(\frac{\rho x}{\sqrt{1-\rho^2}}\right)dx
 + \int_{0}^{\infty}\phi(x)\left[1- \Phi\left(\frac{w-\rho x}{\sqrt{1-\rho^2}}\right)\right]dx = \frac{1}{2}-\frac{1}{2\pi}\cos^{-1}\rho
\end{align}
Because
\begin{align}\notag
g^\prime(\rho) =& \int_{0}^{w}\phi(x)\phi\left(\frac{\rho x}{\sqrt{1-\rho^2}}\right) \frac{x}{(1-\rho^2)^{3/2}}xdx
+\int_{0}^{\infty}\phi(x)\phi\left(\frac{w-\rho x}{\sqrt{1-\rho^2}}\right)\frac{x-\rho w}{(1-\rho)^{3/2}}dx\\\notag
=&\int_{0}^{w}\frac{1}{2\pi}e^{-\frac{x^2}{2(1-\rho^2)}} \frac{x}{(1-\rho^2)^{3/2}}dx +\int_{0}^{\infty}\frac{1}{2\pi}e^{-\frac{(x-\rho w)^2+w^2-w^2\rho^2}{2(1-\rho^2)}} \frac{x-\rho w}{(1-\rho^2)^{3/2}}xdx\\\notag
=&\frac{1}{2\pi}\frac{1}{(1-\rho^2)^{1/2}}\left(1-e^{-\frac{w^2}{2(1-\rho^2)}}\right)
+\frac{1}{2\pi}\frac{1}{(1-\rho^2)^{1/2}}e^{-\frac{w^2}{2}}e^{-\frac{\rho^2w^2}{2(1-\rho^2)}}\\\notag
=&\frac{1}{2\pi}\frac{1}{(1-\rho^2)^{1/2}}
\end{align}
we know
\begin{align}\notag
g(\rho) =& \int_0^\rho g^\prime(\rho) d\rho + g(0) \\\notag
=& \frac{1}{2\pi}\sin^{-1}\rho + (\Phi(w)-1/2)/2 + (1-\Phi(w))/2\\\notag
=& \frac{1}{2\pi}\sin^{-1}\rho + \frac{1}{4} = \frac{1}{2}-\frac{1}{2\pi}\cos^{-1}\rho
\end{align}
Also,
\begin{align}\notag
\frac{\partial P_{w,2}}{\partial \rho} =& \frac{1}{\pi}\frac{1}{\sqrt{1-\rho^2}} -4 \int_{0}^{w}\phi(x)\phi\left(\frac{-w+\rho x}{\sqrt{1-\rho^2}}\right)\frac{x-\rho w}{(1-\rho^2)^{3/2}}dx\\\notag
=&\frac{1}{\pi}\frac{1}{\sqrt{1-\rho^2}}-\frac{4}{2\pi}\frac{1}{(1-\rho^2)^{1/2}}e^{-\frac{w^2}{2}}\left(e^{-\frac{\rho^2w^2}{2(1-\rho^2)}}-
e^{-\frac{(w-\rho w)^2}{2(1-\rho^2)}}\right)\\\notag
=&\frac{1}{\pi}\frac{1}{\sqrt{1-\rho^2}}-\frac{2}{\pi}\frac{1}{\sqrt{1-\rho^2}}\left(e^{-\frac{w^2}{2(1-\rho^2)}}-
e^{-\frac{w^2}{1+\rho}}\right)\\\notag
\end{align}
Thus, combining the results, we obtain
\begin{align}\notag
Var\left(\hat{\rho}_{w,2}\right) = \frac{V_{w,2}}{k} + O\left(\frac{1}{k^2}\right)
\end{align}
where
\begin{align}\notag
V_{w,2} = \frac{\pi^2(1-\rho^2)P_{w,2}(1-P_{w,2})}{\left[1-2e^{-\frac{w^2}{2(1-\rho^2)}} + 2e^{-\frac{w^2}{1+\rho}}\right]^2}
\end{align}
This completes the proof.


\end{document}